\documentclass{article}

\usepackage{microtype}
\usepackage{graphicx}
\usepackage{subfigure}
\usepackage{booktabs} 
\usepackage[ruled,vlined]{algorithm2e}
\usepackage{graphicx}
\usepackage{changepage}
\usepackage{subcaption}
\usepackage{multirow}
\usepackage{xcolor} 
\usepackage{caption}
\usepackage{amsfonts}
\usepackage{algpseudocode}
\usepackage{ulem}
\usepackage{booktabs}

\usepackage[accepted]{icml2025}

\usepackage{amsmath}
\usepackage{amssymb}
\usepackage{mathtools}
\usepackage{amsthm}

\usepackage[capitalize,noabbrev]{cleveref}

\theoremstyle{plain}
\newtheorem{theorem}{Theorem}[section]

\newtheorem{lemma}[theorem]{Lemma}

\theoremstyle{definition}
\newtheorem{definition}[theorem]{Definition}

\theoremstyle{remark}

\usepackage[textsize=tiny]{todonotes}

\begin{document}

\twocolumn[
\icmltitle{Real-Time Privacy Risk Measurement with Privacy Tokens for Gradient Leakage}



\icmlsetsymbol{equal}{*}

\begin{icmlauthorlist}
\icmlauthor{Jiayang Meng}{equal,yyy}
\icmlauthor{Tao Huang}{equal,comp}
\icmlauthor{Hong Chen}{yyy}
\icmlauthor{Xin Shi}{comp}
\icmlauthor{Qingyu Huang}{comp}
\icmlauthor{Chen Hou}{comp}

\end{icmlauthorlist}

\icmlaffiliation{yyy}{School of Information, Renmin University of China, Beijing, China}
\icmlaffiliation{comp}{School of Computer and Big Data, Minjiang University, Fuzhou, Fujian, China}

\icmlcorrespondingauthor{Hong Chen}{chong@ruc.edu.cn}

\icmlkeywords{Machine Learning, ICML}

\vskip 0.3in
]



\printAffiliationsAndNotice{\icmlEqualContribution} 

\begin{abstract}
The widespread deployment of deep learning models in privacy-sensitive domains has amplified concerns regarding privacy risks, particularly those stemming from gradient leakage during training. Current privacy assessments primarily rely on post-training attack simulations. However, these methods are inherently reactive, unable to encompass all potential attack scenarios, and often based on idealized adversarial assumptions. These limitations underscore the need for proactive approaches to privacy risk assessment during the training process. To address this gap, we propose the concept of privacy tokens, which are derived directly from private gradients during training. Privacy tokens encapsulate gradient features and, when combined with data features, offer valuable insights into the extent of private information leakage from training data, enabling real-time measurement of privacy risks without relying on adversarial attack simulations. Additionally, we employ Mutual Information (MI) as a robust metric to quantify the relationship between training data and gradients, providing precise and continuous assessments of privacy leakage throughout the training process. Extensive experiments validate our framework, demonstrating the effectiveness of privacy tokens and MI in identifying and quantifying privacy risks. This proactive approach marks a significant advancement in privacy monitoring, promoting the safer deployment of deep learning models in sensitive applications.
\end{abstract}

\section{Introduction}

The remarkable progress of deep learning in fields such as natural language processing \cite{otter2020survey}, computer vision \cite{o2020deep}, and autonomous systems \cite{zhang2020autonomous} has brought immense benefits. However, with these advancements come pressing privacy concerns \cite{rigaki2023survey} \cite{nasr2018comprehensive} \cite{shafee2021privacy}, especially as deep learning models are increasingly deployed in privacy-sensitive sectors like healthcare, finance, and autonomous driving. In particular, models trained on sensitive data can unintentionally expose private information during and after the training process. A key privacy risk stems from gradient leakage, a crucial component in optimizing model parameters \cite{zhu2019deep} \cite{zhao2020idlg} \cite{gao2023automatic} \cite{shafee2021privacy} \cite{rigaki2023survey}, may unintentionally disclose sensitive information. 

While the threat of gradient leakage is well-documented, existing privacy assessments are limited by their reliance on post-training attack simulations \cite{nasr2018comprehensive} \cite{shafee2021privacy} \cite{rigaki2023survey}. These approaches are reactive, impractical for comprehensive real-world testing, and often rely on idealized adversarial assumptions that can not reflect real-world conditions. Moreover, they offer no mechanism for monitoring privacy risks during the training process itself, leaving model trainers blind to potential data exposure until it is too late.

\textbf{Key Questions to Address.} Given the limitations of traditional privacy measures, this paper aims to tackle the following crucial questions:

\begin{itemize}
\item \textbf{Q1:} How much information do gradients reveal about the underlying training data?
\item \textbf{Q2:} Can we quantify the relationship between the training data and gradients, enabling the direct assessment of privacy leakage during model training?
\end{itemize}

Inspired by the concept of the class token in Vision Transformers \cite{dosovitskiy2020image}, which aggregates information from different image patches to make global predictions, we propose embedding gradients into vector representations, which we define as "privacy tokens", to capture the privacy information contained within them. By combining these privacy tokens with the private data features, we can quantitatively assess the relationship between gradient and data privacy, thereby enabling the direct evaluation of privacy risks arising from gradient leakage during training. Similar to how the class token bridges the relationship between image patches and their global class prediction, privacy tokens of intermediate gradients could provide a novel and continuous means of assessing privacy leakage throughout the training process—without requiring adversarial attacks.

\textbf{Why Privacy Token of Intermediate Gradients Can Be Utilized For Privacy Analysis (For Q1).} Privacy tokens of intermediate gradients, similar to class token in Vision Transformers, aggregate and encapsulate key information from the private training data at various stages of training. Intermediate gradients reflect the impact of training data on model’s parameters, acting as a compact representation of how training data influences model. Consequently, privacy tokens of gradients provide a global perspective on privacy risks during training, helping us quantify data exposure without relying on costly attack simulations.

\textbf{Why Privacy Token is Useful for Privacy Protection (For Q2).} Analyzing gradients as privacy tokens, when combined with the private data features, enables real-time privacy monitoring throughout the training process. Proper embeddings of private gradients provide actionable insights into how much information leaked via gradients before training completes, allowing model trainers to detect privacy risks early and enabling timely protective strategies during training. Continuously quantifying this leakage reduces the need for costly post-training attacks and streamlines the process of safeguarding sensitive data in deep learning systems.

Therefore, we attempt to obtain the proper embeddings of the training data and the gradients, then use Mutual Information (MI) as a metric to quantify the amount of private information shared between them. MI, a theoretically grounded and continuous measure, offers a precise way to estimate privacy leakage and provides deeper insights into how susceptible a model may be to gradient-based privacy attacks.

Based on this idea, we introduce the concept of privacy token corresponding to the private gradients and propose a novel framework based on the Mutual Information Neural Estimator (MINE) which combines the privacy token with the embeddings of training data and then estimates the mutual information between training data and gradients. Our approach utilizes MINE to approximate the mutual information through neural networks, ensuring scalability for high-dimensional data. This framework allows for continuous, real-time privacy monitoring during the model training process, offering a proactive solution for privacy management in deep learning.

\textbf{Contributions of This Work.} This paper introduces the following key contributions:

\begin{enumerate} 
\item \textbf{Introduction of Privacy Tokens:} We introduce the concept of privacy tokens derived from intermediate gradients, representing the sensitive information encoded in gradients. These tokens, combined with training data embeddings, enable a continuous, proactive analysis of privacy risks during training. 
\item \textbf{Privacy Leakage Framework Based on MINE Network for MI Estimation:} We design a specialized MINE network for MI estimation that incorporates intermediate layer outputs of the evaluated model as feature representations of raw training data, combined with privacy tokens representing the features of gradients. This framework provides a theoretically grounded and practical metric for evaluating gradient-based privacy risks during model training. 
\item \textbf{Empirical Investigation of Privacy Risks:} Through extensive experiments, we demonstrate that privacy tokens capture sensitive gradient information effectively and that MI reliably quantifies data-gradient correlation. Additionally, we analyze how various factors, including model architecture, dataset characteristics, and training configurations influence, offering deeper insights into privacy risks in deep learning.

\end{enumerate}

\section{Related Work}


\subsection{Gradient-Based Privacy Attacks}

Gradients, as core components of model optimization, contain information about the training data that adversaries can exploit to infer sensitive attributes or reconstruct raw training samples. Numerous attack methods demonstrate the feasibility of extracting private information from gradients, including model inversion \cite{fredrikson2015model} \cite{zhang2020secret} \cite{wang2021variational}, gradient inversion \cite{zhu2019deep} \cite{zhao2021exploiting} \cite{liang2023egia} \cite{ye2024gradient} \cite{fang2023gifd}, and membership inference attacks (MIA) \cite{shokri2017membership} \cite{truex2019demystifying} \cite{choquette2021label} \cite{hu2022membership}. These approaches reveal critical vulnerabilities in gradient sharing, particularly in collaborative settings such as federated learning. However, these methods often rely on specific assumptions about the attacker's prior knowledge and capabilities, which may not be realistic in practice. For example, classical model inversion and gradient inversion attacks, as explored by \cite{fredrikson2015model} and \cite{zhu2019deep}, generally assume that the attacker has detailed access to gradients or partial knowledge of the data distribution. Classical MIAs, such as those proposed by \cite{shokri2017membership}, also depend on assumptions about the attacker's access to similar or representative data samples. These assumptions can be idealized, failing to account for the variability and constraints of real-world attack scenarios. In addition, attack-based approaches cannot comprehensively evaluate data sensitivity, as it is infeasible to anticipate and test every possible privacy attack. Furthermore, they lack a mechanism for monitoring privacy risks during the training phase, leaving model trainers unaware of possible data breaches the risks become irreversible. These limitations emphasize the need for a generalized privacy metric that does not rely solely on specific attacks to assess privacy risks timely. Our work extends this body of research by focusing on quantifying the amount of privacy leakage present in the gradients, without relying on adversarial attacks.

\subsection{Differentially Private Privacy-Preserving Mechanisms}

In response to gradient-based privacy risks, privacy-preserving techniques using differential privacy (DP) have been developed to secure machine learning models. DP aims to lower the sensitivity level of data by limiting information leakage in model training, achieved by introducing controlled noise into gradients or model outputs, ensuring a quantifiable measure of privacy protection as explored by \cite{abadi2016deep} \cite{jayaraman2019evaluating} \cite{nasr2020improving}. However, while DP can lower the sensitivity level of data, it does not explicitly quantify the relationship between the decreased data privacy level and the potential increase in an attacker’s information gain. Current privacy-preserving methods do not provide a direct, continuous measure of the impact these techniques have on data sensitivity relative to potential adversarial capabilities, limiting their utility in assessing how protective measures affect privacy risk in practice. Our work attempts to quantify sensitivity levels between data and gradients, providing insights for privacy budget allocation in DP. And it could also present the reduction of privacy leakage when applying DP in deep learning.

\subsection{Mutual Information and Privacy Risk Assessment}

Mutual information (MI) has been widely used to quantify dependencies between random variables across various fields and is emerging as a potential metric for privacy risk assessment in machine learning \cite{wang2021improving} \cite{he2021drmi} \cite{gao2023detecting}. This concept can be extended to the privacy domain, where MI can quantify the extent to which data characteristics are retained in gradients.

However, directly estimating MI for high-dimensional data is challenging, leading to the development of neural network-based estimators called Mutual Information Neural Estimator (MINE) \cite{belghazi2018mine}. MINE approximates MI by training a neural network to distinguish between joint and product distributions, enabling MI estimation in complex, high-dimensional deep learning settings. 

Although MI has not been widely applied to privacy quantification, it holds significant potential for measuring information leakage. Recent studies have started to explore this avenue \cite{farokhi2020modelling} \cite{wang2021privacy}, examining the utility of MI in quantifying privacy risks by measuring the dependencies between sensitive attributes and model outputs or gradients. This work builds on these foundations by applying MI as a privacy risk metric specifically for gradient leakage. By estimating the MI between training data and gradients, we aim to provide a continuous and theoretically grounded measure of privacy risk that complements existing binary privacy evaluations.

\subsection{Intermediate Representations in Deep Learning}
Intermediate representations, including feature maps and activations, have been studied in the context of model interpretability and adversarial attacks. Researchers have investigated how these intermediate features can expose information about the input data~\cite{kurakin2016adversarial}~\cite{machado2021adversarial}. The use of intermediate gradients, however, has been less explored, especially in the context of privacy leakage. By drawing an analogy to the class token in Vision Transformers, which acts as a global summary of the input data, we propose that the proper embeddings of intermediate gradients can serve as \textit{privacy tokens}. These tokens offer a compact and continuous representation of privacy risks, allowing model trainers to monitor the privacy status of the training data timely throughout the training process. This novel perspective of analyzing intermediate gradients as privacy tokens has not been explored in existing literature and constitutes a key contribution of our work.




\section{Methodology}

This section presents the methodology for estimating mutual information (MI) between raw training samples and gradients to quantify the potential privacy leakage through the privacy token of intermediate gradients in deep learning models. We begin with a formal definition of privacy tokens and MI in the context of privacy estimation. Subsequently, we describe the design of the general approach to MI estimation via the Mutual Information Neural Estimator (MINE) and then outline the specific Mutual Information Estimation (MIE) algorithm tailored for this study.

\subsection{Mutual Information Based on Privacy Tokens As A Privacy Measurement}
First, we give a formal definition of the privacy tokens corresponding to the private gradients and then we discuss how to utilize mutual information as a privacy measurement.
\begin{definition}[Privacy Token]
    Given a deep learning model $F(\mathbf{x};W)$ with parameters $W$ and a training dataset $\mathcal{X}$, for a batch of data $\mathbf{x}_{\text{batch}}$ during the training process, the privacy token of the corresponding gradient $g_{\text{batch}}$ is defined as the embedding vector obtained by passing the gradient through a trainable feature extractor. Let the feature extractor be denoted as $Grad\_Extractor$, then the privacy token of the gradient $g_{\text{batch}}$ is given by 
\begin{equation}
    Emb(\mathbf{g}_{\text{batch}}) = Grad\_Extractor(\mathbf{g}_{\text{batch}})
\end{equation}
\end{definition}
Mutual information (MI) is a measure of the dependency between two random variables, capturing the amount of information that one variable reveals about the other. For two variables \( \mathcal{X} \) (the private training data) and \( G \) (the corresponding gradients), mutual information \( I(\mathcal{X}; G) \) quantifies the reduction in uncertainty about one variable given knowledge of the other. Formally, the mutual information between \( \mathcal{X} \) and \( G \) is defined as:
\begin{equation}
    I(\mathcal{X}; G) = \mathbb{E}_{\mathcal{X}, G} \left[ \log \frac{p(\mathcal{X}, G)}{p(\mathcal{X}) p(G)} \right]
\end{equation}
In the context of gradient leakage, \( I(\mathcal{X}; G) \) represents the shared information between the training data and gradients, with higher mutual information indicating a greater risk of privacy leakage. Estimating the mutual information \( I(\mathcal{X}; G) \) reveals the extent to which gradients expose information about the original data. This measure is essential for privacy analysis, as it enables researchers to assess the potential for gradient leakage, ultimately guiding the development of privacy-preserving machine learning models.

To estimate the mutual information between the original data and gradients during training of deep learning models, we encounter a significant challenge: the joint distribution of the raw data samples and the gradients lacks a precise analytical form. This complicates the direct computation of mutual information within deep learning frameworks.

From a practical perspective, model trainers may be more interested in determining the minimum amount of private information about the original data that gradients might leak during training. This leads to the problem of estimating a lower bound for the mutual information. To address this, we employ the Donsker-Varadhan representation, which offers a way to formulate a tractable lower bound.
\begin{lemma}[Donsker-Varadhan Representation]\label{lem:DV_representation}
    For two probability distributions \( P \) and \( Q \), their Kullback-Leibler (KL) divergence can be expressed as:
\begin{equation}
    D_{\text{KL}}(P \parallel Q) = \sup_{T} \left( \mathbb{E}_{P} [T] - \log \mathbb{E}_{Q} [e^{T}] \right)
\end{equation}
\end{lemma}
This lemma allows us to estimate the KL divergence between \( P \) and \( Q \) by utilizing a function \( T \), which is often parameterized by a neural network, thus providing a flexible approach for the high-dimensional MI estimation.

Applying this representation to estimate the mutual information between the original data \( \mathcal{X} \) and the gradients \( G \) during the training process, we leverage the Mutual Information Neural Estimator (MINE), a neural network-based approach that approximates mutual information by maximizing a lower bound. MINE trains a neural network, \( T_{\theta} \), parameterized by \( \theta \), to distinguish between joint distributions $p(\mathcal{X}, G)$ and products of marginals $p(\mathcal{X}) p(G)$, providing MI estimation. According to Lemma \ref{lem:DV_representation}, the lower bound for the mutual information \( I(\mathcal{X}; G) \) can be expressed as:
\begin{equation}
    \scriptstyle I(\mathcal{X}; G) \geq \mathbb{E}_{p(\mathcal{X}, G)} [T_{\theta}(\mathcal{X}, G)] - \log \left( \mathbb{E}_{p(\mathcal{X}) p(G)} [e^{T_{\theta}(\mathcal{X}, G)}] \right)
    \label{eq4}
\end{equation}
By maximizing the lower bound, MINE estimates a computationally feasible lower bound of mutual information, which represents the minimum potential privacy risk from gradient leakage during training, even in complex, high-dimensional settings. This approximation lays the groundwork for mutual information estimation in high-dimensional settings, as elaborated in the subsequent sections.

\subsection{MINE Network Based on Gradient Privacy Token}

In the process of estimating mutual information (MI), it is evident that we cannot directly use the original training data and the corresponding gradients to estimate MI between them. MINE network must effectively capture and process the features of both original training data and gradients. 

For the training data, deep learning models inherently function as feature extractors. The model's intermediate layers produce representations of the training data that capture meaningful features. Consequently, we directly leverage the outputs of these intermediate layers as feature representations of the data. This approach bypasses the need for an extra feature extractor for the training data, using the model's internal representations as direct inputs to the MINE network. 

For the gradients, at this point, privacy token plays a key role! We utilize a proper feature extractor, such as an Autoencoder or Transformer, to compress high-dimensional gradients. The Autoencoder or Transformer could be trained to produce compact, representative feature vectors for the gradients, the \textit{privacy token} we defined before. The output embeddings are then used as inputs to the MINE network.  


\begin{figure}[ht]
    \centering
    \vspace{-2mm}
    \subfigure[Designed Based on Autoencoder]{
		\includegraphics[width=0.48\textwidth]{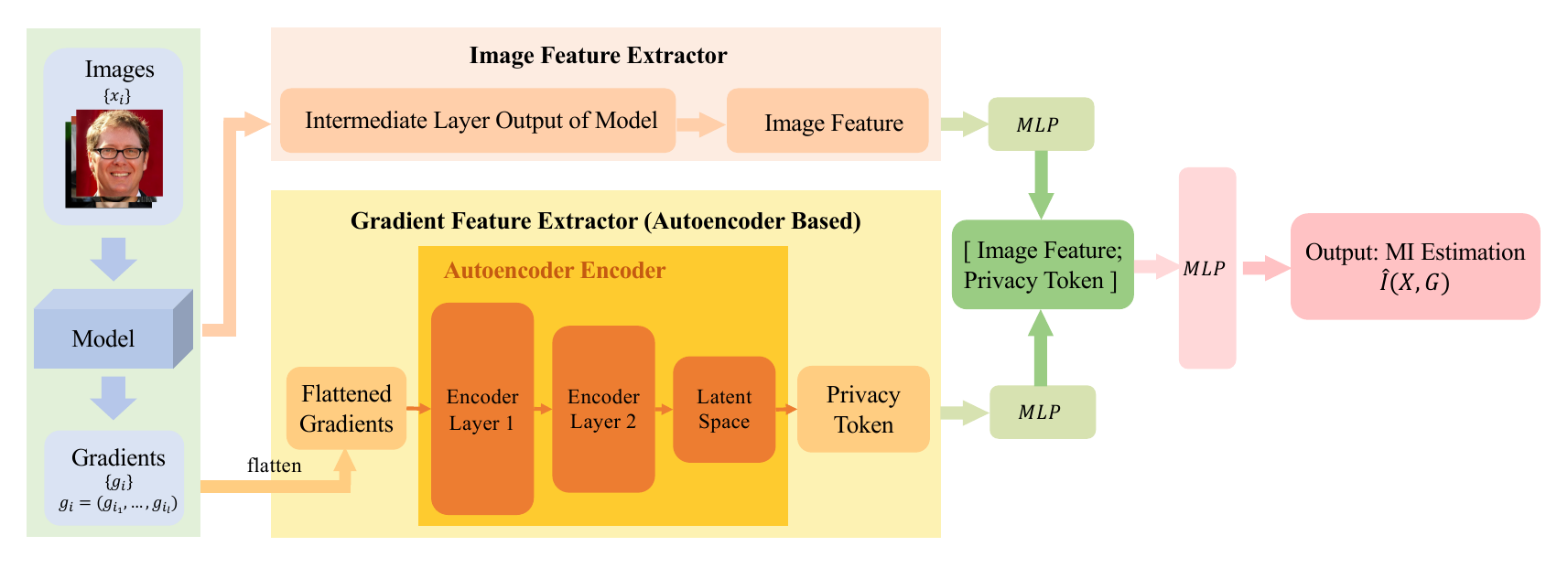}
		\label{mine_auto}
    }
    \subfigure[Designed Based on Transformer]{
		\includegraphics[width=0.48\textwidth]{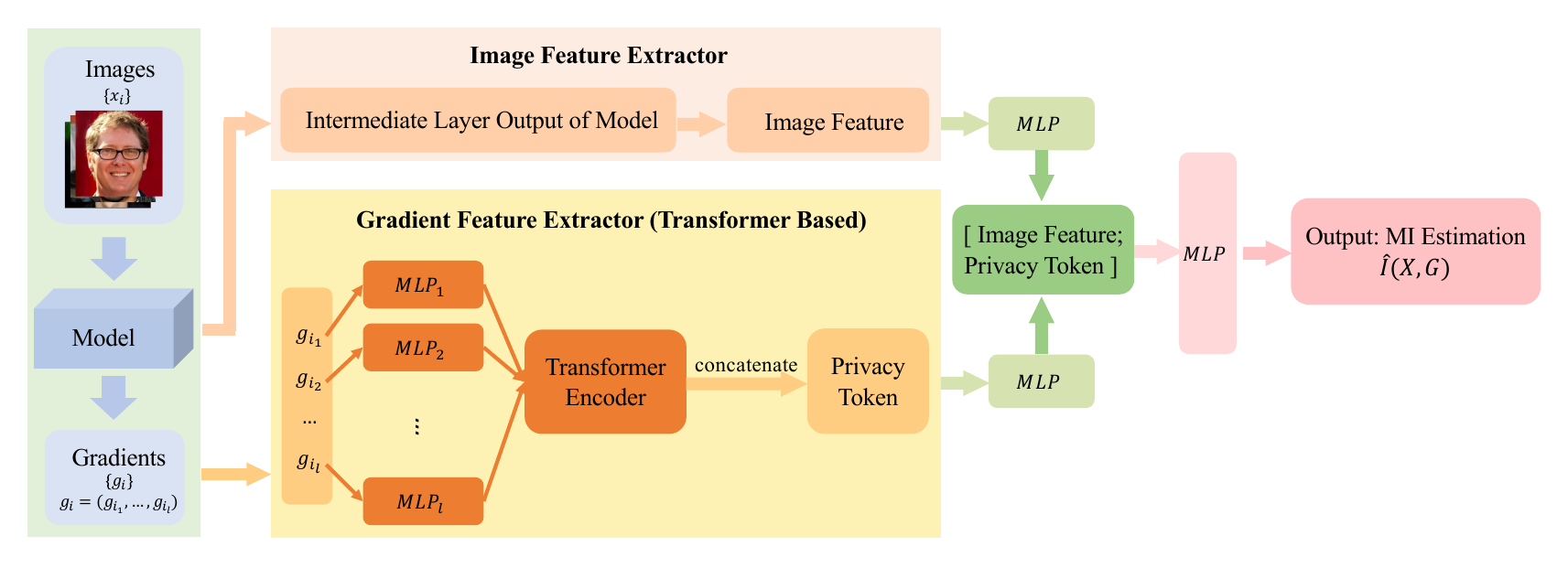}
		\label{mine_trans}
    }
    \caption{MINE Networks.}
    \label{mine_network}
\end{figure}

The overview of MINE network is presented in Figure \ref{mine_network}. This design leverages the model’s intrinsic feature extraction capabilities from the input and the encoding capability of the Autoencoder or Transformer encoder to compress gradient information, forming a compact and effective representation for mutual information estimation. Specifically, the Autoencoder takes flattened gradient vectors as input, with the encoder's latent representation as the privacy token. To match Transformer encoder's required input dimension, we process the gradients from different intermediate layers through distinct Multilayer Perceptron (MLP) networks, obtaining outputs of the same dimension. These outputs are then passed through the Transformer encoder to extract features, which are concatenated to form the privacy token. MINE network enables efficient and accurate estimation of mutual information between training data and gradients, contributing to a detailed analysis of gradient leakage.

\subsection{Mutual Information Estimation (MIE) Algorithm}

To adapt MINE for estimating mutual information between the training data \( \mathcal{X} \) and gradients \( G \), we propose Algorithm \ref{alg: MIE} consisting of three key components:

\textbf{Gradient Collection and Model Training.} During model training, we compute gradients for mini-batches of the training data using stochastic gradient descent (SGD). For each mini-batch, we store the current batch of data and its corresponding gradients in \( \mathcal{D}_{\text{pos}} \). Mismatched data-gradient pairs are created by pairing data batches with randomly permuted gradients from other batches, stored in \( \mathcal{D}_{\text{neg}} \). Specifically, to more precisely monitor changes in mutual information during the training process, we define the "sub-epoch" as the checkpoint unit. Each sub-epoch consists of a fixed number of iterations, rather than using the entire training dataset. This process constructs the data necessary for training the MINE network to distinguish between matched data-gradient pairs and dismatched pairs.

\textbf{MINE Network Training.} Once matched and mismatched samples are collected, they are used to compute the expectations required for MINE loss. Specifically, when privacy tokens are derived from the Autoencoder, MINE loss is augmented with \textit{AE loss}, which measures the discrepancy between the encoder's input and the decoder's output, reflecting the latent space's capacity to represent the input. The MINE network is then trained using the MINE loss.

\begin{algorithm}[H]
    \caption{Mutual Information Estimation}
    \label{alg: MIE}
    \KwIn{Model $F(\mathbf{x}; W)$ with parameters $W$, Dataset $\mathcal{X}$, Number of model training sub-epochs $N_{\text{model}}$, Number of iterations per sub-epoch of model training $N_{\text{sub-epoch}}$, Number of MINE network training iterations $N_{\text{MINE}}$, Learning rates $\alpha$ for model training and $\beta$ for MINE network training.}
    \KwOut{Estimated $\hat{I}(\mathcal{X'}; G')$ between data $\mathcal{X'}=\bigcup_{i=1}^{L} \mathbf{x}'_i$ and gradients $G'=\bigcup_{i=1}^{L} \mathbf{g}'_i$ to be evaluated.}
    Initialize the MINE network $T_{\theta}(\mathbf{x}, \mathbf{g})$. \\
    \textit{// Data Preparation.} \\
    Construct the training data loader from $\mathcal{X}$. \\
    \textit{// Model Training and Data-Gradient Pair Collection.} \\
    Initialize positive sample list $\mathcal{D}_{\text{pos}} = \emptyset$.\\
    Initialize negative sample list $\mathcal{D}_{\text{neg}} = \emptyset$.\\
    \textit{// Model Training (Mini-batch SGD).} \\
    \For{$t = 1, 2, \dots, N_{\text{model}}$}{
        \For{$i = 1, 2, \dots, N_{\text{sub-epoch}}$}{
            Compute the gradients of the $i$-th batch $\mathbf{x}_{\text{batch}}^{(i)}$: $\mathbf{g}_{\text{batch}}^{(i)} = \nabla_W F(\mathbf{x}_{\text{batch}}^{(i)}; W)$.\\ 
            Update the model parameters using the optimizer: $W \leftarrow W - \alpha \cdot \mathbf{g}_{\text{batch}}^{(i)}$.\\
            \textit{// Collect Matched Data-Gradient Pair.} \\
            Store the current batch data and its corresponding gradient $(\mathbf{x}_{\text{batch}}^{(i)}, \mathbf{g}_{\text{batch}}^{(i)})$ in $\mathcal{D}_{\text{pos}}$. 
            }
         \textit{// Construct Mismatched Data-Gradient Pair.} \\
         Shuffle gradients $g_{\text{batch}}$ across matched data-gradient pairs to form mismatched pairs: $(\mathbf{x}_{\text{batch}}^{(i)}, \mathbf{g}_{\text{batch}}^{(j)})$, where $i = 1,2,..., N_{\text{sub-epoch}}$ and $i \neq j$.\\
         Store mismatched pairs in $\mathcal{D}_{\text{neg}}$.
         }
    \textit{// MINE Network Training.} \\
    \For{$t = 1, 2, \dots, N_{\text{MINE}}$}{
         Sample a batch of samples \( \{(\mathbf{x}_m^{+}, \mathbf{g}_m^{+})\}_{m=1}^{M} \) from \( \mathcal{D}_{\text{pos}} \).\\ 
         Sample a batch of samples $\{(\mathbf{x}_n^{-}, \mathbf{g}_n^{-})\}_{n=1}^{N}$ from $\mathcal{D}_{\text{neg}}$.\\
         Estimate positive expectation:\\ $\hat{\mathbb{E}}_{\text{pos}} = \frac{1}{M} \sum_{m=1}^M \frac{1}{|\mathbf{x}_{m}^{+}|} \sum_{i=1}^{|\mathbf{x}_{m}^{+}|} [T_{\theta}(\mathbf{x}_{m}^{+}, \mathbf{g}_{m}^{+})]_i$ \\
         Estimate negative expectation (log domain):\\ 
         $\hat{\mathbb{E}}_{\text{neg}} = \frac{1}{N} \sum_{n=1}^N \frac{1}{|\mathbf{x}_{n}^{-}|} \sum_{i=1}^{|\mathbf{x}_{n}^{-}|} [e^{T_{\theta}(\mathbf{x}_{n}^{-}, \mathbf{g}_{n}^{-})}]_i$\\
         Compute MINE loss: $\mathcal{L}(\theta) = -\hat{\mathbb{E}}_{\text{pos}} + \log\left( \hat{\mathbb{E}}_{\text{neg}} \right)$ \\
         \If{Gradients feature extractor is Autoencoder-based}{
            $\mathcal{L}(\theta) \gets \mathcal{L}(\theta) + \mathcal{L}_{\text{AE loss}}(\theta)$
         } 
         Update MINE network parameters: $\theta \leftarrow \theta - \beta \nabla_{\theta} \mathcal{L}(\theta)$ 
    }
    \textit{// Mutual Information Estimation.} \\
    Compute the positive expectation for $\mathcal{X'}$ and $G'$:\\
    $\hat{\mathbb{E}}_{\text{pos}}' = \frac{1}{L} \sum_{(\mathbf{x}', \mathbf{g}') \in (\mathcal{X'}, G')} \frac{1}{|\mathbf{x}'|} \sum_{i=1}^{|\mathbf{x}'|} [T_{\theta}(\mathbf{x}', \mathbf{g}')]_i$. \\
    Shuffle gradients $\mathbf{g}'$ of the $G'$ to obtain $G_{\text{shuffled}}'$.\\
    Compute the negative expectation for $\mathcal{X'}$ and $G'$ :\\
    $\hat{\mathbb{E}}_{\text{neg}}' = \frac{1}{L} \sum_{(\mathbf{x}', \mathbf{g}_{\text{shuffled}}') \in (\mathcal{X'}, G_{\text{shuffled}}')} \frac{1}{|\mathbf{x}'|} \sum_{i=1}^{|\mathbf{x}'|} [e^{T_{\theta}(\mathbf{x}', \mathbf{g}_{\text{shuffled}}')}]_i$.\\
    Estimate MI:
    $\hat{I}(\mathcal{X'}; G') = \hat{\mathbb{E}}_{\text{pos}}' - \log\left( \hat{\mathbb{E}}_{\text{neg}}' \right)$.\\
    \Return the estimated mutual information $\hat{I}(\mathcal{X'}; G')$.
\end{algorithm}

\textbf{Mutual Information Estimation.} Using the trained MINE network, we first compute the positive expectation with the given images and gradients to be evaluated, then shuffle the gradients to calculate the negative expectation. The mutual information is estimated based on the previously described lower bound, providing an estimate of \( I(\mathcal{X'}; G') \), which indicates the extent of gradient leakage.

\section{Experiments}\label{experiment_section}

\subsection{Experiment Setup}\label{setup_of_exp}

\paragraph{Datasets and Models.} We use CIFAR-10 ($64 \times 64$ pixels) and CelebA-HQ ($256 \times 256$ pixels) as our experimental dataset. Experiments are conducted using LeNet and AlexNet for CIFAR-10, and CNN and MLP for CelebA-HQ. Their architectures are as follows:

\textit{LeNet (CIFAR-10):} The LeNet consists of three convolutional layers with ReLU activations, followed by a fully connected layer. The output of the convolutional layers is flattened into a 1D vector of length 768, serving as the feature representation of the input image, which is then passed through the fully connected layer. We quantify the mutual information between the gradients of the three convolutional layers and the input images.

\textit{AlexNet (CIFAR-10):} The AlexNet comprises five convolutional layers with ReLU activations and max-pooling, followed by an adaptive average pooling layer and three fully connected layers for classification. The output of the adaptive average pooling layer serves as the feature representation. Given the large size of gradient vectors and insights from \cite{huang2024intermediate} highlighting the importance of gradients closer to the output for classification, we measure the mutual information between the gradients of the last three convolutional layers and the input images.

\textit{CNN (CelebA-HQ):} The CNN consists of two convolutional layers with ReLU activations, followed by max pooling for downsampling and two fully connected layers for embedding and classification. The output of the first fully connected layer serves as the feature embedding for the input data. And we quantify the mutual information between the gradients of the convolutional layers and the input images.

\textit{MLP (CelebA-HQ):} The MLP is a Multilayer Perceptron network with four fully connected layers, each followed by ReLU activations, except the output layer, which predicts classifications. The output of the second layer serves as the embedding of the input. We quantify the mutual information between the gradients of all layers and the input images.
\paragraph{Image Reconstruction Attack Setup.} For the CIFAR-10 ($64 \times 64$ pixels) dataset, we apply the DLG method introduced in \cite{zhu2019deep} for gradient-based image reconstruction attacks. DLG iteratively optimizes a randomly initialized input image to minimize the difference between its gradient in the target model and the leaked gradient. For the CelebA-HQ ($256 \times 256$ pixels) dataset, we adopt the DSG-based method proposed in \cite{huang2024gradient}, which utilizes a gradient-guided conditional diffusion model to reconstruct images and performs better for reconstructing high-resolution images.

Other detailed settings are presented in Appendix \ref{settings}.

\subsection{Experiment Result} \label{sec:exp}

\subsubsection{Trends of Mutual Information} 
As shown in Algorithm \ref{alg: MIE}, during the training process of the monitored model, the trainer continuously collects data-gradient pairs to train the MINE network. Subsequently, the mutual information of the data-gradient pairs to be evaluated, such as those collected pairs up to the current stage, can be estimated. In our experiments, we save model checkpoints at various sub-epochs during training. From these checkpoints, we extract 200 batches of data from the training set to ensure the reliability and accuracy of the estimates, and use the models from different sub-epochs to calculate the corresponding matched data-gradient pairs. After shuffling the gradients across batches, we generated mismatched data-gradient pairs. Using the trained MINE network, we estimate the mutual information $I_{\{a,t\}}(\mathcal{X};G)$ and $I_{\{a,t\}}(\mathcal{X};G')$ for matched and mismatched pairs for each batch across different sub-epochs, where $a$ and $t$ denote the Autoencoder-based and Transformer-based gradient extractors using to derive privacy tokens, respectively. The difference, $I_{\{a,t\}}(\mathcal{X};G)- I_{\{a,t\}}(\mathcal{X};G^{\prime})$, is then calculated and averaged over 200 batches per sub-epoch. These differences indicate the difficulty of distinguishing between matched and mismatched pairs, reflecting the correlation between the gradients and the original training data. Additionally, we document the test accuracy of the model before each sub-epoch. The results are presented in the first five columns of Table \ref{attack}, with additional details provided in Appendix \ref{MI-result}. 

Based on these results, it can be observed that: 

$\bullet$ First, mutual information differences are consistently greater than zero, suggesting that matched data-gradient pairs have higher mutual information than mismatched pairs, except when the model is randomly initialized and untrained. In this case, randomly initialized parameters make distinguishing between matched and mismatched pairs more challenging. Once the attackers gain access to the gradients, they can reduce uncertainty in inferring private data. 

$\bullet$ Second, the variation in the mutual information difference follows a distinct pattern: it initially increases and then decreases as training progresses. This is expected, as the model progressively learns to capture input features and adapts to the downstream task. Before training, with random initialization, outputs are highly uncertain and may vary significantly for similar inputs. As gradients are closely tied to the outputs, distinguishing between matched and mismatched data-gradient pairs is challenging, resulting in the lowest mutual information difference early in training. As training progresses, the model learns general features from the data, and the parameters are gradually optimized. The outputs become more closely related to the inputs, strengthening the correlation between matched images and gradients, causing the mutual information difference to increase. In the later stages, the feature extraction process stabilizes, and the outputs of the feature extraction layers are constrained within specific ranges. Consequently, gradients for training data records tend to fall within a similar range, leading to a decrease in the mutual information difference. This suggests that gradients are more sensitive in the intermediate stages and less sensitive in the later stages.

$\bullet$ Third, although the mutual information differences calculated with different gradient feature extractors vary numerically, their trends throughout training are consistently aligned, indicating that all effective gradient embeddings can reliably compute mutual information. The Transformer-based extractor appears to be more meaningful in practical terms, albeit at the cost of higher computational complexity. Therefore, when selecting a feature extractor, it is essential to strike a balance between computational efficiency and the accuracy of mutual information estimation.

\begin{table*}[htp]
\caption{Mutual Information and Performance of Privacy Attack. The maximum mutual information difference across checkpoints is marked in bold. Peak attack performance points are bolded, and the worst performance is both bolded and italicized.}
\resizebox{2\columnwidth}{!}{%
\begin{tabular}{c|c|c|cc|cccc}
\toprule[1pt]
\textbf{Model} & \textbf{Sub-Epoch} & \textbf{Test Accuracy} & \textbf{$I_a(\mathcal{X};G) - I_a(\mathcal{X};G^{\prime})$} & \textbf{$I_t(\mathcal{X};G) - I_t(\mathcal{X};G^{\prime})$} & \textbf{SSIM $\uparrow$}        & \textbf{MSE $\downarrow$}             & \textbf{PSNR $\uparrow$}         & \textbf{LPIPS $\downarrow$}         \\ \toprule[1pt]

\multirow{7}{*}{\textbf{LeNet}} & 0         & 10.39\%       & 0.03288       & 0.01588      & 0.999039710 & \textbf{0.030865720} & \textbf{15.1052361 dB} & 2.7979574e-06 \\
&1         & 17.93\%       & 0.98077       & 1.56315       & \textbf{0.999248564} & 0.039145544 & 14.0731764 dB & \textbf{2.0446041e-06} \\
&5         & 26.13\%       & 2.18896       & 2.70519       & 0.998636901 & 0.059506238 & 12.2543755 dB & 1.0335207e-05 \\ 
&8         & 29.70\%       & 2.59941       & 3.09143       & \textbf{0.999070644} & \textbf{0.042847046} & \textbf{13.7697935 dB} & 5.4185015e-06 \\ 
&60        & 47.45\%       & \textbf{4.15026}       & \textbf{4.30479}       & 0.998919308 & 0.057304602 & 12.4181042 dB & \textbf{3.7715349e-06} \\ 
&100       & 51.24\%       & 3.88313       & 3.97720       & 0.998631895 & 0.075161338 & 11.2400541 dB & 6.3348662e-06 \\ 
&200       & 57.06\%       & 3.30983       & 3.55665       & \textit{\textbf{0.994546950}} & \textbf{\textit{0.153886259}} & \textbf{\textit{8.1280022 dB}} & \textbf{\textit{2.2399030e-05}} \\ \toprule[1pt]

\multirow{7}{*}{\textbf{AlexNet}} & 0         & 10.00\%       & -0.00109       & -0.00197      & \textbf{0.997909069} & \textbf{0.026695648} & \textbf{15.7355957 dB} & \textbf{4.4740254e-06} \\ 
&5         & 21.45\%       & 2.81713       & 2.84957       & 0.994904280 & 0.058183897 & 12.3519716 dB & 8.1643147e-06 \\ 
&15        & 30.55\%       & \textbf{3.01831}       & \textbf{2.97529}       & 0.993920386 & 0.102552183 & 9.8905506 dB & \textbf{7.1724528e-06} \\ 
&30        & 36.71\%       & 2.90797       & 2.64001       & \textbf{0.996570732} & \textbf{0.054689482} & \textbf{10.9417013 dB} & 1.5824698e-05 \\ 
&50        & 43.51\%       & 2.73608       & 2.31924       & 0.994314969 & 0.135054663 & 8.6949043 dB & 2.0395855e-05 \\
&120       & 53.21\%       & 1.85140       & 1.40926       & 0.992240727 & 0.132450566 & 8.7794619 dB & 2.6198821e-05 \\ 
&200       & 57.42\%       & 1.42161       & 1.15953       & \textbf{\textit{0.993263040}} & \textbf{\textit{0.138820484}} & \textbf{\textit{8.5754642 dB}} & \textbf{\textit{1.5810241e-05}} \\\toprule[1pt]

\multirow{7}{*}{\textbf{CNN}} & 0         & 64.05\%       & 0.00033       & -0.00633      & \textbf{0.999986172} & \textbf{0.000584106} & \textbf{32.3350868 dB} & \textbf{1.3302498e-06} \\ 
&1         & 68.85\%       & 0.27901       & 1.20039       & 0.999980211 & 0.000718946 & 31.4330406 dB & 2.2014940e-06 \\
&5         & 73.23\%       & 0.67145       & 3.04729       & 0.999900758 & 0.000981016 & 27.4228268 dB & 5.3508622e-06 \\ 
&10        & 75.94\%       & 1.29205       & 4.44203       & 0.999701738 & 0.003754674 & \textbf{29.2542801 dB} & 1.9714682e-05 \\ 
&20        & 86.46\%       & 1.66154       & \textbf{6.18565}       & 0.999265015 & 0.008026601 & 20.9546833 dB & 5.0416951e-05 \\ 
&50        & 90.42\%       & \textbf{1.69969}       & 5.76775       & 0.998914242 & 0.011613548 & 19.3503513 dB & 8.2043822e-05 \\ 
&100       & 94.17\%       & 1.39744       & 5.63270       & \textbf{\textit{0.997994052}} & \textbf{\textit{0.011776168}} & \textbf{\textit{19.2899590 dB}} & \textbf{\textit{8.5325912e-05}} \\ \toprule[1pt]

\multirow{7}{*}{\textbf{MLP}} & 0    & 36.15\%       & 5.83049e-04    & 0.0013300     & 0.999995708 & \textbf{5.223742846e-05} & \textbf{42.8201828 dB} & \textbf{2.2593352e-07} \\
&1         & 67.81\%       & 0.67829       & 0.41561       & \textbf{0.999997795} & 7.14464113e-05  & 41.4601936 dB & 3.8103249e-07 \\ 
&5         & 76.15\%       & 1.80015       & 1.39898       & 0.999986649 & 0.000239780     & 34.6880188 dB & 4.0843397e-06 \\ 
&10        & 79.27\%       & \textbf{2.27453}       & 1.78570       & 0.999967337 & 0.000623932     & 32.0486298 dB &  6.0732955e-06 \\ 
&20        & 87.08\%       & 2.0571        & \textbf{1.90173}       & 0.999921143 & 0.001258478     & 29.0015430 dB & 1.6891203e-05 \\ 
&50        & 91.46\%       & 1.37297       & 1.72956       & 0.997579276 & 0.021300912     & 16.7160187 dB & 2.9218782e-04  \\ 
&100       & 92.81\%       & 1.2793        & 1.70819       & \textbf{\textit{0.997521698}} & \textbf{\textit{0.021320218}}     & \textbf{\textit{16.7120838 dB}} & \textbf{\textit{3.0333982e-04 }} \\ \toprule[1pt]
\end{tabular}}
\label{attack}
\end{table*}

\subsubsection{Mutual Information and Capability of Privacy Attacks}

Mutual information quantifies the dependence between the training data and the gradients. Therefore, the mutual information difference between matched and mismatched data-gradient pairs reflects the difficulty of distinguishing them and indicates the attacker's ability to execute a successful attack. 

\begin{figure}[ht]
    \centering
    \vspace{-2mm}
    \includegraphics[width=0.48\textwidth]{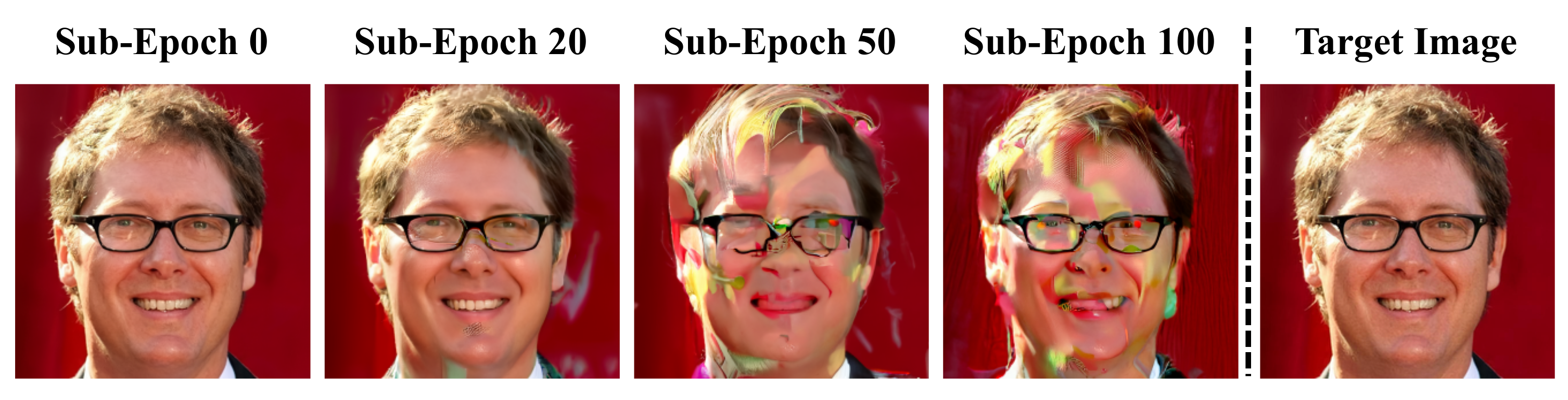}
    \caption{Attack Results on CelebA-HQ (CNN).}
    \label{Attacks-cnn}
\end{figure}

\begin{figure}[ht]
    \centering
    \vspace{-2mm}
    \includegraphics[width=0.48\textwidth]{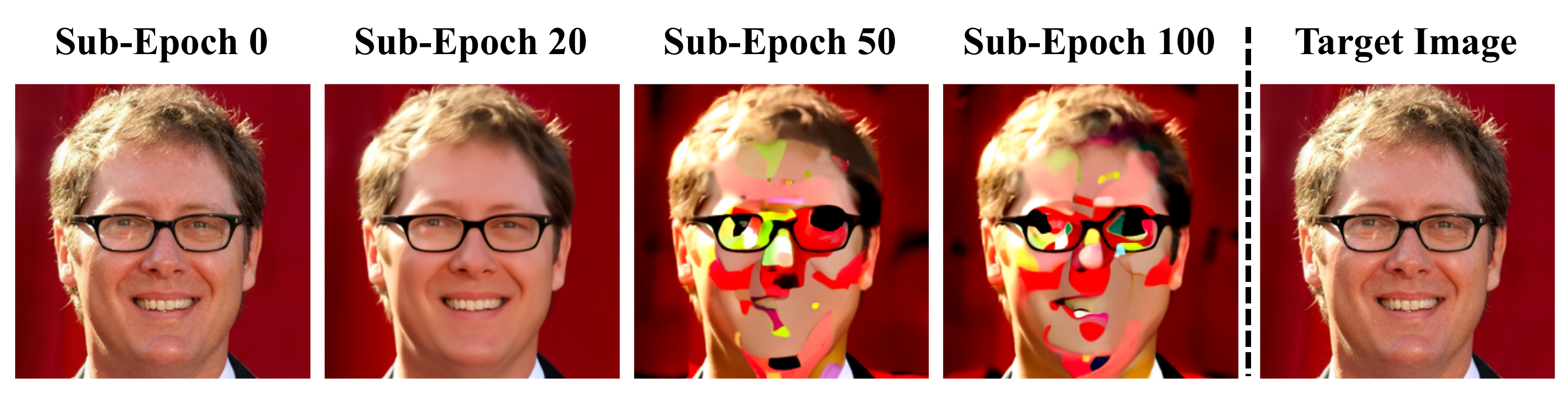}
    \caption{Attack Results on CelebA-HQ (MLP).}
    \label{Attacks-mlp}
\end{figure}

To assess the relationship between the mutual information difference and the attacker's capabilities, we use the gradient-based image reconstruction attack, which enables the model to reconstruct original image samples from gradients. In our experiment, we randomly select a data sample from the training set as the target image, compute its gradients across multiple sub-epochs, and feed these target gradients into the attack model to reconstruct the target image. The reconstructed images then compared to the target image using relevant metrics. The results are shown in the last four columns of Table \ref{attack}, with examples of the reconstructed data provided in Figures \ref{Attacks-cnn} and \ref{Attacks-mlp}. 

From the results, we can draw the following conclusions: 

$\bullet$ First, as the mutual information difference decreases, the quality of the reconstructed data deteriorates, as reflected by the metrics. This suggests that the mutual information difference computed by our method is strongly correlated with attack performance. This correlation is logical, as the mutual information difference reflects the difficulty of distinguishing between matched and mismatched data-gradient pairs. As the difference decreases, distinguishing between these pairs becomes increasingly challenging, resulting in poorer reconstruction attack performance. This implies that, in the later sub-epochs, even if the attacker has access to gradient information, they are unlikely to obtain the desired outcome in the privacy attacks. Decreasing mutual information difference progressively restricts the attacker’s information gain, even with the leaked gradients.

$\bullet$ Second, we observe that gradients in the intermediate stages of model training exhibit greater sensitivity, as indicated by both the mutual information differences and attack performance metrics. Interestingly, we also find that gradients in the later stages of training are less sensitive, resulting in poorer attack performance when the attacker captures gradients from this phase. In the context of differential privacy, this suggests that gradients in later stages are more vulnerable to privacy-preserving noise. Our findings indicate that, during the later training stages, adding minimal or no noise is crucial for preserving the utility of the gradients.

$\bullet$ Third, unexpectedly, the reconstructed results exhibit remarkable accuracy during the initial sub-epochs of model training, which contradicts the fact that the mutual information difference is in an increasing phase at this stage. This phenomenon can likely be attributed to the effects of model random initialization. Early-stage gradient-guided reconstruction attacks benefit from the random small weights that break parameter symmetry, making gradient information more distinguishable. In this phase, the model is in a high-loss region with more sensitive gradients, which attackers can exploit more easily. Additionally, simpler activation patterns and a more intuitive loss surface facilitate smoother and higher-quality reconstructions.

$\bullet$ Finally, a notable finding emerges from the earlier learning stages, particularly during sub-epochs when the mutual information difference increases, where we observe relatively high-quality reconstructions (as Figures \ref{Attacks-cnn} and \ref{Attacks-mlp}). In Table \ref{attack}, we note that the mutual information difference starts to decrease around sub-epoch 20 for CNN and MLP models (although it may also decrease at the sub-epochs in the vicinity of 20, such as 10 and 50). Interestingly, the reconstructed images at these sub-epochs are nearly identical to the target image, indicating that the gradients in these sub-epochs, where mutual information increases, are more sensitive. Thus, we conclude that the most critical focus for model parameter protection should be on the gradients during the early stages of training.

\textbf{Summary:} Our method demonstrates a strong correlation between the mutual information difference of matched and mismatched data-gradient pairs and the effectiveness of privacy attacks. The mutual information, derived from privacy tokens, provides a reliable measure of attack potential without the need for actual privacy attacks. Furthermore, existing differential privacy protection strategies typically focus on model performance, advocating for higher noise injection in the early training stages and reduced noise in later stages. Our findings reinforce this approach from the perspective of privacy leakage, as gradients in the early and intermediate training stages are more sensitive, whereas those in later stages exhibit reduced sensitivity. Thus, our work highlights the importance of adjusting noise levels based on gradient sensitivities to enhance privacy protection.

Ablation experiments are presented in Appendix \ref{ablation_study}.

\section{Conclusion and Future Work}

In this paper, we introduce the concept of \textit{privacy tokens} to proactively assess privacy risks in deep learning models. By embedding gradients into privacy tokens and quantifying their relationship with training data using Mutual Information (MI) via the MINE network, our approach enables real-time privacy monitoring, overcoming the limitations of traditional post-training attack simulations and advancing privacy-aware deep learning. This work advances privacy-aware deep learning, providing a more precise and adaptive framework for evaluating privacy risks during training.

Looking ahead, we will enhance the robustness of privacy tokens through improved gradient embedding techniques to ensure more accurate and resilient privacy risk assessments. Additionally, We will integrate differential privacy (DP) with MI-based privacy quantification, enabling adaptive privacy budget allocation by dynamically adjusting noise based on quantified privacy leakage at different training stages to optimize the balance between privacy protection and model utility.

\bibliography{example_paper}
\bibliographystyle{icml2025}

\newpage
\appendix
\onecolumn
\section{Detailed Experimental Settings}\label{settings}

\paragraph{Hyperparameters.}
During model training, for the CIFAR-10 dataset, $N_{\text{Model}}$ is set to 200, with each sub-epoch consisting of 50 iterations ($N_{sub-epoch}$ equals 50) and a batch size of 20. The Adam optimizer is employed with an initial learning rate of $9 \times 10^{-4}$, and a StepLR scheduler reduces the learning rate by a factor of 0.9 every 20 sub-epochs. For the CelebA-HQ dataset, due to the model’s rapid convergence, $N_{\text{Model}}$ is set to 100, with each sub-epoch consisting of 10 iterations ($N_{sub-epoch}$ equals 10) and a batch size of 20. The initial learning rate of the Adam optimizer is set to $1 \times 10^{-4}$, which is reduced by a factor of 0.9 every 10 sub-epochs using the StepLR scheduler.

\paragraph{Settings of MINE Network.}
For the original data, we utilize the outputs from the intermediate layers of the monitored model as its features, as detailed in Section \ref{setup_of_exp}. For the gradients, two feature extraction methods are implemented: an Autoencoder-based approach and a Transformer-based approach for generating privacy tokens. The Autoencoder aims to learn a compact, low-dimensional representation of the input data. It consists of two primary components: encoder and decoder. The encoder compresses the input data from its original dimensionality into a lower-dimensional latent space using two fully connected layers with LeakyReLU activations. The decoder then reconstructs the input data from this lower-dimensional representation back to the original dimensionality through two fully connected layers, with the final layer applying a ReLU activation to ensure non-negative outputs. The Transformer-based approach incorporates multiple single-layer MLP networks, each mapping flattened gradients from different layers into 512-dimensional vectors. These vectors are subsequently processed by a single Transformer encoder, which employs multi-head self-attention (with 8 heads), a 2048-dimensional feedforward network, and a dropout rate of 0.1. The output of the Transformer encoder is then concatenated into a 1D vector to represent the gradient features. The features extracted from both images and gradients are independently embedded into 128-dimensional vectors using two fully connected layers. These vectors are concatenated to form a 256-dimensional image-gradient composite vector, which is further processed through two additional fully connected layers to produce a scalar output. 

During the training of the MINE network, $N_{\text{MINE}}$ is set to 200. The batch sizes for calculating the positive and negative expectations ($M$ and $N$ in Algorithm \ref{alg: MIE}) are set to 10\% of the sample sizes in $\mathcal{D}{\text{pos}}$ and $\mathcal{D}{\text{neg}}$, respectively. For the Autoencoder-based MINE networks, the Autoencoder component is optimized using the Adam optimizer with a learning rate of $1 \times 10^{-5}$, which is applied to the parameters of the Autoencoder, while the other MINE components use a higher learning rate of $1 \times 10^{-4}$. For the Transformer-based MINE networks, the MLP networks, which standardize the dimensions of gradients across different intermediate layers, and the Transformer encoder are optimized with the Adam optimizer at a learning rate $1 \times 10^{-5}$, and the remaining MINE components are optimized with a higher learning rate of $1 \times 10^{-4}$.

\paragraph{Machine Configuration.} 
All experiments are run over a GPU machine with one Intel(R) Xeon(R) Gold 5218R CPU @ 2.10GHz with 251 GB DRAM and 4 NVIDIA RTX A6000 GPUs (each GPU has 48 GB DRAM). The operating system is Linux. We implement our experiment with Pytorch 2.0.1.

\section{Mutual Information Estimation Results}\label{MI-result}
\begin{figure}[htp]
    \centering
    \vspace{-2mm}
    \subfigure[Sub-Epoch 0 (Autoencoder)]{
		\includegraphics[width=0.34\textwidth]{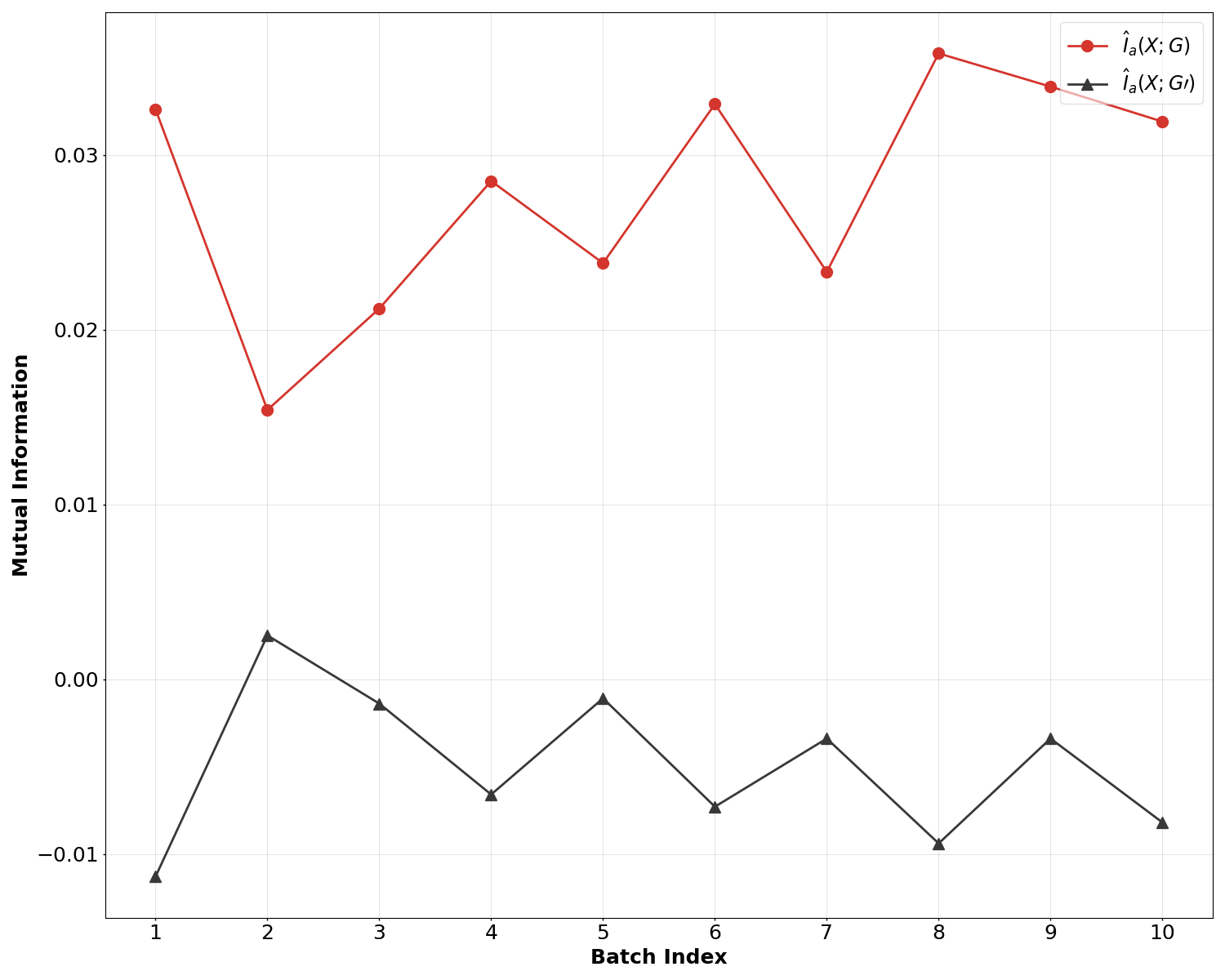}
		\label{let-a-0}
    }
    \subfigure[Sub-Epoch 8 (Autoencoder)]{
		\includegraphics[width=0.34\textwidth]{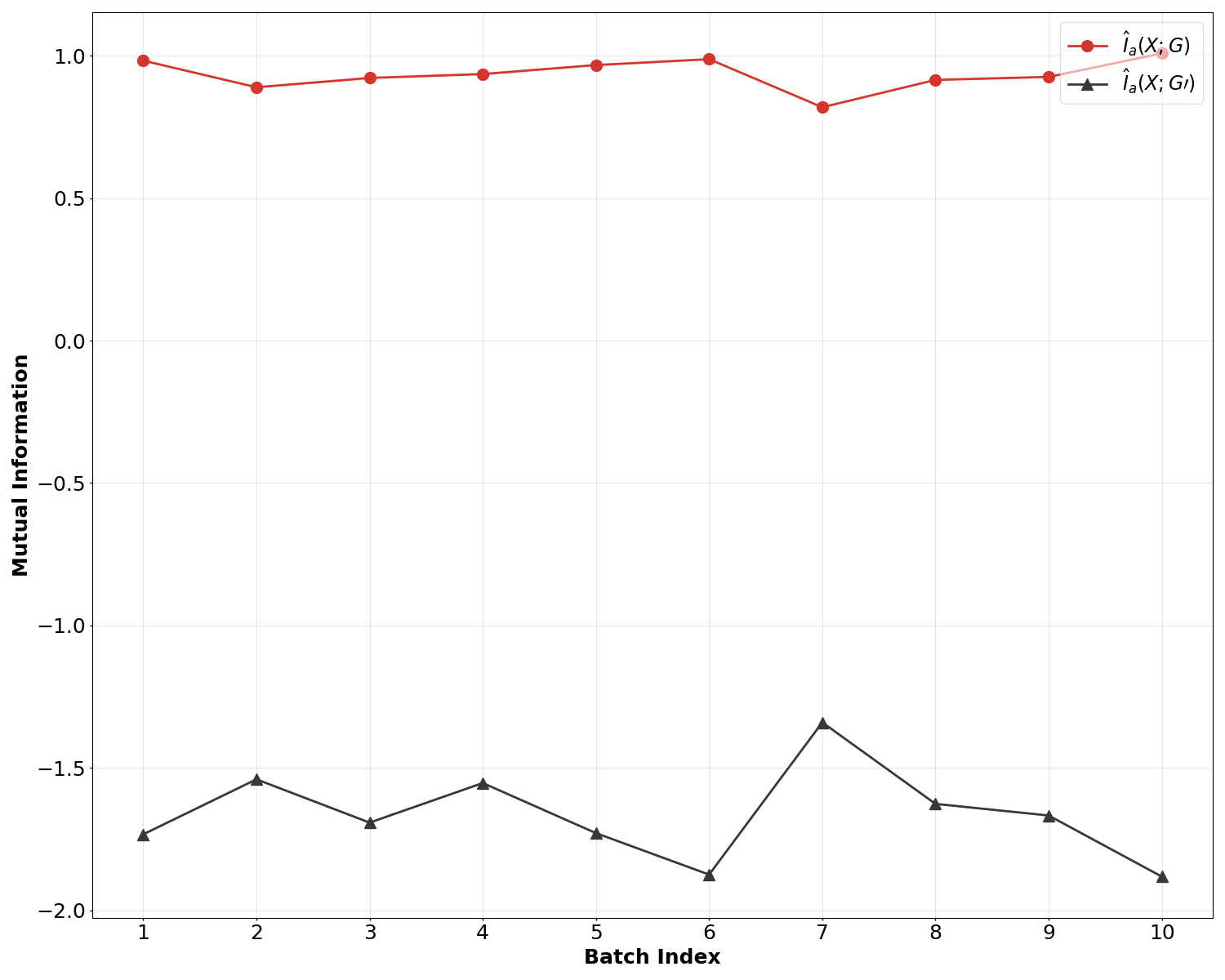}
		\label{let-a-8}
    }\\
    \subfigure[Sub-Epoch 60 (Autoencoder)]{
		\includegraphics[width=0.34\textwidth]{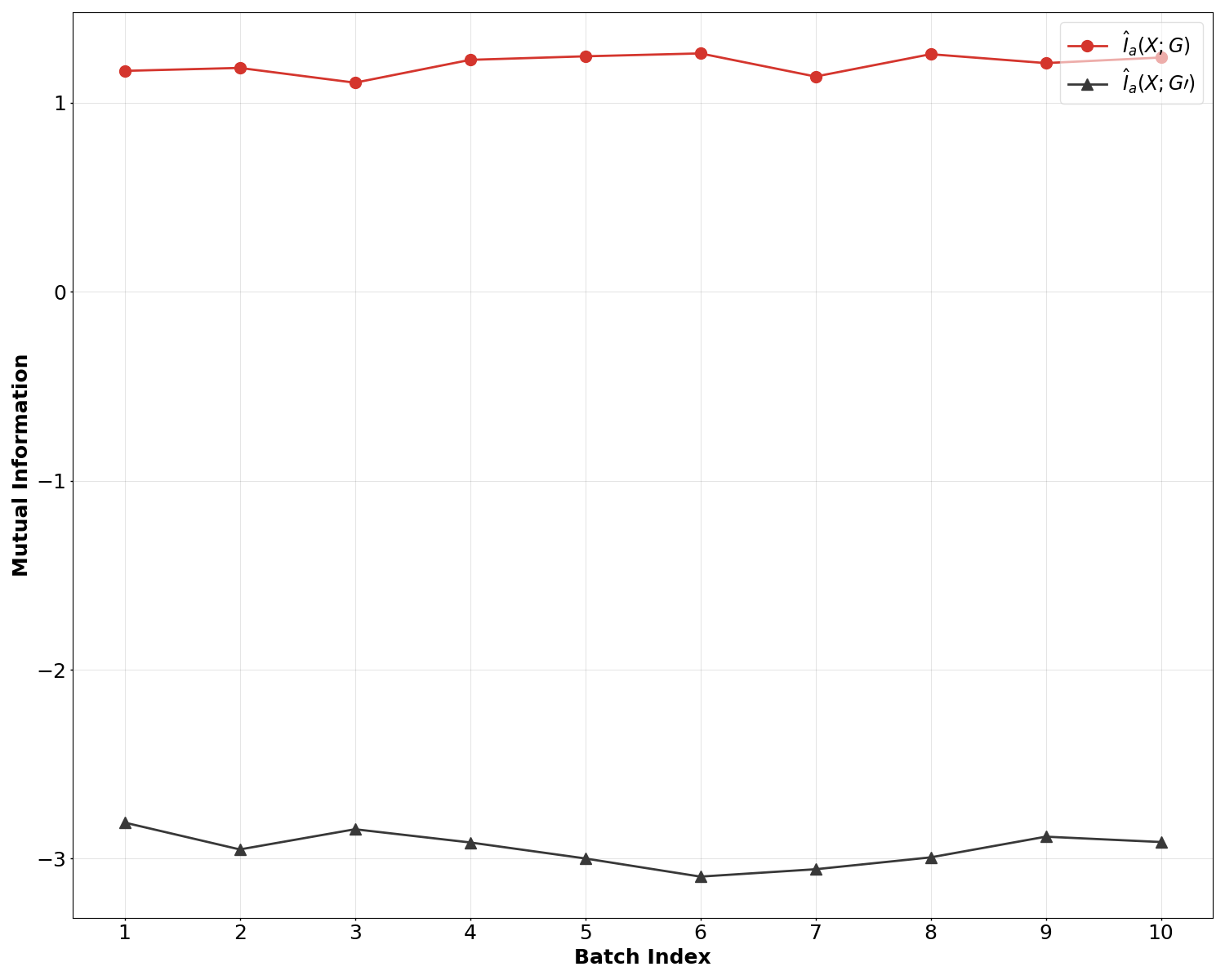}
		\label{let-a-60}
    }
    \subfigure[Sub-Epoch 200 (Autoencoder)]{
		\includegraphics[width=0.34\textwidth]{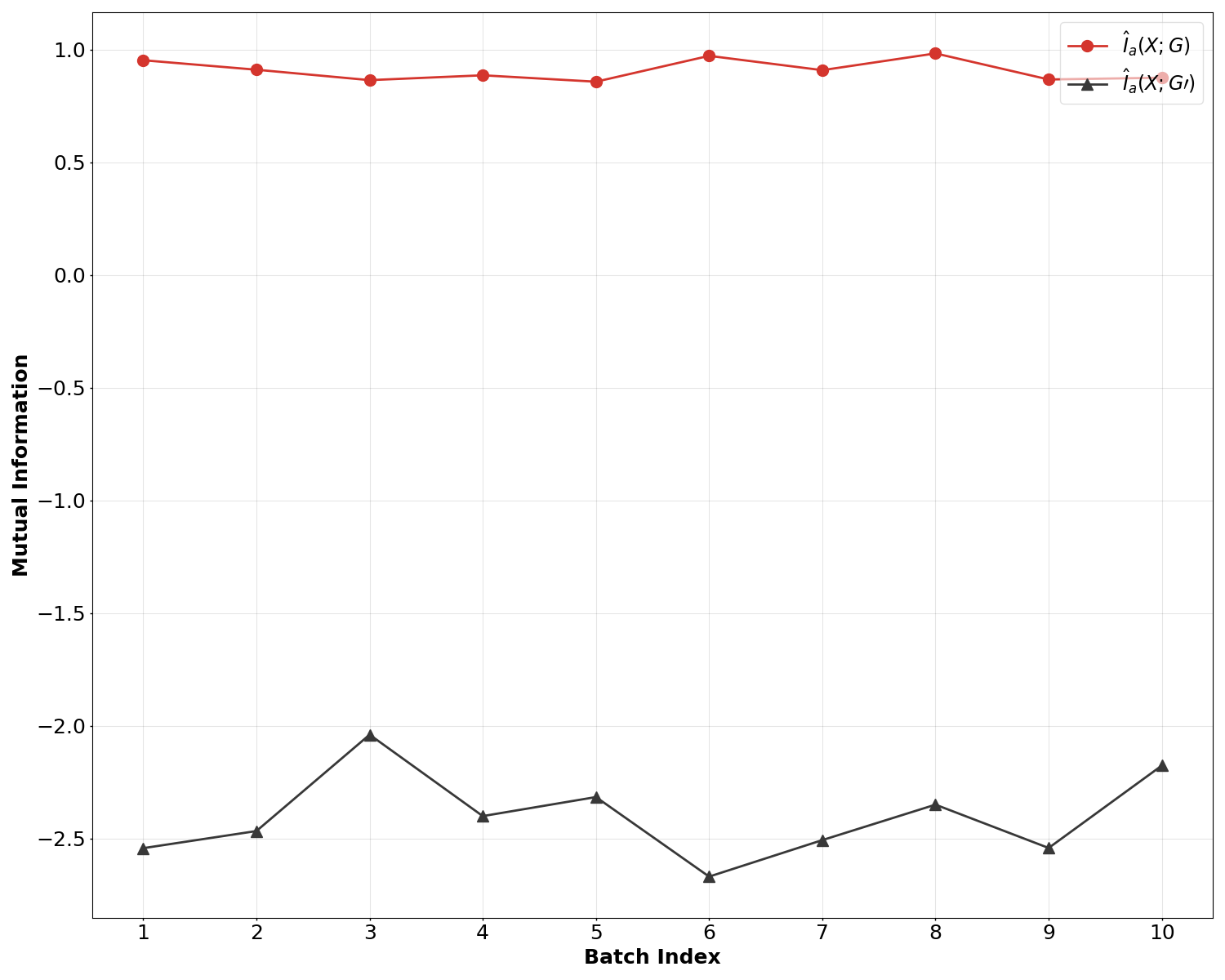}
		\label{let-a-200}
    }\\
    \subfigure[Sub-Epoch 0 (Transformer)]{
		\includegraphics[width=0.34\textwidth]{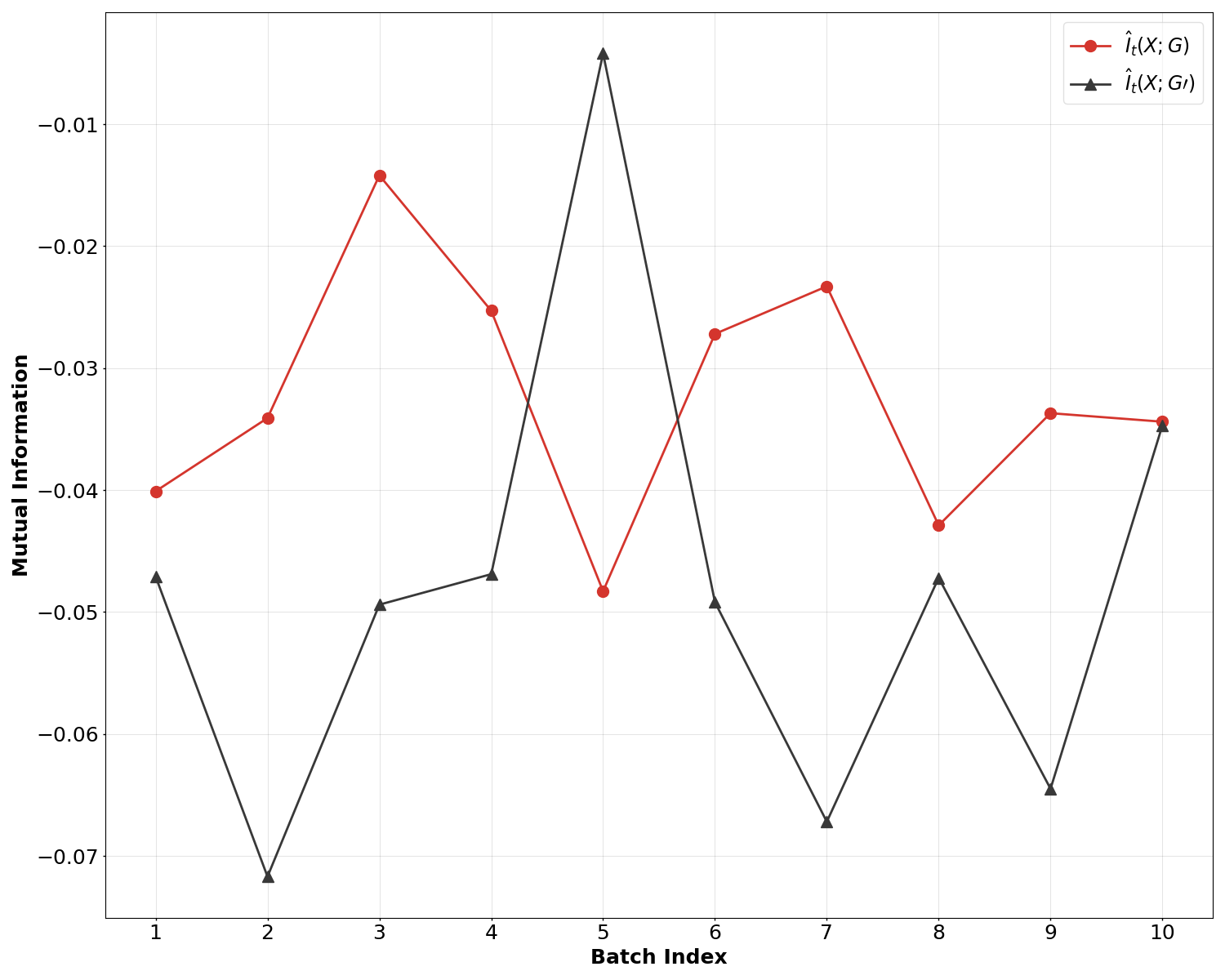}
		\label{let-t-0}
    }
    \subfigure[Sub-Epoch 8 (Transformer)]{
		\includegraphics[width=0.34\textwidth]{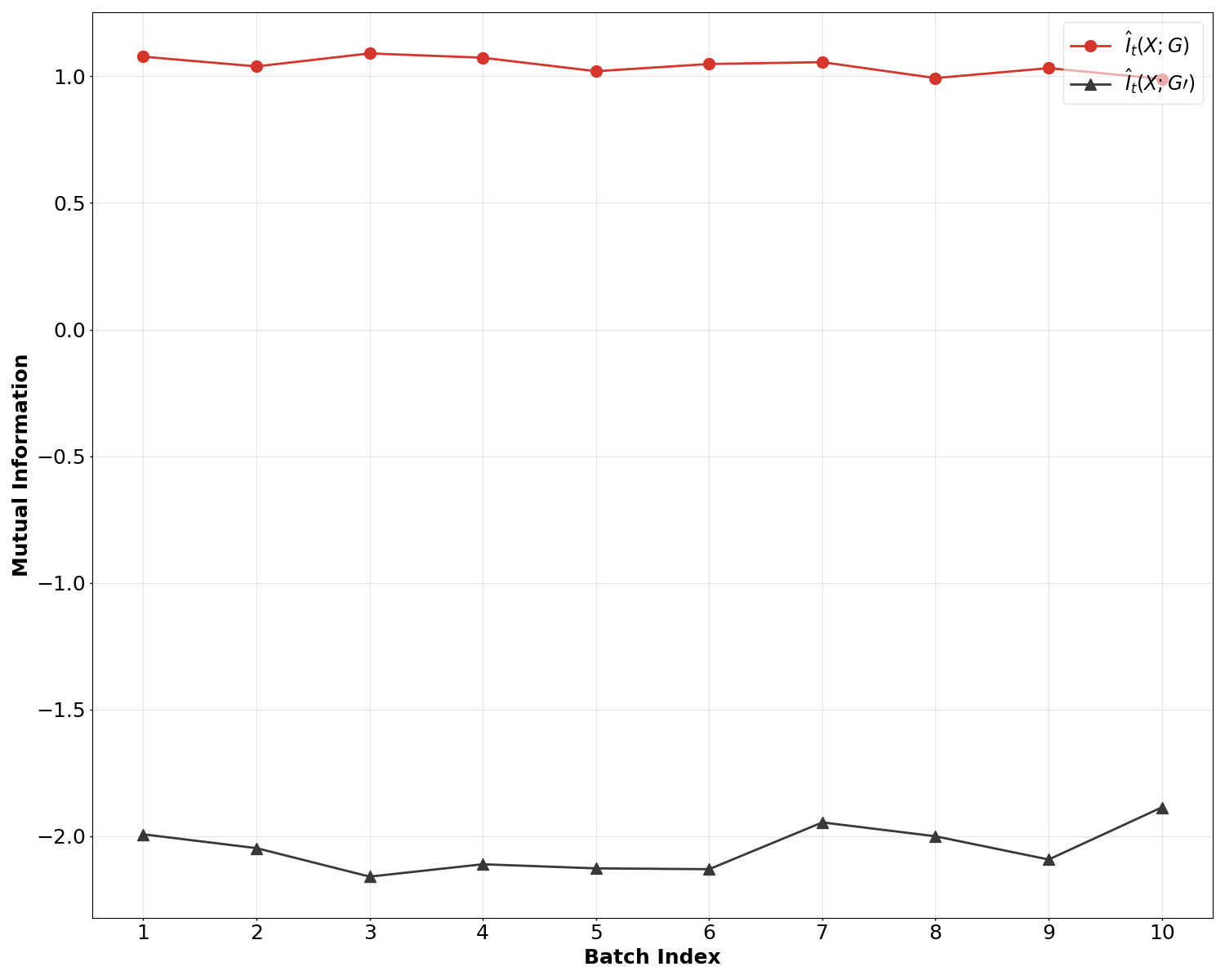}
		\label{let-t-8}
    }\\
    \subfigure[Sub-Epoch 60 (Transformer)]{
		\includegraphics[width=0.34\textwidth]{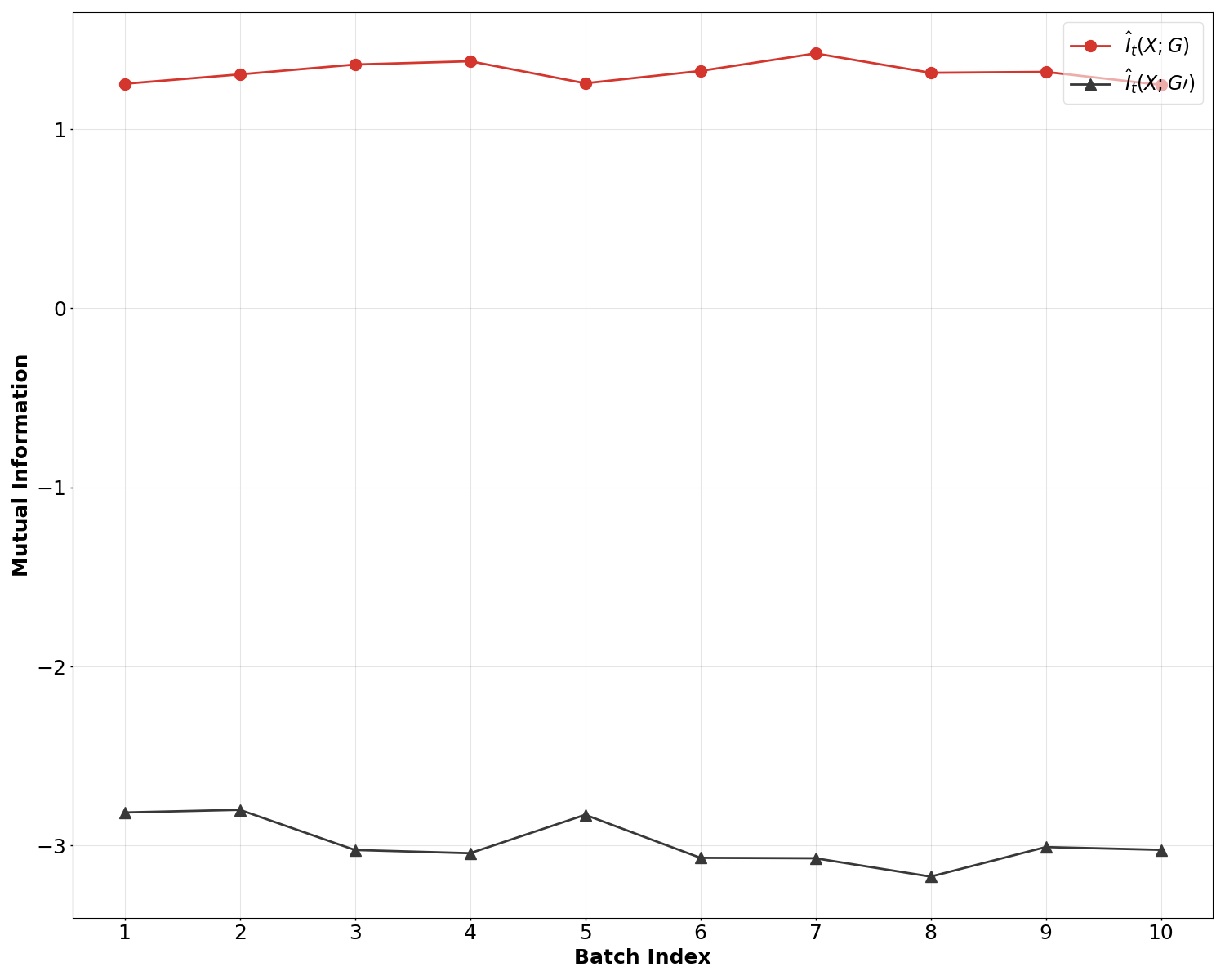}
		\label{let-t-60}
    }
    \subfigure[Sub-Epoch 200 (Transformer)]{
		\includegraphics[width=0.34\textwidth]{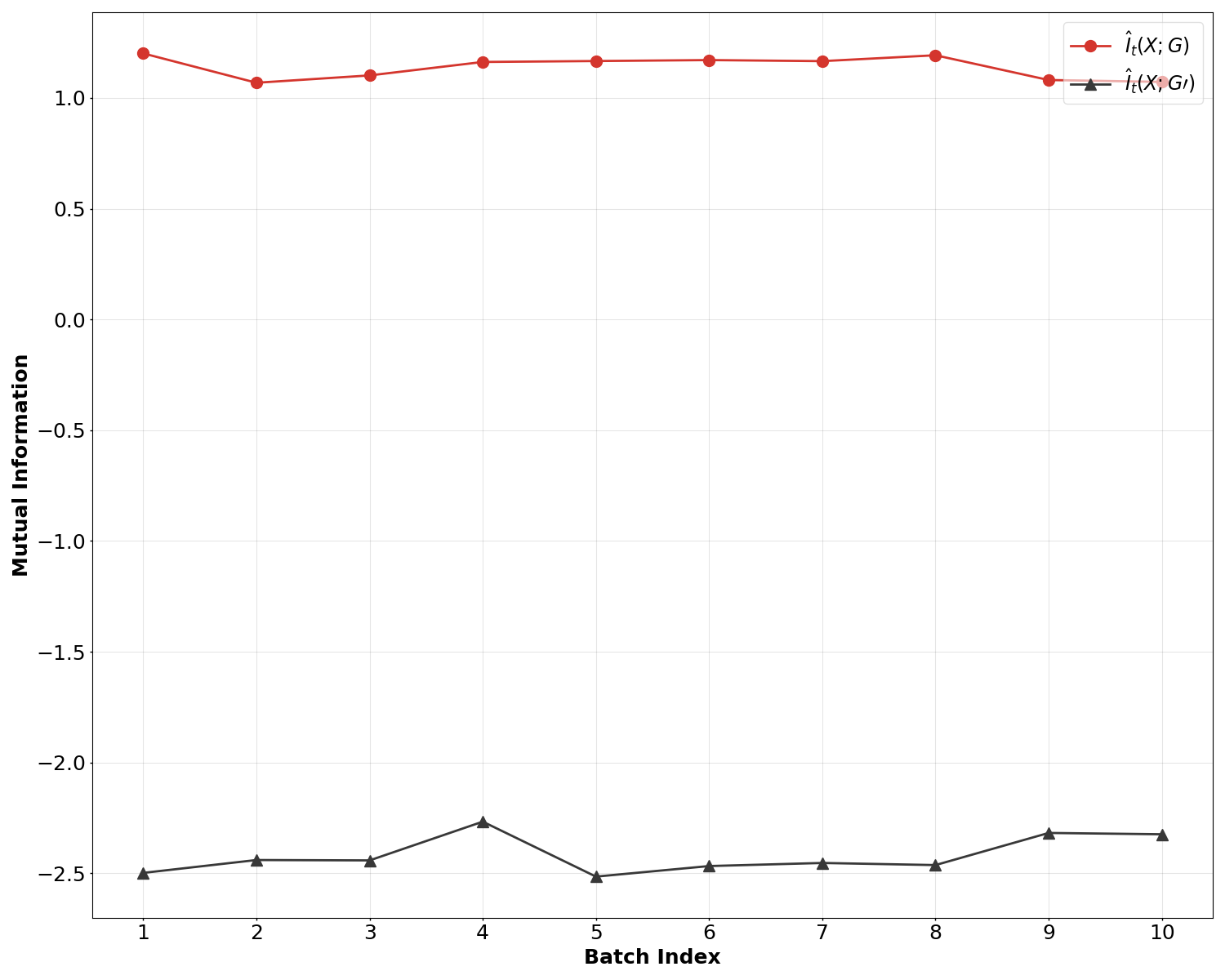}
		\label{let-t-200}
    }
    \caption{Mutual information estimation on LeNet (CIFAR-10).}
    \label{lenet}
\end{figure}

\begin{figure}[htp]
    \centering
    \vspace{-2mm}
    \subfigure[Sub-Epoch 0 (Autoencoder)]{
		\includegraphics[width=0.34\textwidth]{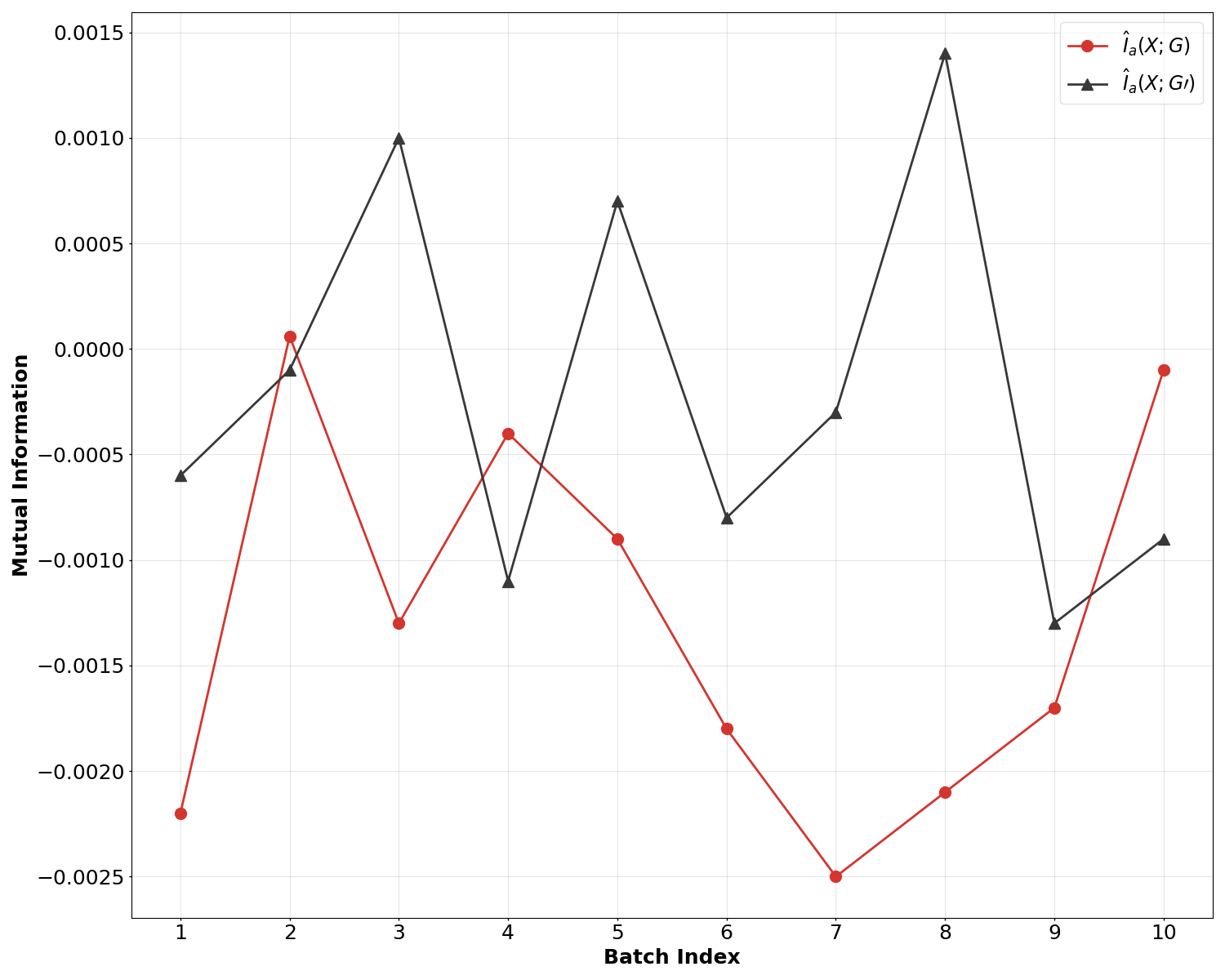}
		\label{ale-a-0}
    }
    \subfigure[Sub-Epoch 30 (Autoencoder)]{
		\includegraphics[width=0.34\textwidth]{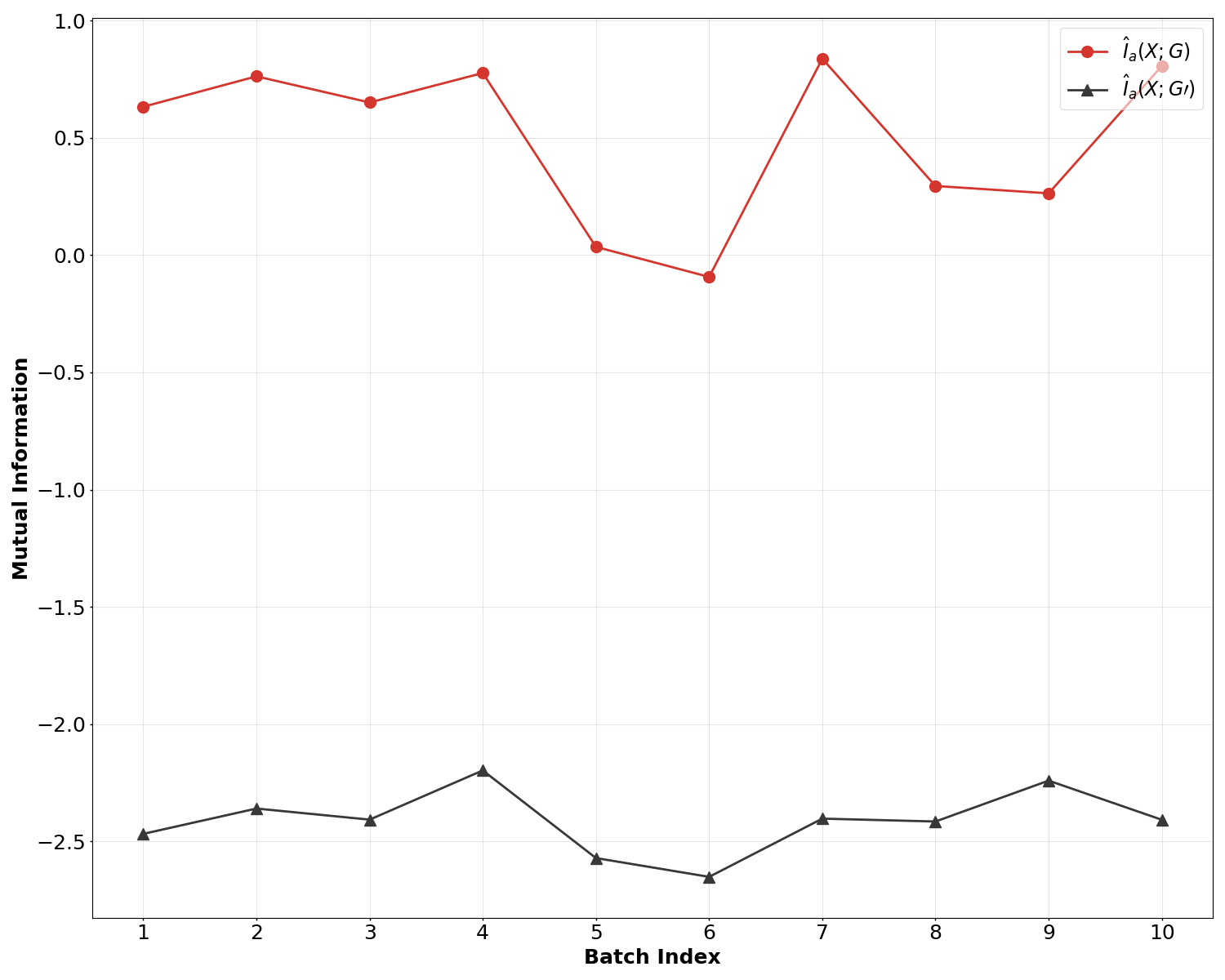}
		\label{ale-a-30}
    }
    \subfigure[Sub-Epoch 120 (Autoencoder)]{
		\includegraphics[width=0.34\textwidth]{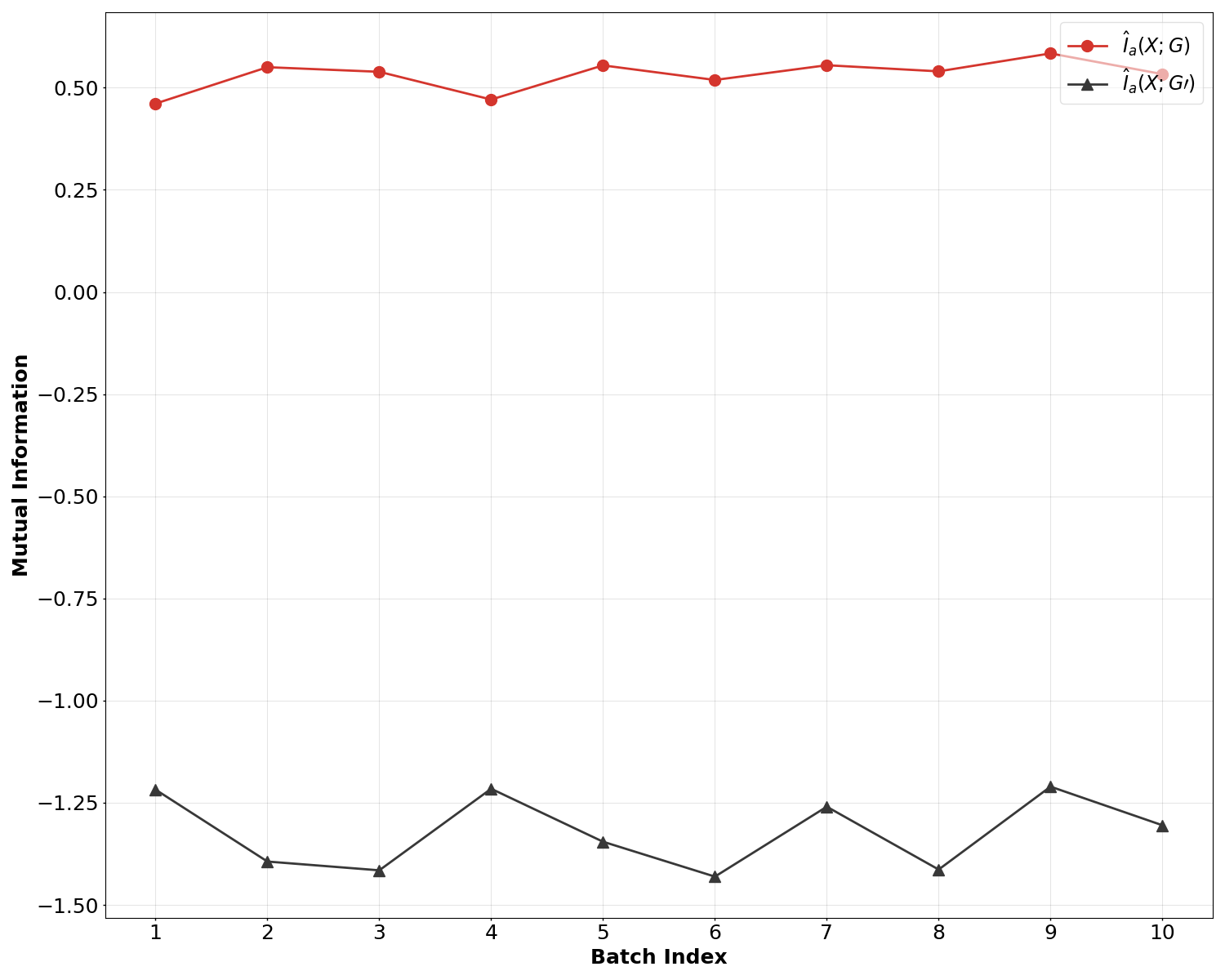}
		\label{ale-a-120}
    }
    \subfigure[Sub-Epoch 200 (Autoencoder)]{
		\includegraphics[width=0.34\textwidth]{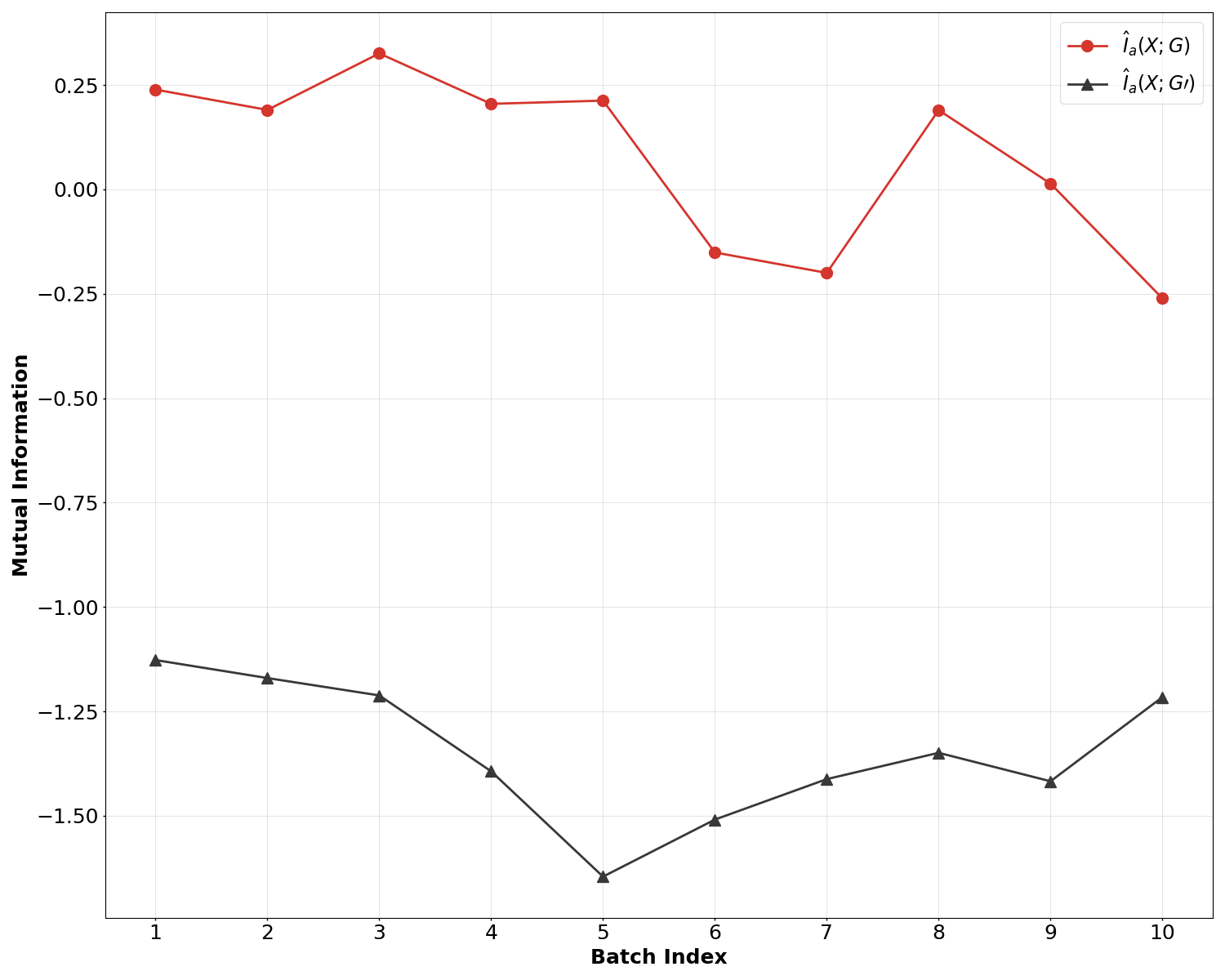}
		\label{ale-a-200}
    }\\
    \subfigure[Sub-Epoch 0 (Transformer)]{
		\includegraphics[width=0.34\textwidth]{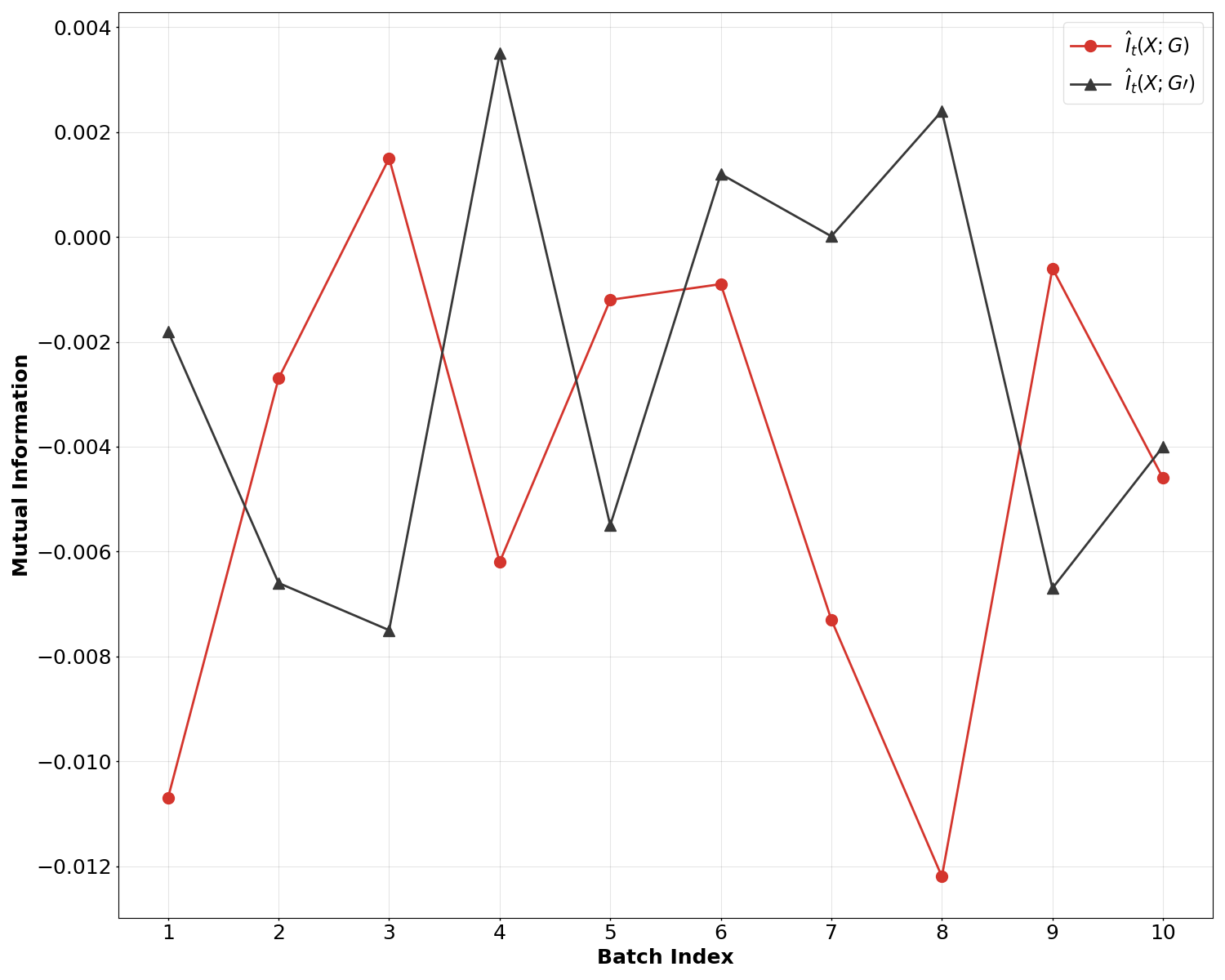}
		\label{ale-t-0}
    }
    \subfigure[Sub-Epoch 30 (Transformer)]{
		\includegraphics[width=0.34\textwidth]{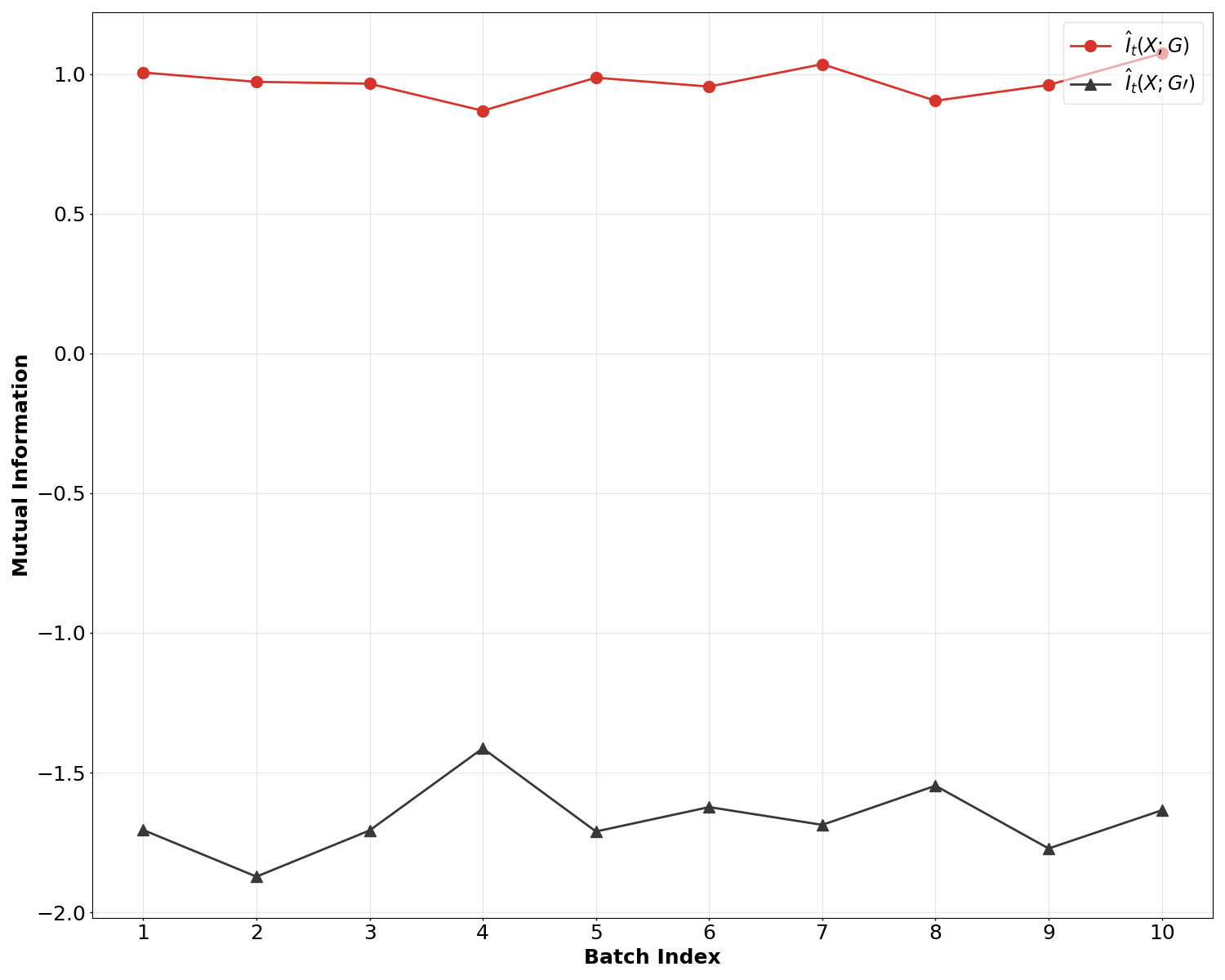}
		\label{ale-t-30}
    }
    \subfigure[Sub-Epoch 120 (Transformer)]{
		\includegraphics[width=0.34\textwidth]{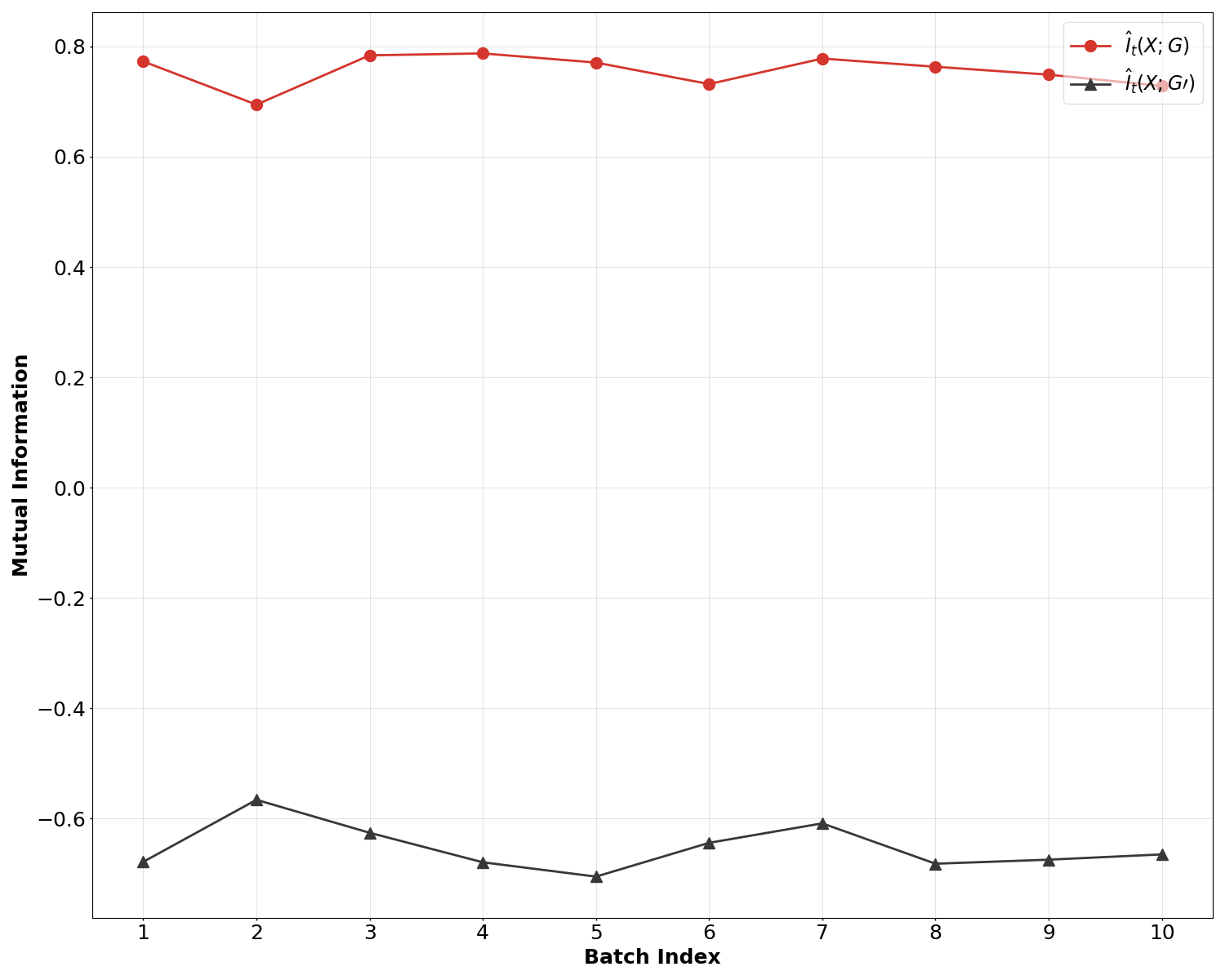}
		\label{ale-t-120}
    }
    \subfigure[Sub-Epoch 200 (Transformer)]{
		\includegraphics[width=0.34\textwidth]{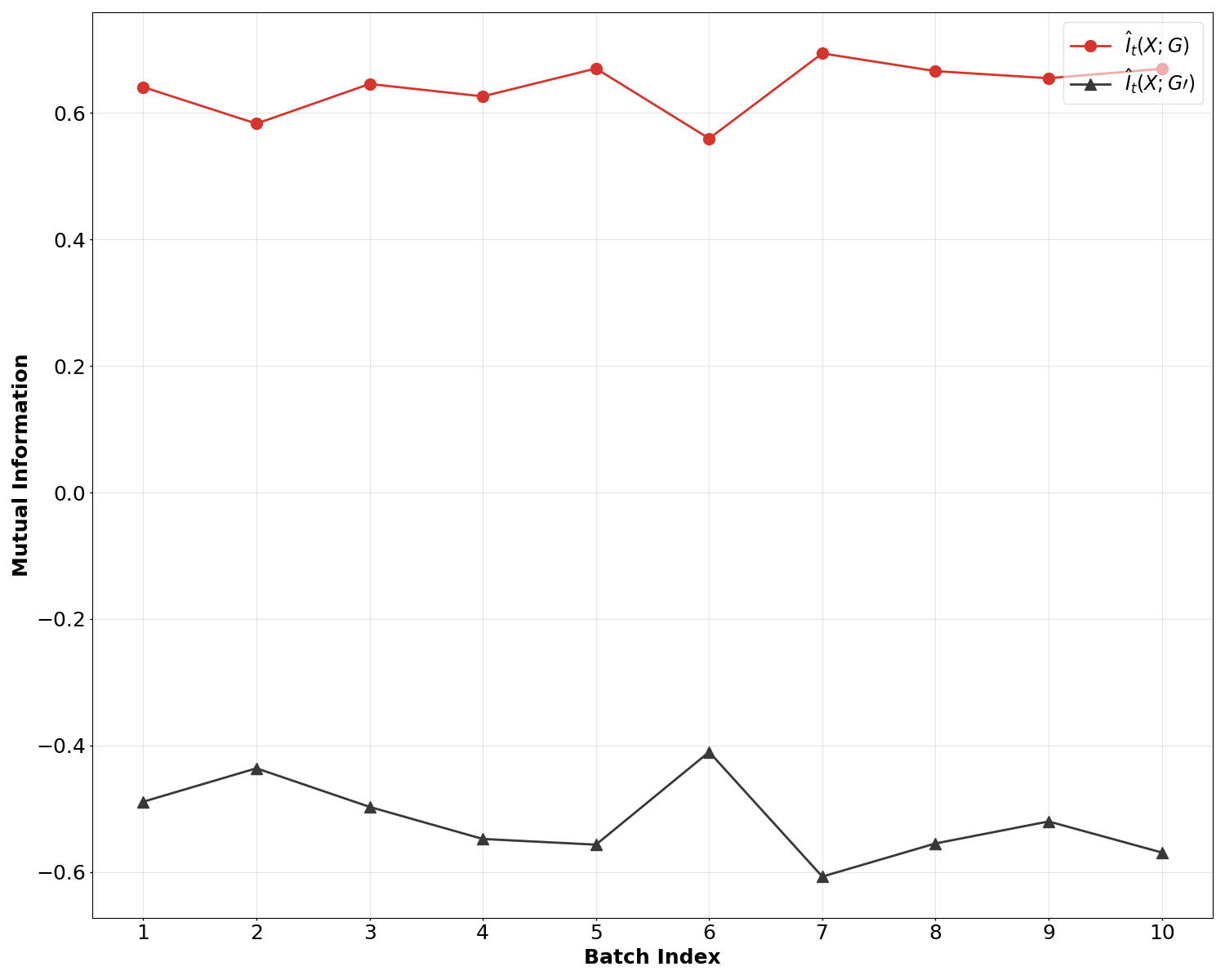}
		\label{ale-t-200}
    }
    \caption{Mutual information estimation on AlexNet (CIFAR-10).}
    \label{alexnet}
\end{figure}

\begin{figure}[htp]
    \centering
    \vspace{-2mm}
    \subfigure[Sub-Epoch 0 (Autoencoder)]{
		\includegraphics[width=0.34\textwidth]{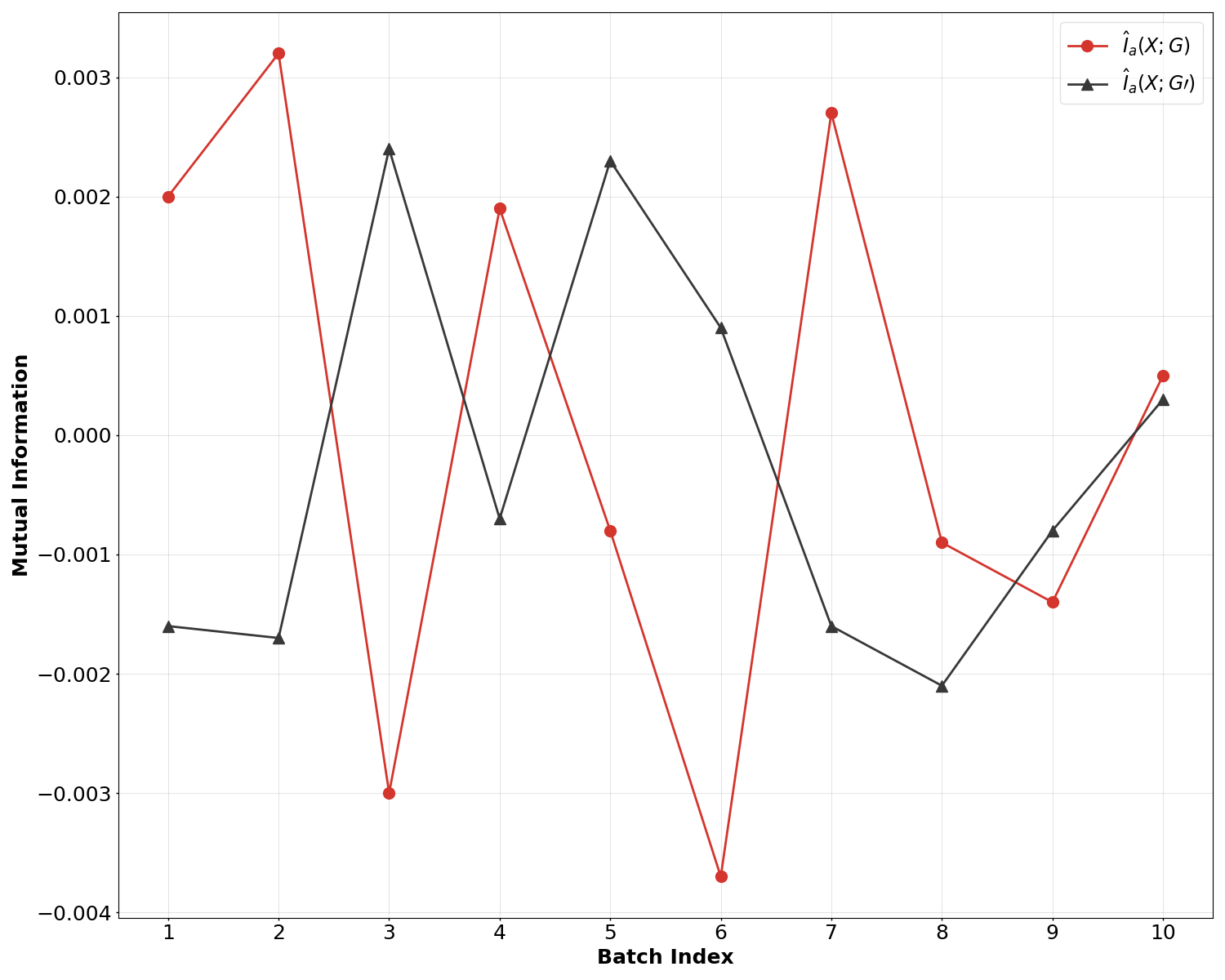}
		\label{cnn-a-0}
    }
    \subfigure[Sub-Epoch 10 (Autoencoder)]{
		\includegraphics[width=0.34\textwidth]{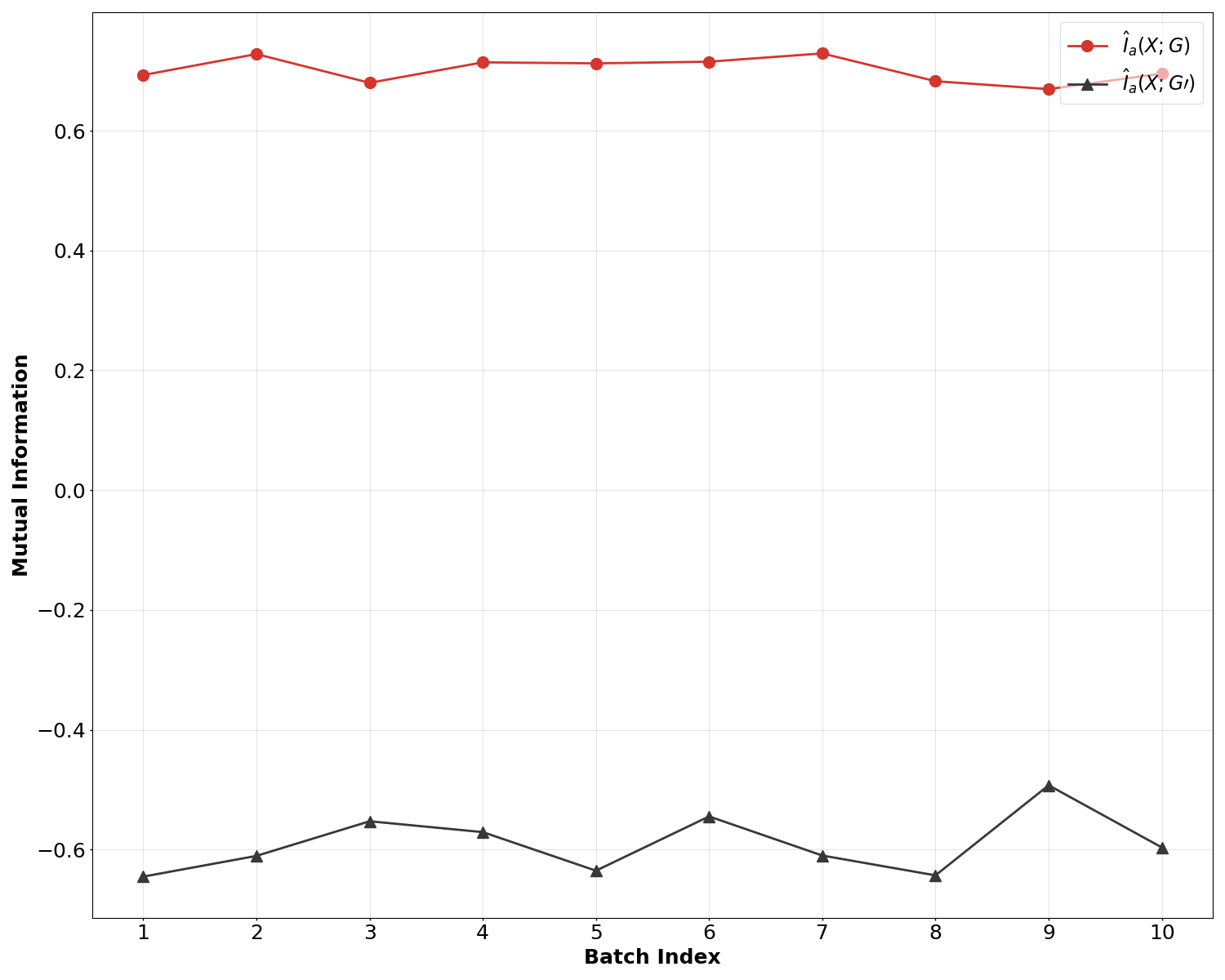}
		\label{cnn-a-10}
    }
    \subfigure[Sub-Epoch 20 (Autoencoder)]{
		\includegraphics[width=0.34\textwidth]{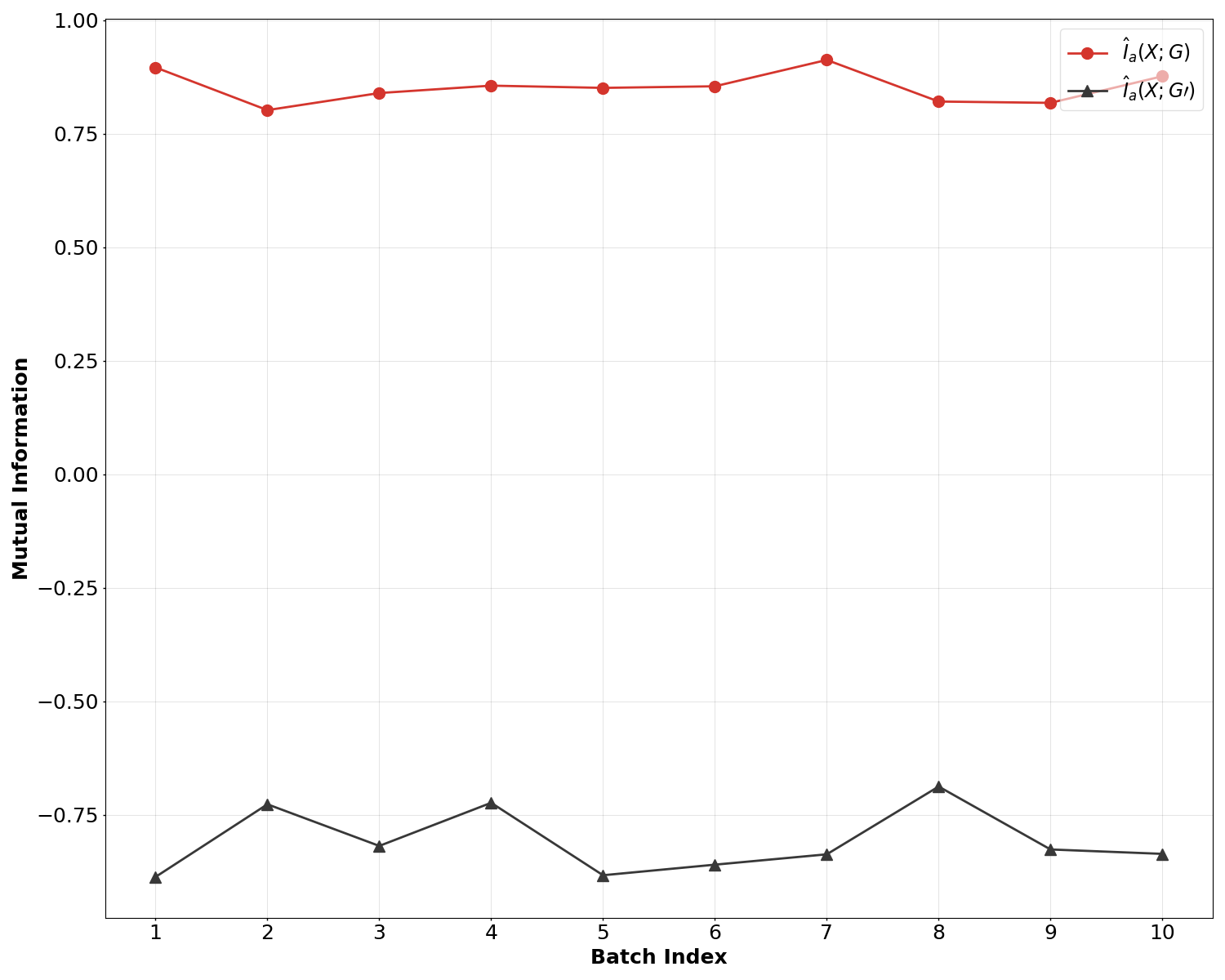}
		\label{cnn-a-20}
    }
    \subfigure[Sub-Epoch 100 (Autoencoder)]{
		\includegraphics[width=0.34\textwidth]{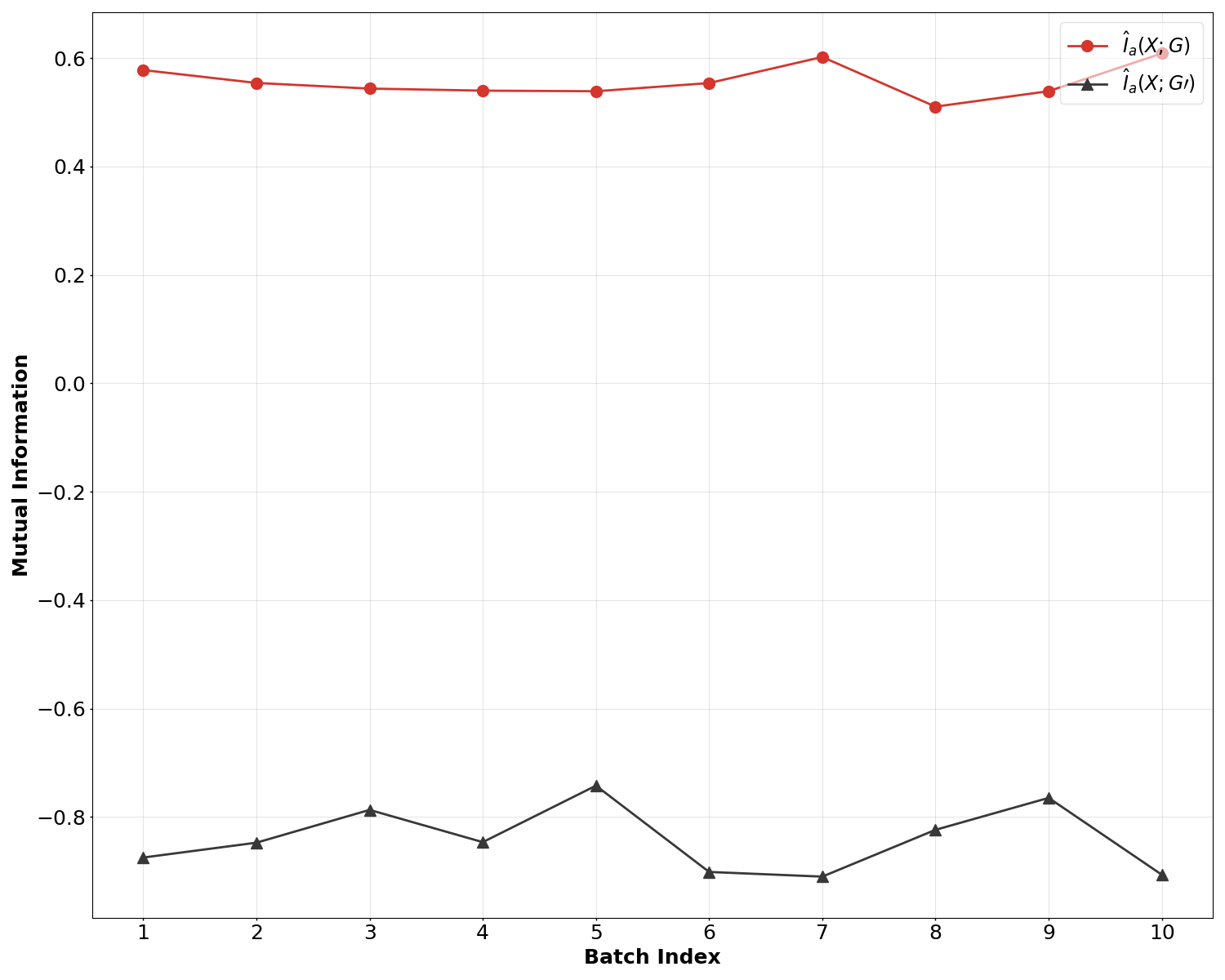}
		\label{cnn-a-100}
    }\\
    \subfigure[Sub-Epoch 0 (Transformer)]{
		\includegraphics[width=0.34\textwidth]{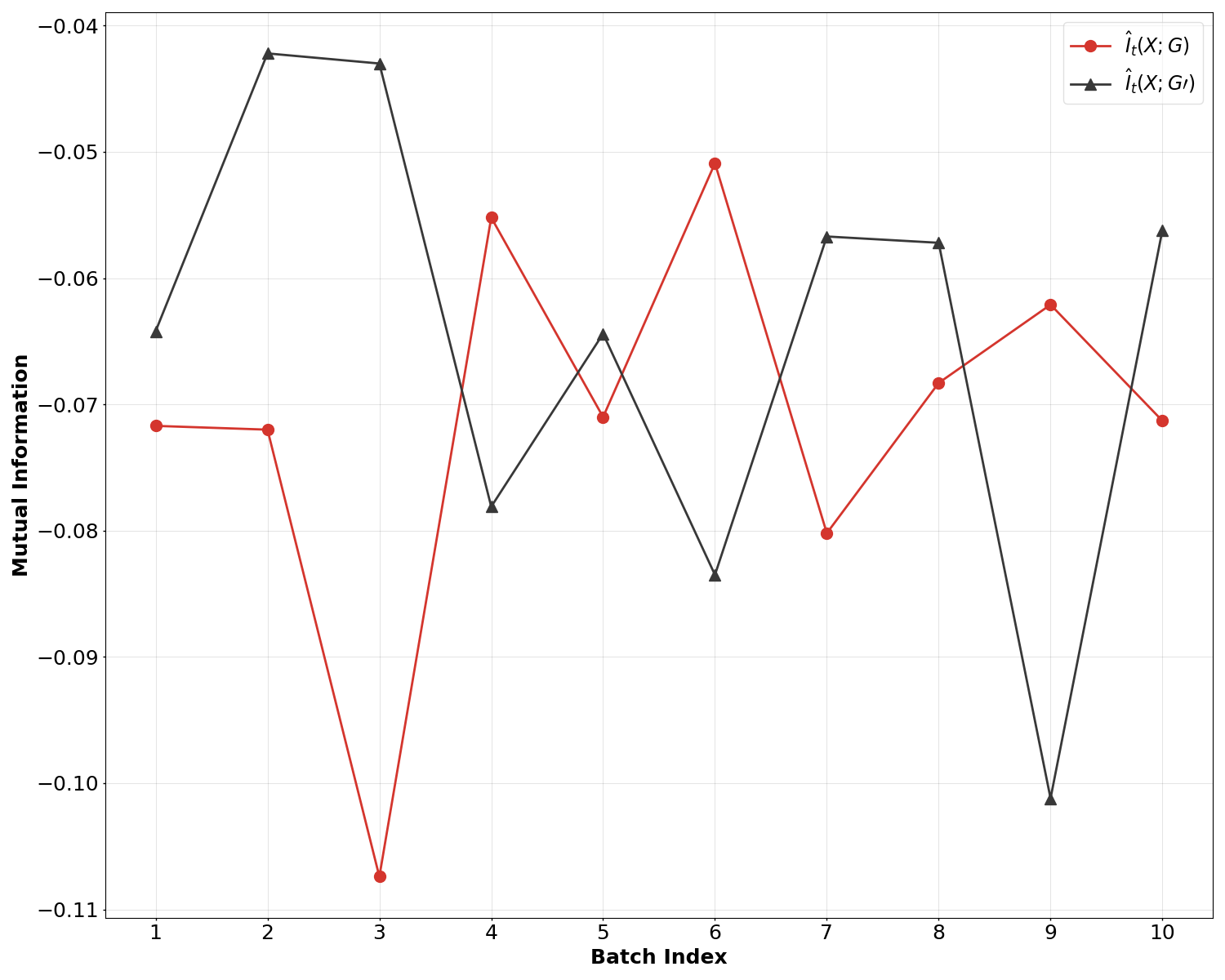}
		\label{cnn-t-0}
    }
    \subfigure[Sub-Epoch 10 (Transformer)]{
		\includegraphics[width=0.34\textwidth]{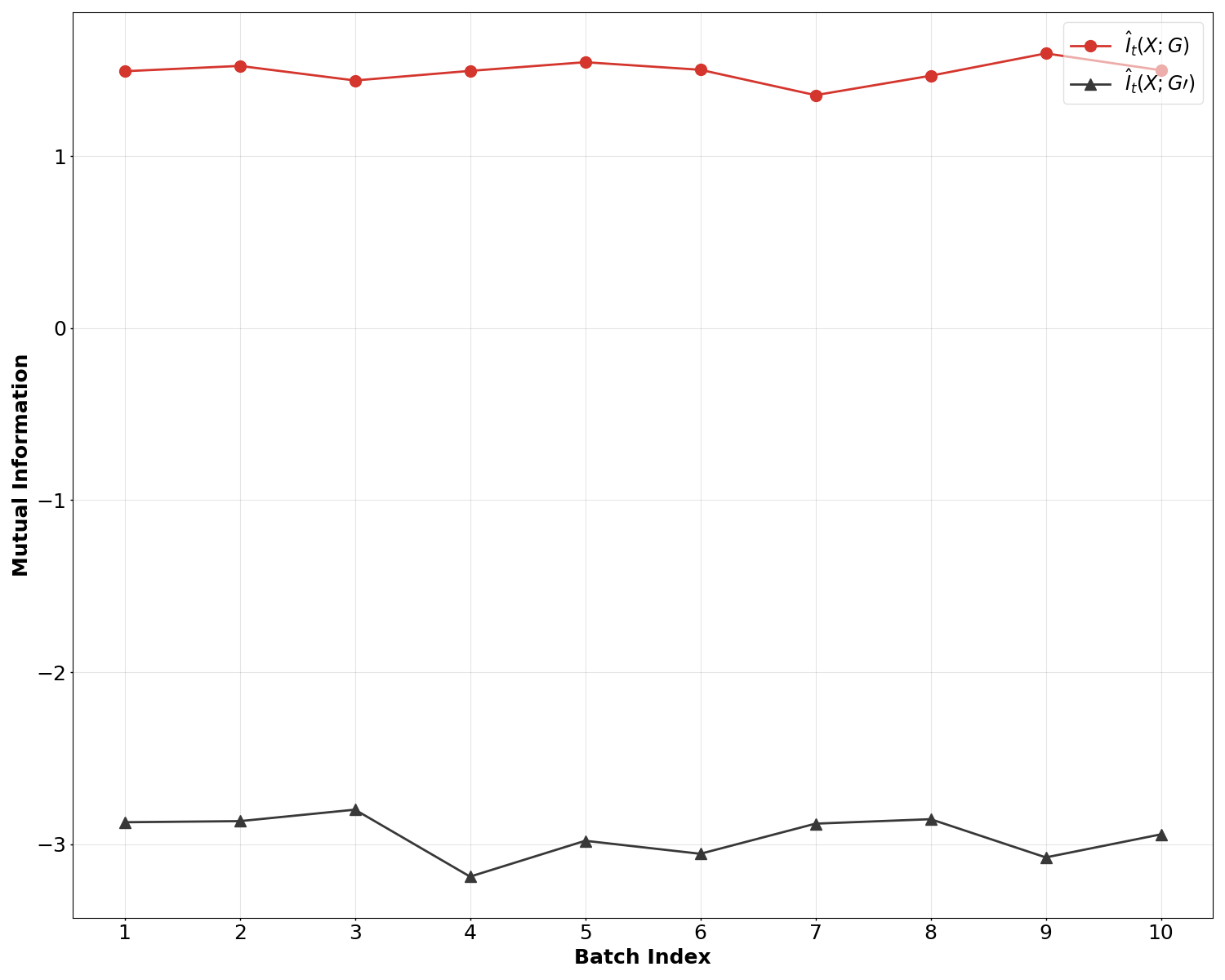}
		\label{cnn-t-10}
    }
    \subfigure[Sub-Epoch 20 (Transformer)]{
		\includegraphics[width=0.34\textwidth]{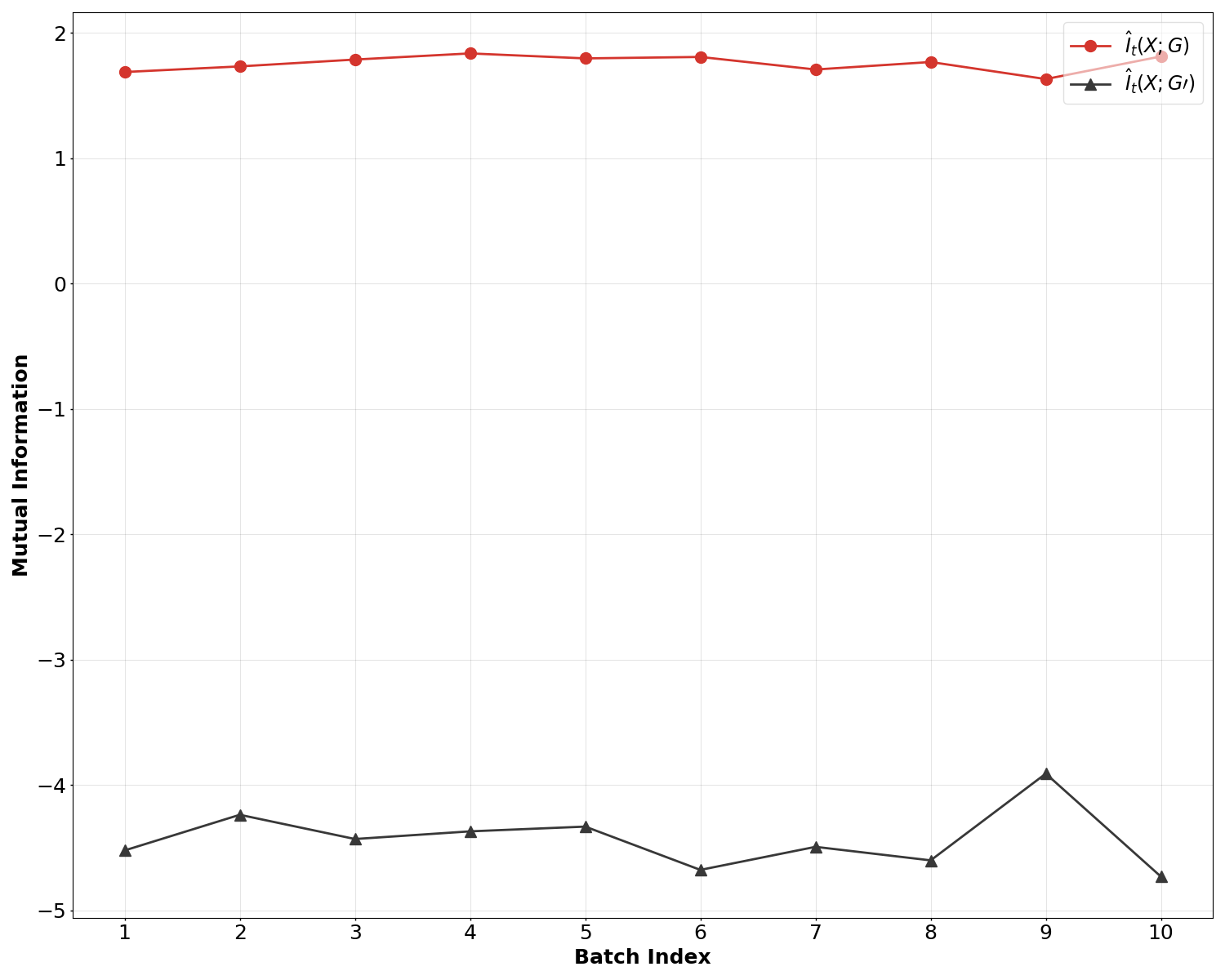}
		\label{cnn-t-20}
    }
    \subfigure[Sub-Epoch 100 (Transformer)]{
		\includegraphics[width=0.34\textwidth]{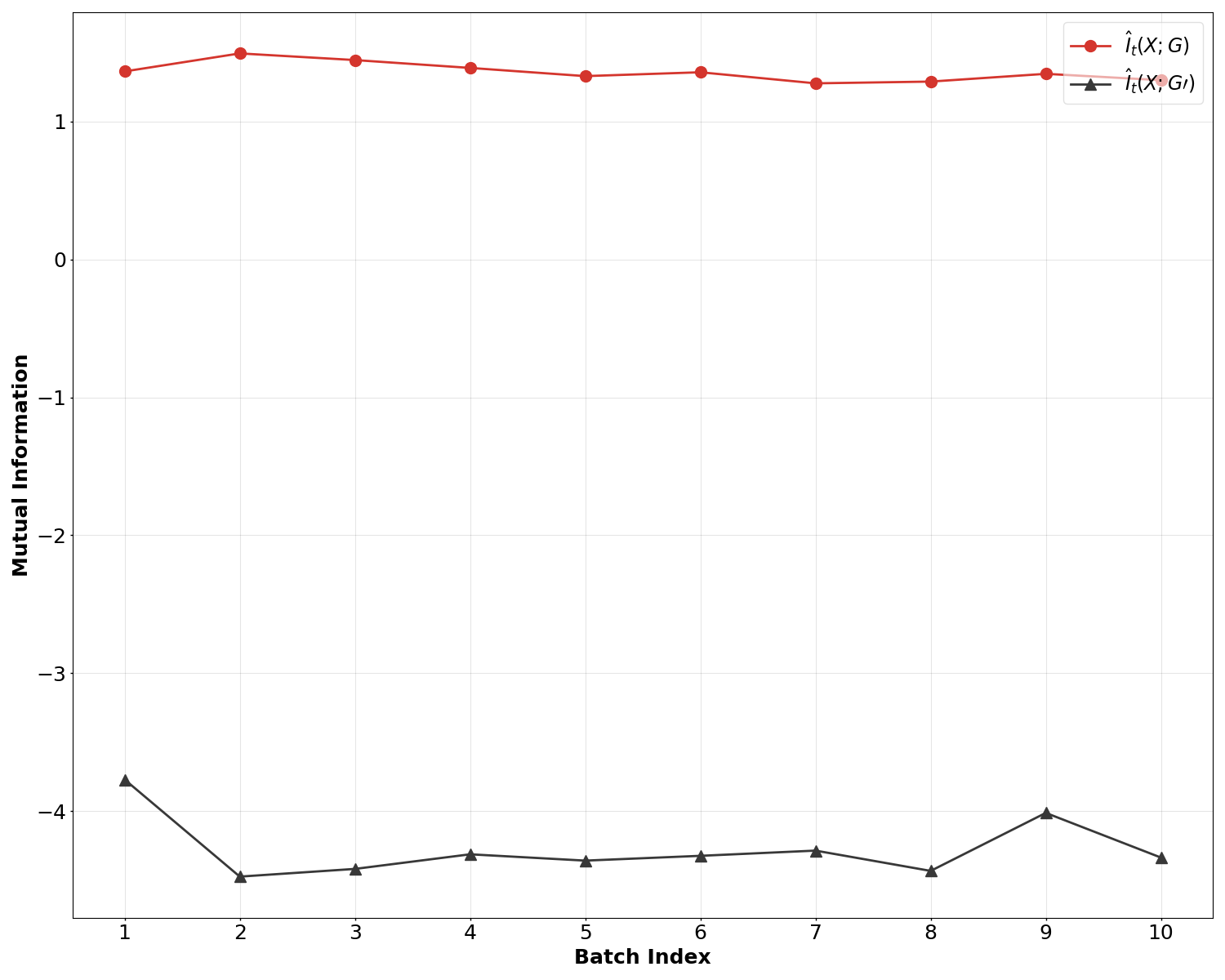}
		\label{cnn-t-100}
    }
    \caption{Mutual information estimation on CNN (CelebA-HQ).}
    \label{cnn}
\end{figure}

\begin{figure}[htp]
    \centering
    \vspace{-2mm}
    \subfigure[Sub-Epoch 0 (Autoencoder)]{
		\includegraphics[width=0.34\textwidth]{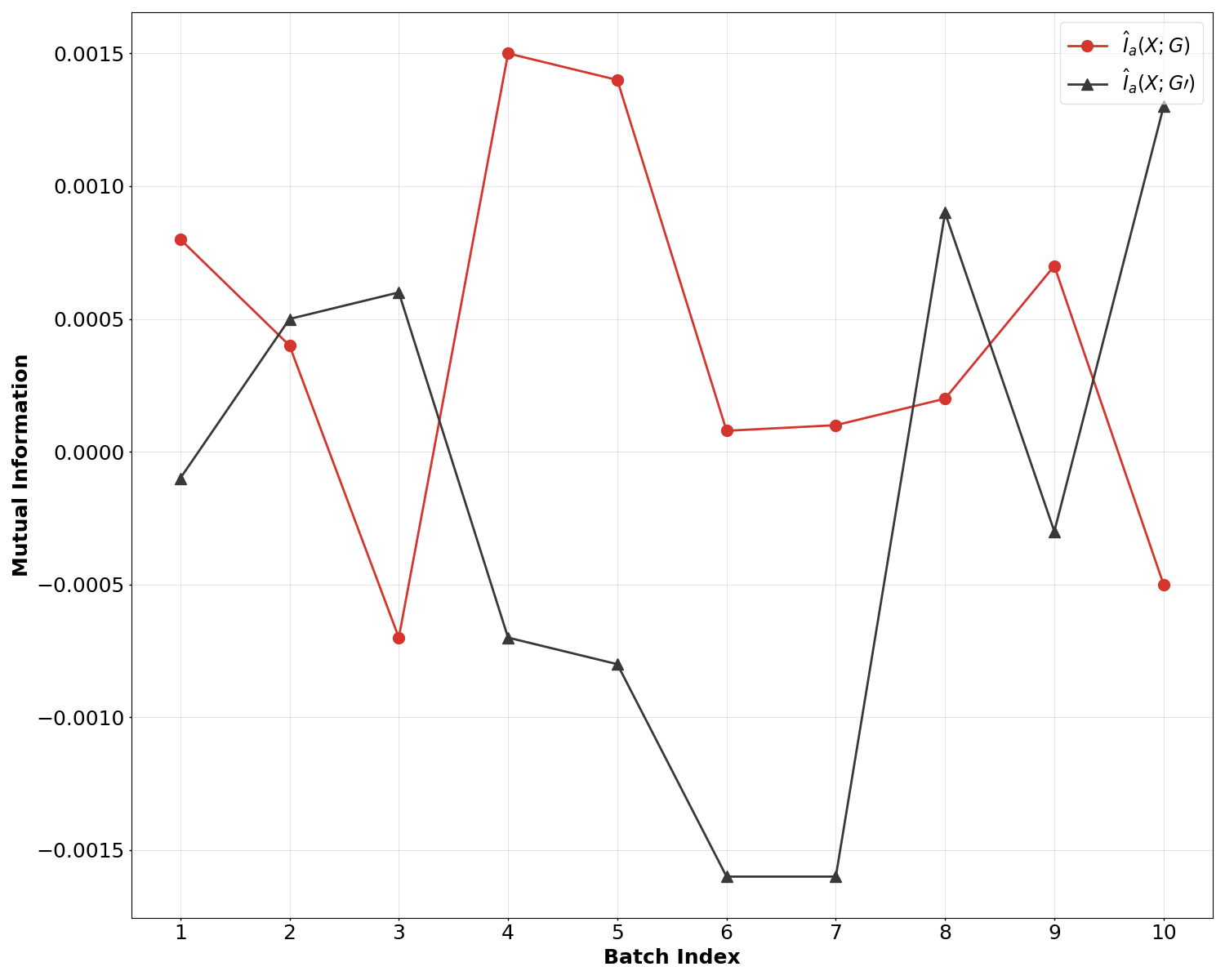}
		\label{mlp-a-0}
    }
    \subfigure[Sub-Epoch 10 (Autoencoder)]{
		\includegraphics[width=0.34\textwidth]{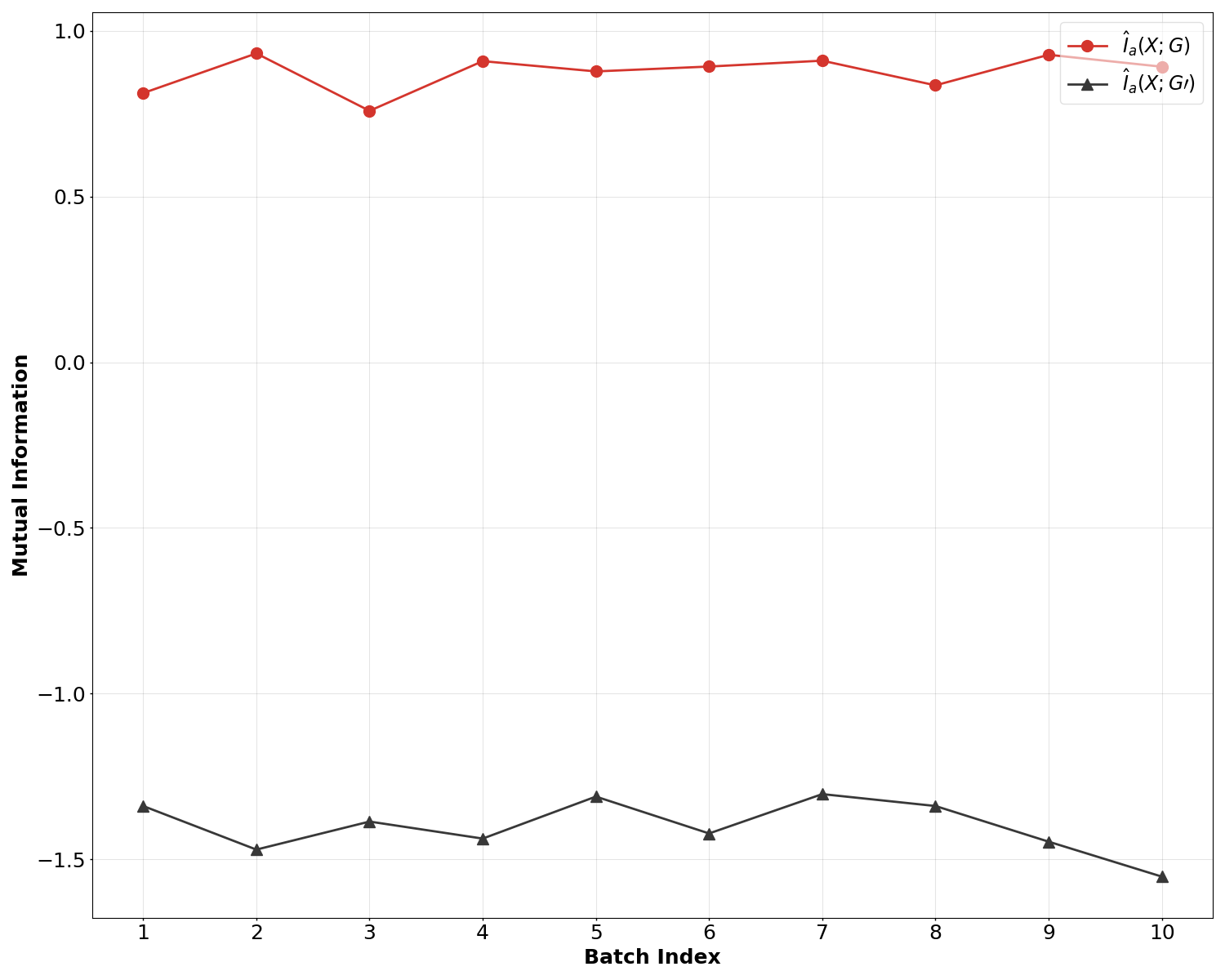}
		\label{mlp-a-10}
    }
    \subfigure[Sub-Epoch 20 (Autoencoder)]{
		\includegraphics[width=0.34\textwidth]{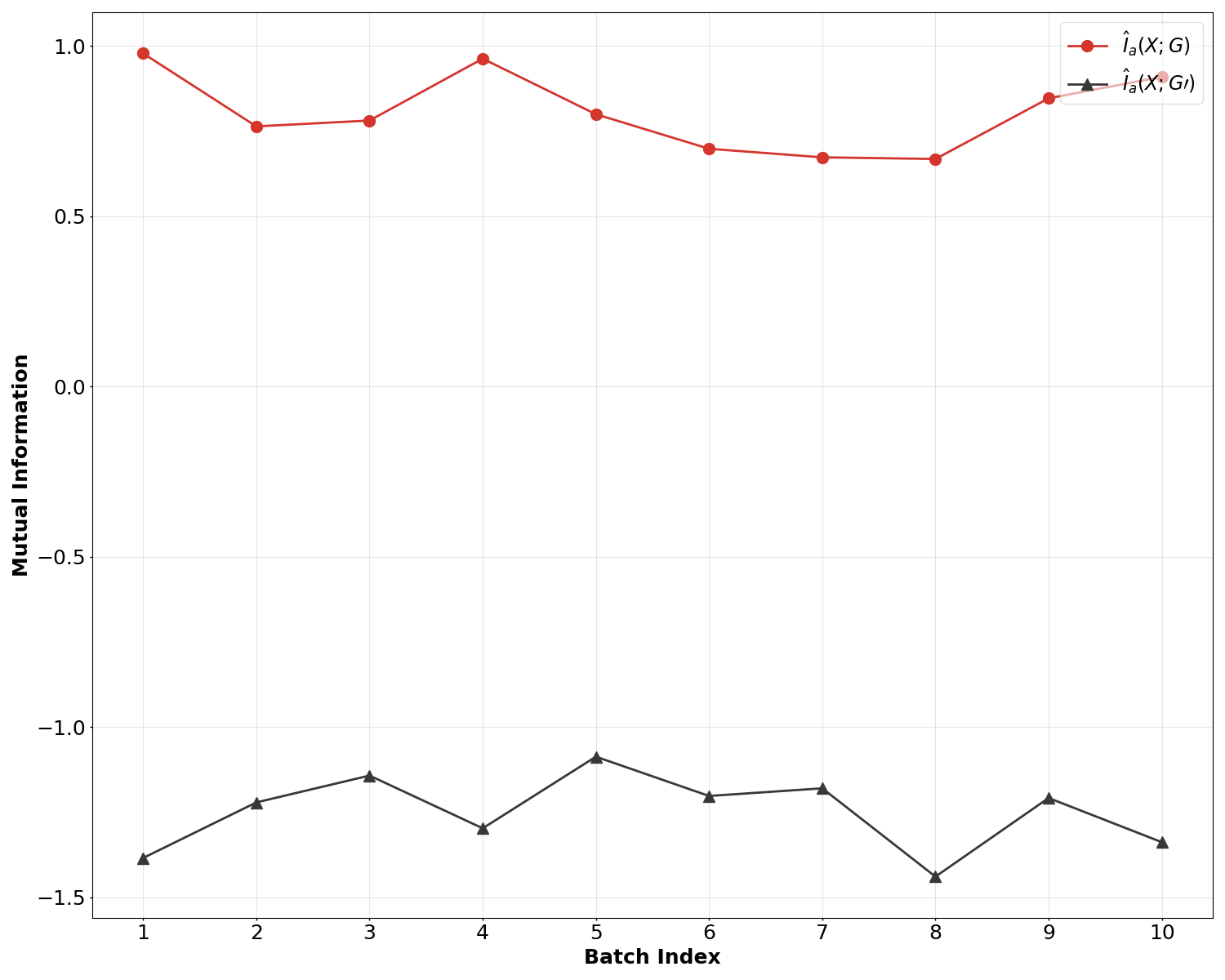}
		\label{mlp-a-20}
    }
    \subfigure[Sub-Epoch 100 (Autoencoder)]{
		\includegraphics[width=0.34\textwidth]{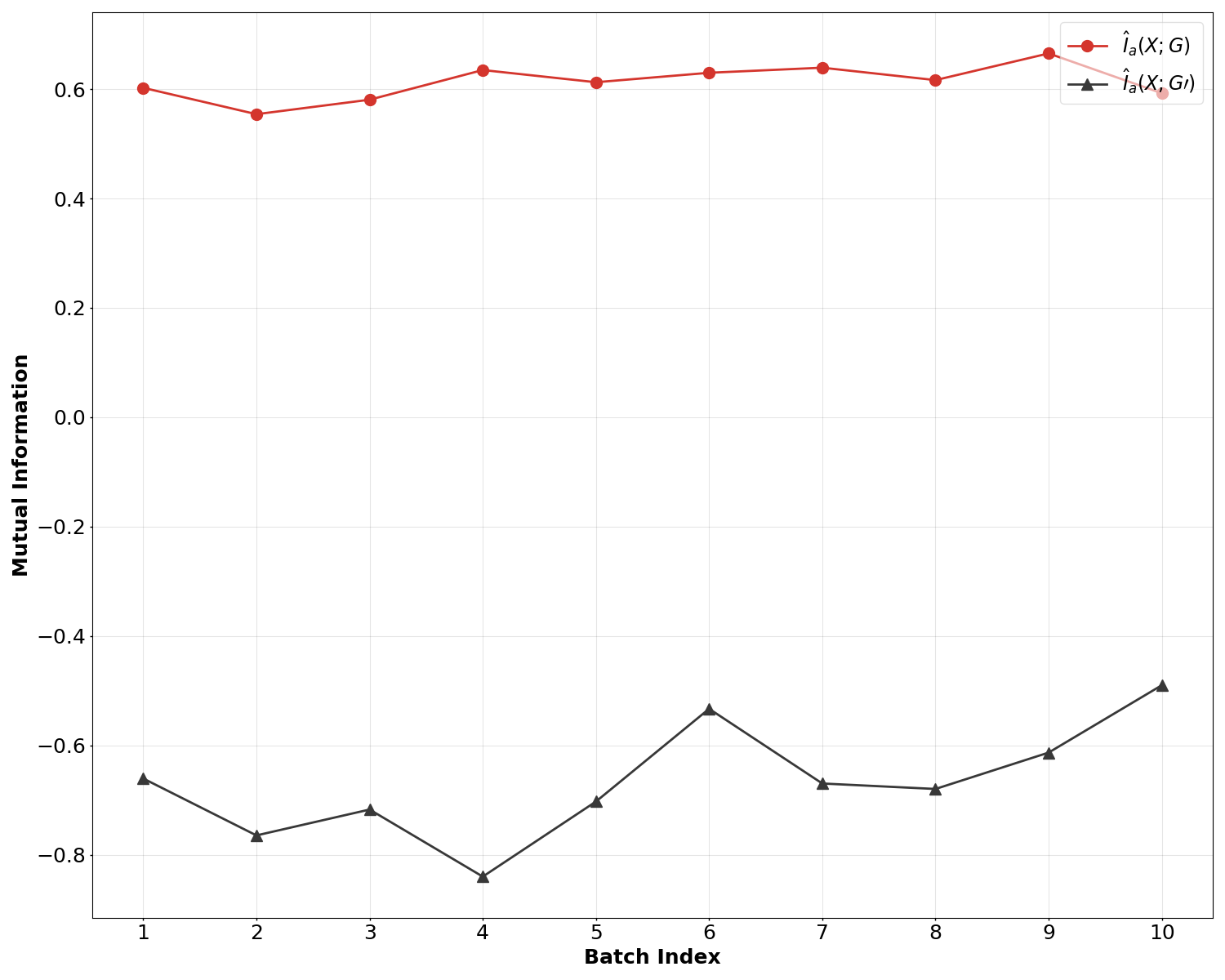}
		\label{mlp-a-100}
    }\\
    \subfigure[Sub-Epoch 0 (Transformer)]{
		\includegraphics[width=0.34\textwidth]{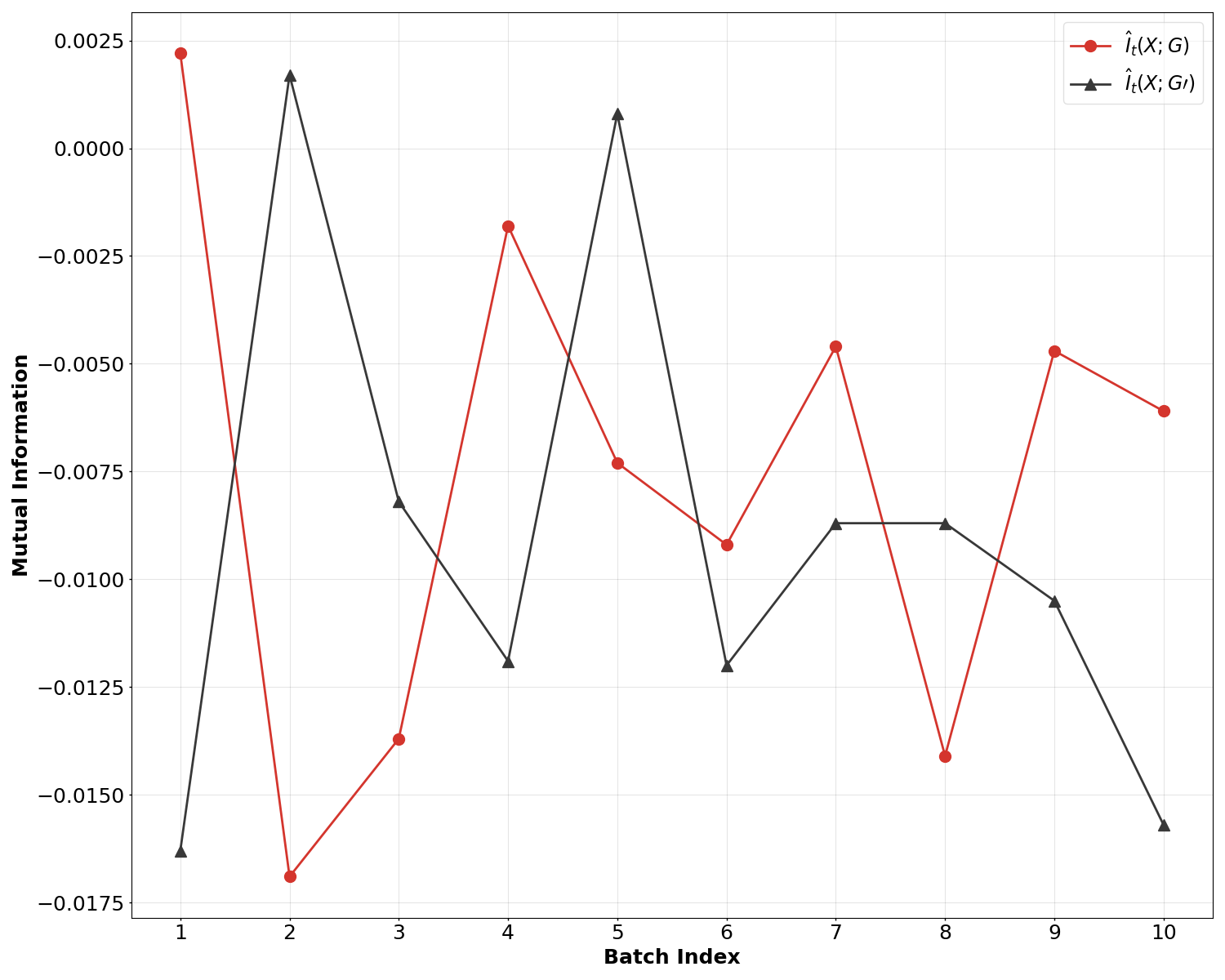}
		\label{mlp-t-0}
    }
    \subfigure[Sub-Epoch 10 (Transformer)]{
		\includegraphics[width=0.34\textwidth]{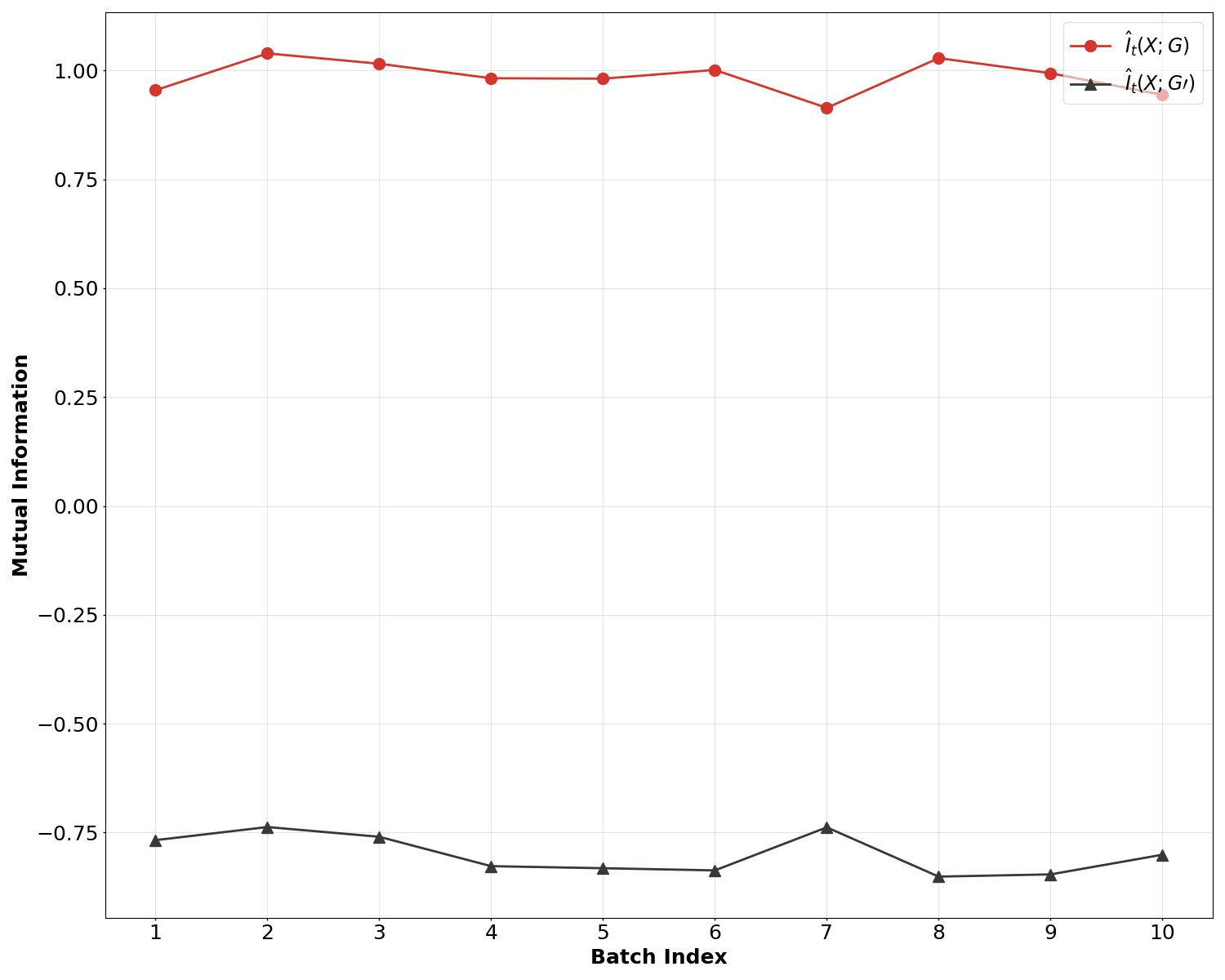}
		\label{mlp-t-10}
    }
    \subfigure[Sub-Epoch 20 (Transformer)]{
		\includegraphics[width=0.34\textwidth]{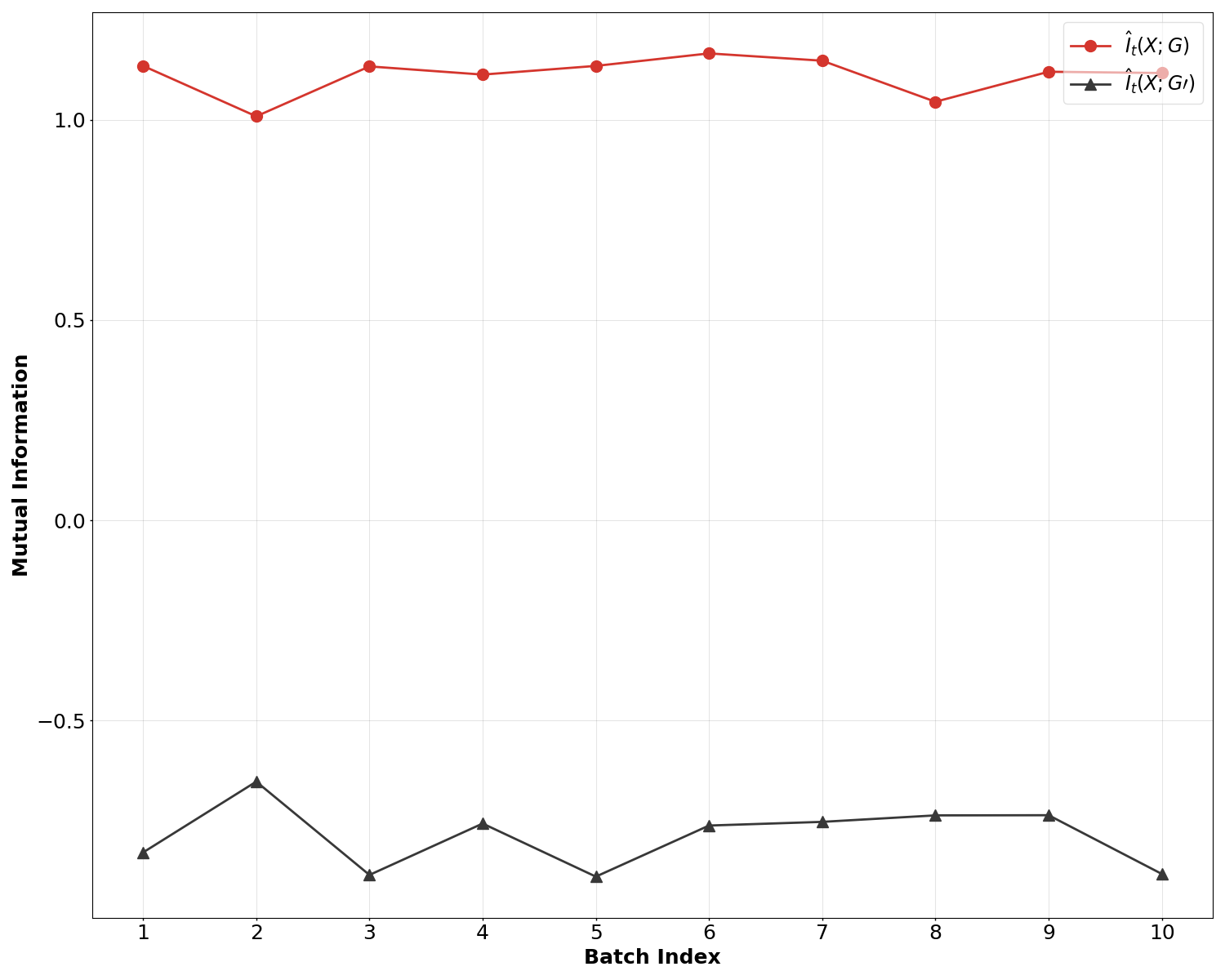}
		\label{mlp-t-20}
    }
    \subfigure[Sub-Epoch 100 (Transformer)]{
		\includegraphics[width=0.34\textwidth]{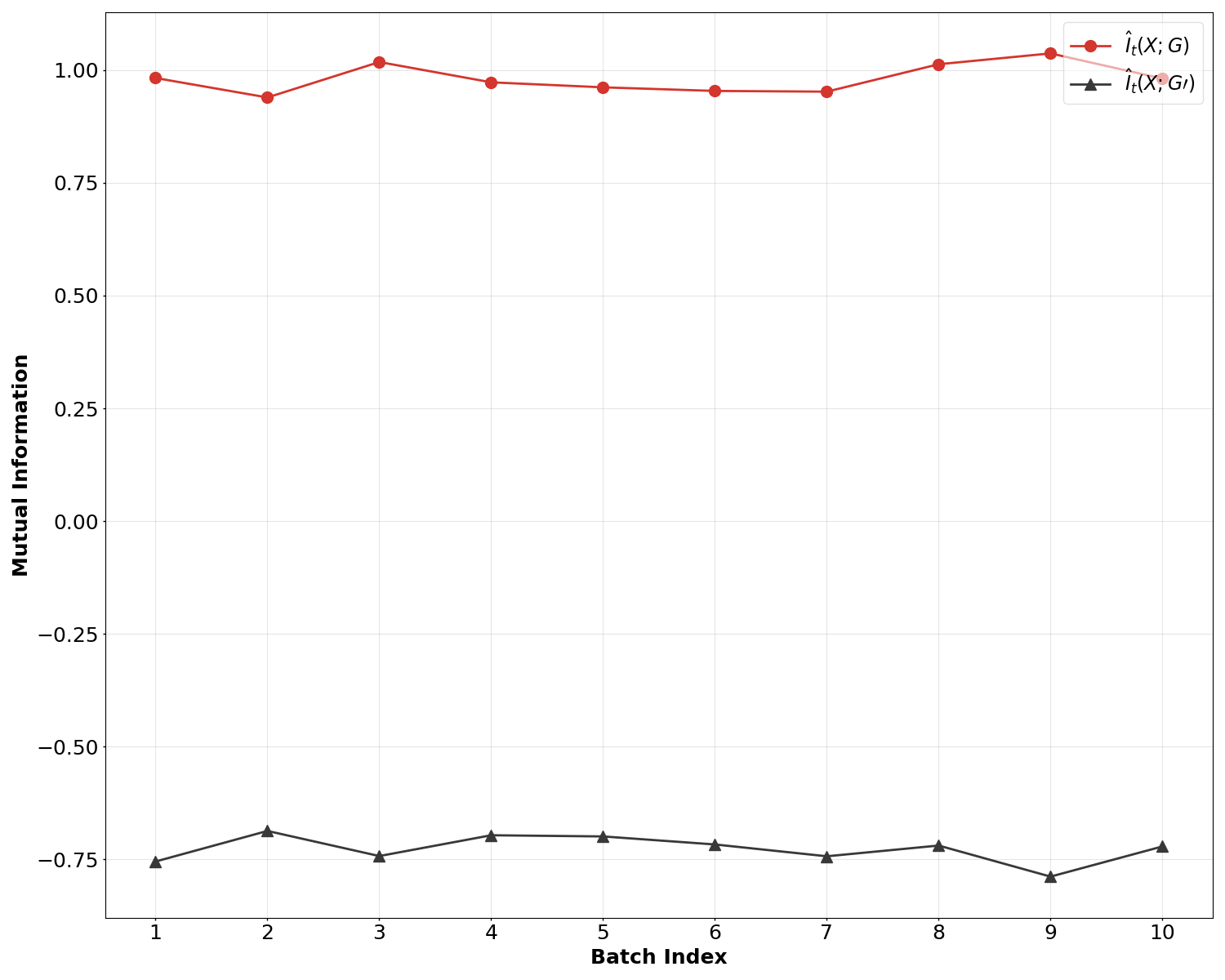}
		\label{mlp-t-100}
    }
    \caption{Mutual information estimation on MLP (CelebA-HQ).}
    \label{mlp}
\end{figure}

In Section \ref{sec:exp}, we present the average difference in mutual information between matched and unmatched data-gradient pairs across 200 batches of the training data. To enhance the clarity of the experimental results, this section provides detailed mutual information values for both matched and unmatched pairs. The results are shown in Figures \ref{lenet}-\ref{mlp}. The red line represents the estimated mutual information $\hat{I}_{\{a,t\}}(\mathcal{X};G)$ of matched data-gradient pairs while the black line represents the estimated mutual information $\hat{I}_{\{a,t\}}(\mathcal{X};G^{\prime})$ of mismatched data-gradient pairs. Notably, except during the model's initialization phase (sub-epoch 0), where training has not yet commenced, no consistent pattern is observed between the mutual information of matched data-gradient pairs and that of unmatched pairs. In subsequent sub-epochs, however, the mutual information of matched pairs consistently exceeds that of unmatched pairs. This indicates that when an attacker gains access to the gradient of the original data, their information gain increases.

\begin{figure}[htp]
    \centering
    \vspace{-2mm}
    \subfigure[Autoencoder]{
		\includegraphics[width=0.42\textwidth]{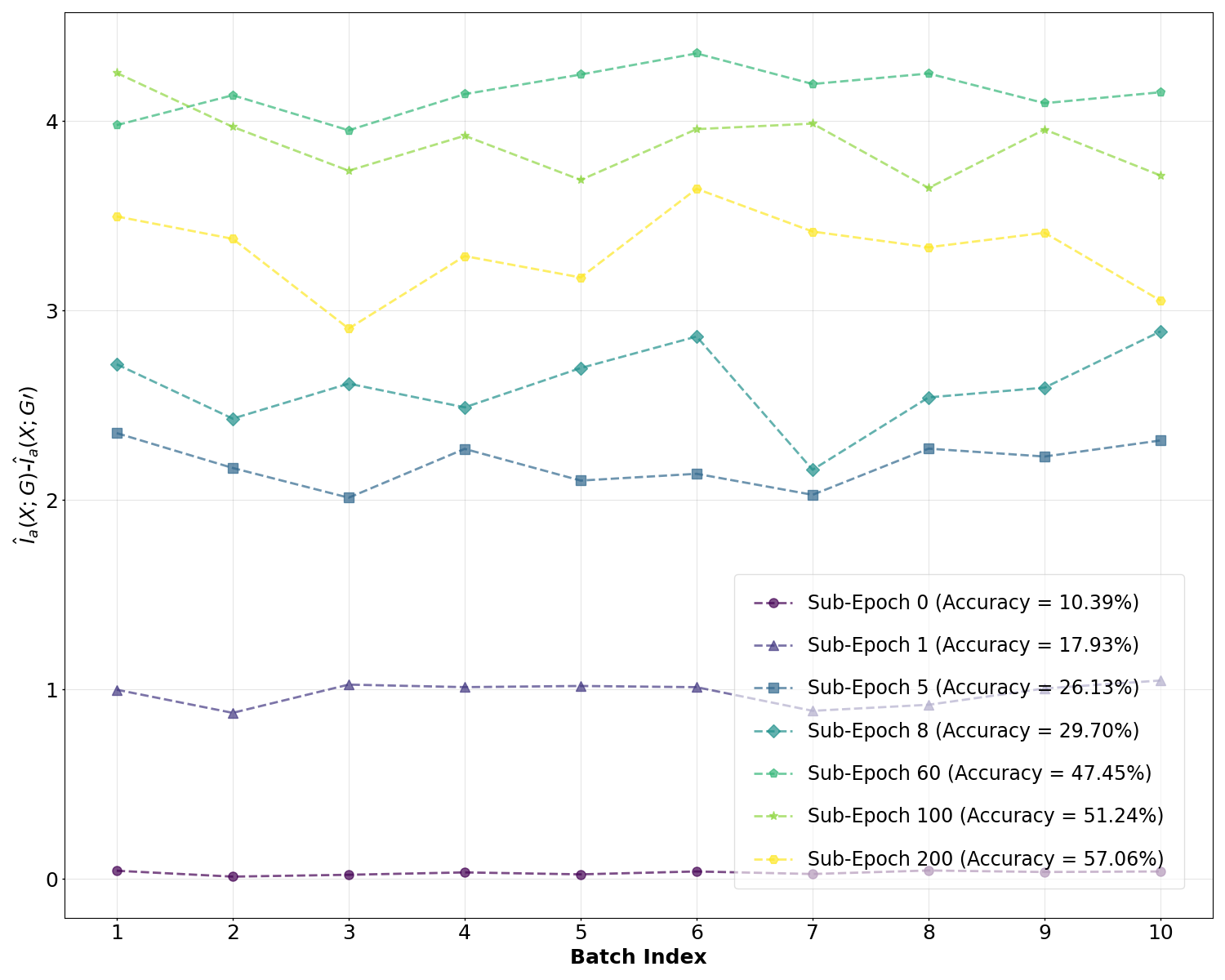}
		\label{diff-let-a}
    }
    \subfigure[Transformer]{
		\includegraphics[width=0.42\textwidth]{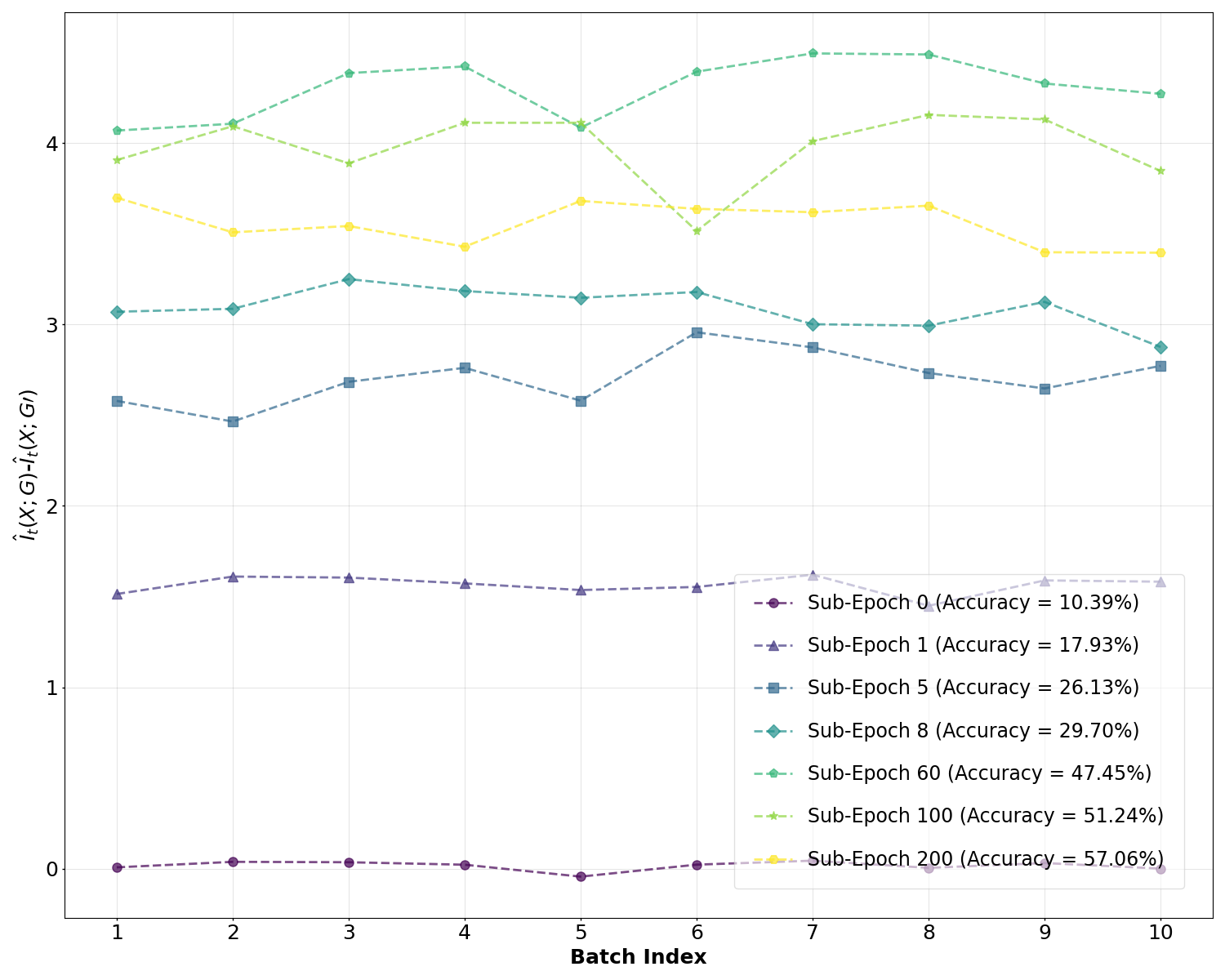}
		\label{diff-let-t}
    }
    \caption{Mutual Information Differences Estimation on LeNet (CIFAR-10).}
    \label{diff-MI-lenet}
\end{figure}

\begin{figure}[htp]
    \centering
    \vspace{-2mm}
    \subfigure[Autoencoder]{
		\includegraphics[width=0.42\textwidth]{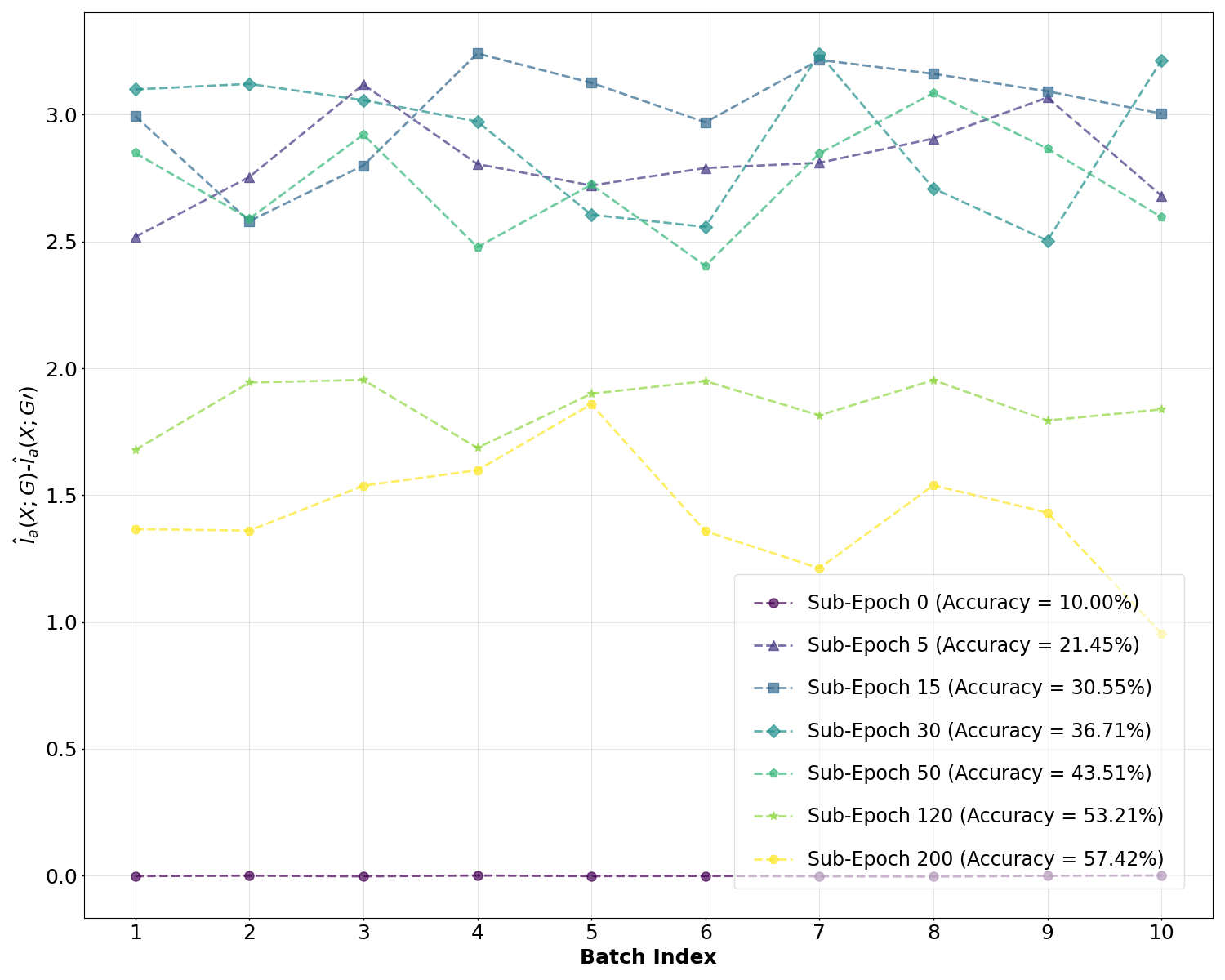}
		\label{diff-alex-a}
    }
    \subfigure[Transformer]{
		\includegraphics[width=0.42\textwidth]{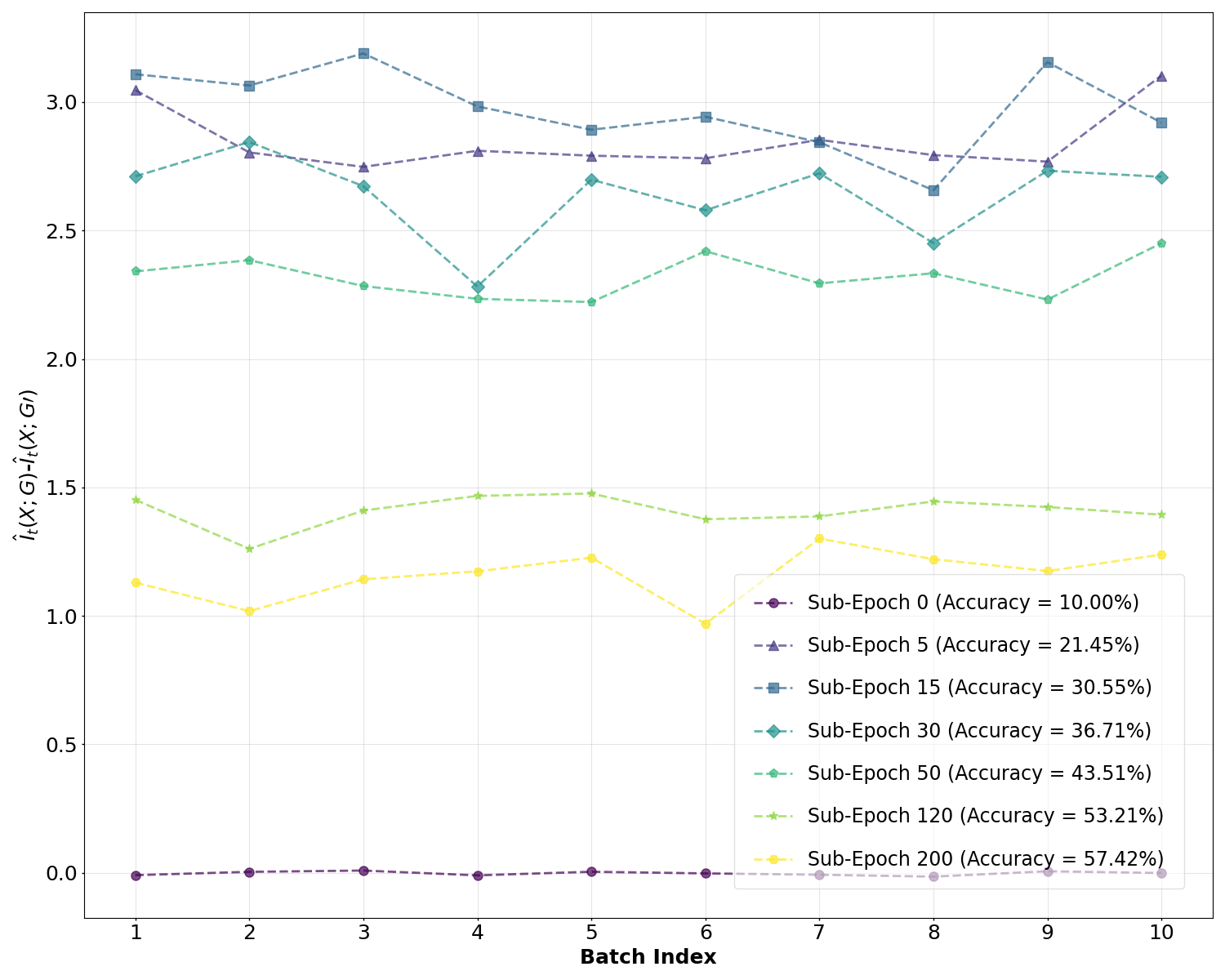}
		\label{diff-alex-t}
    }
    \caption{Mutual Information Differences Estimation on AlexNet (CIFAR-10).}
    \label{diff-MI-alexnet}
\end{figure}

\begin{figure}[htp]
    \centering
    \vspace{-2mm}
    \subfigure[Autoencoder]{
		\includegraphics[width=0.42\textwidth]{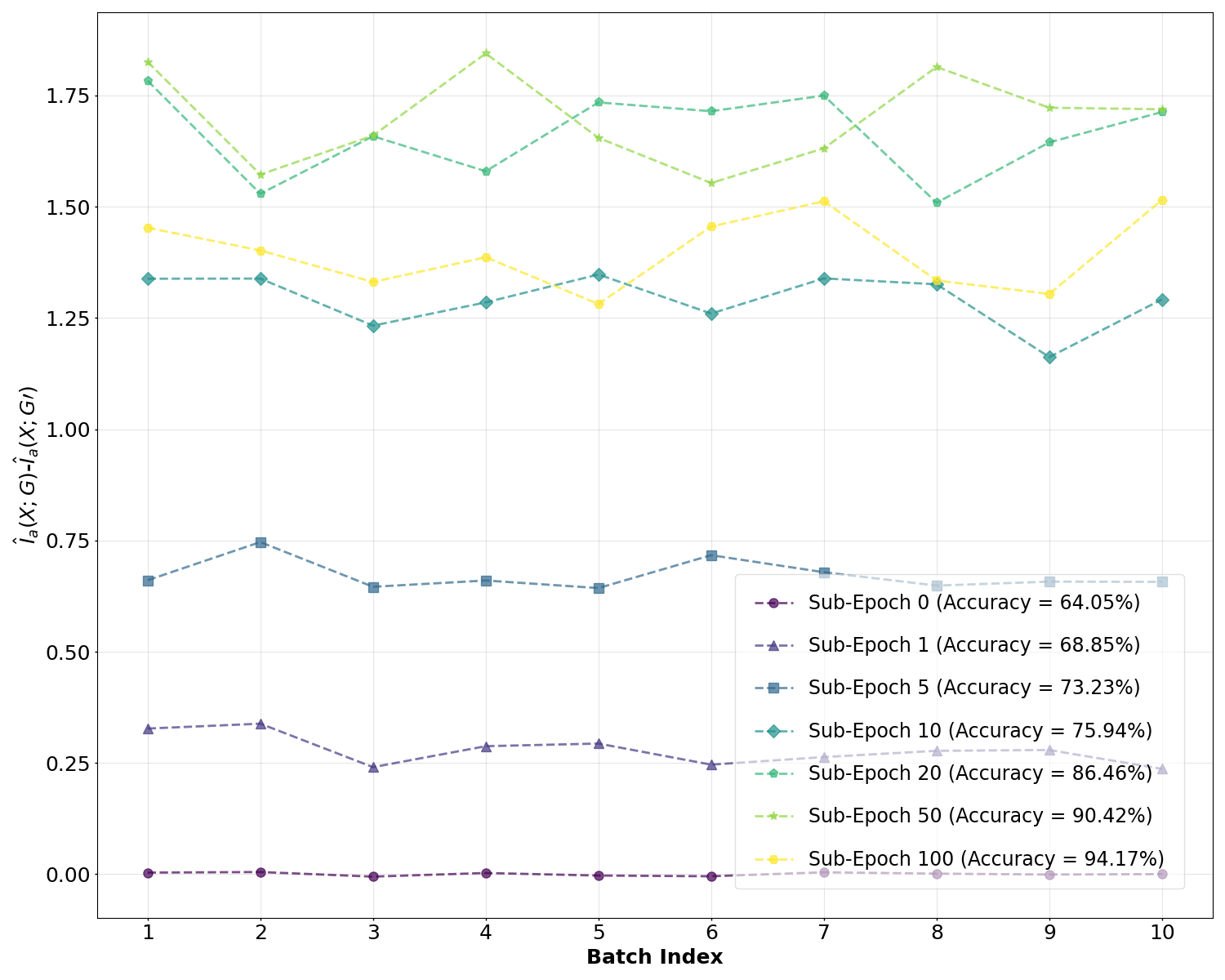}
		\label{diff-cnn-a}
    }
    \subfigure[Transformer]{
		\includegraphics[width=0.42\textwidth]{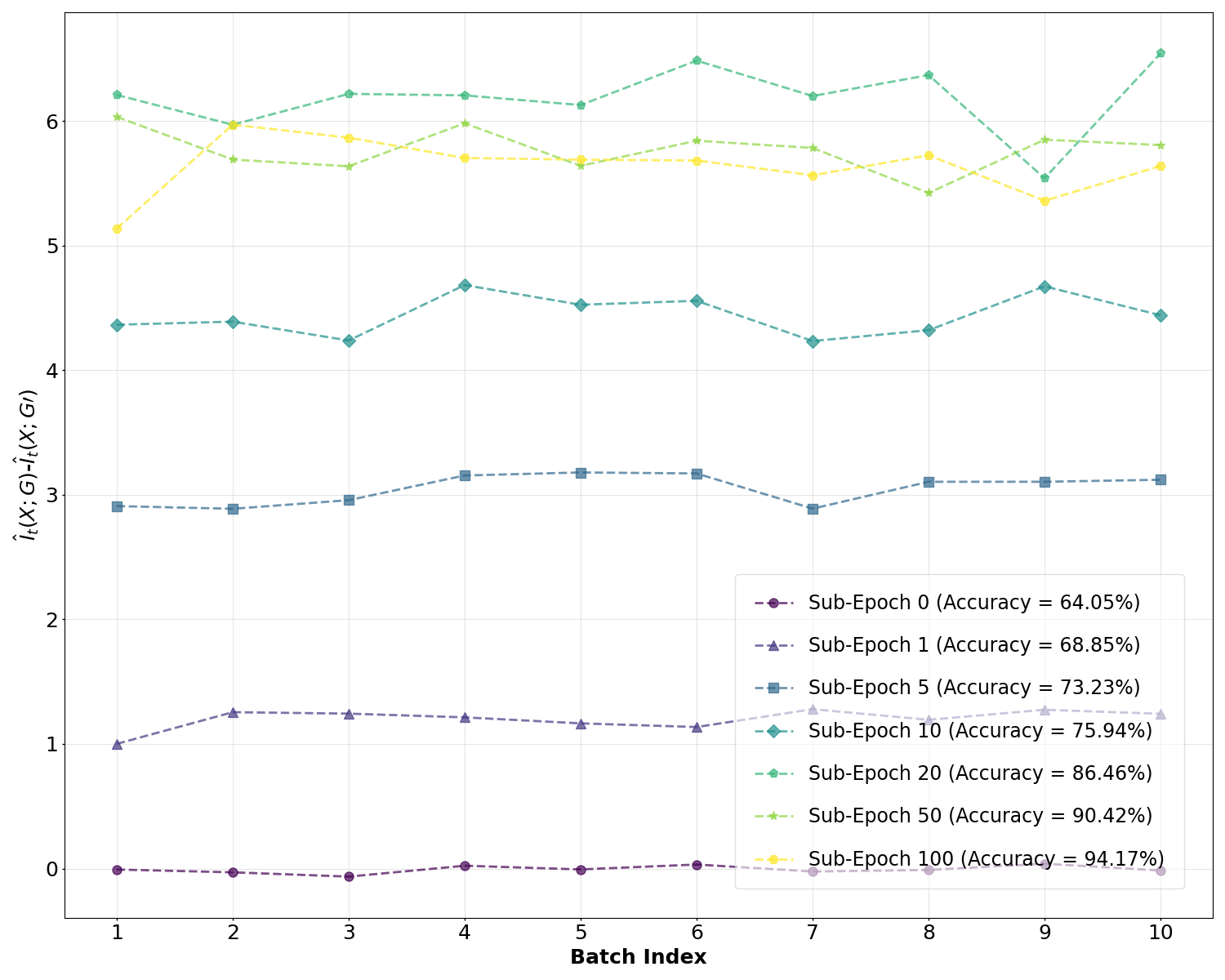}
		\label{diff-cnn-t}
    }
    \caption{Mutual Information Differences Estimation on CNN (CelebA-HQ).}
    \label{diff-MI-cnn}
\end{figure}

\begin{figure}[htp]
    \centering
    \vspace{-2mm}
    \subfigure[Autoencoder]{
		\includegraphics[width=0.42\textwidth]{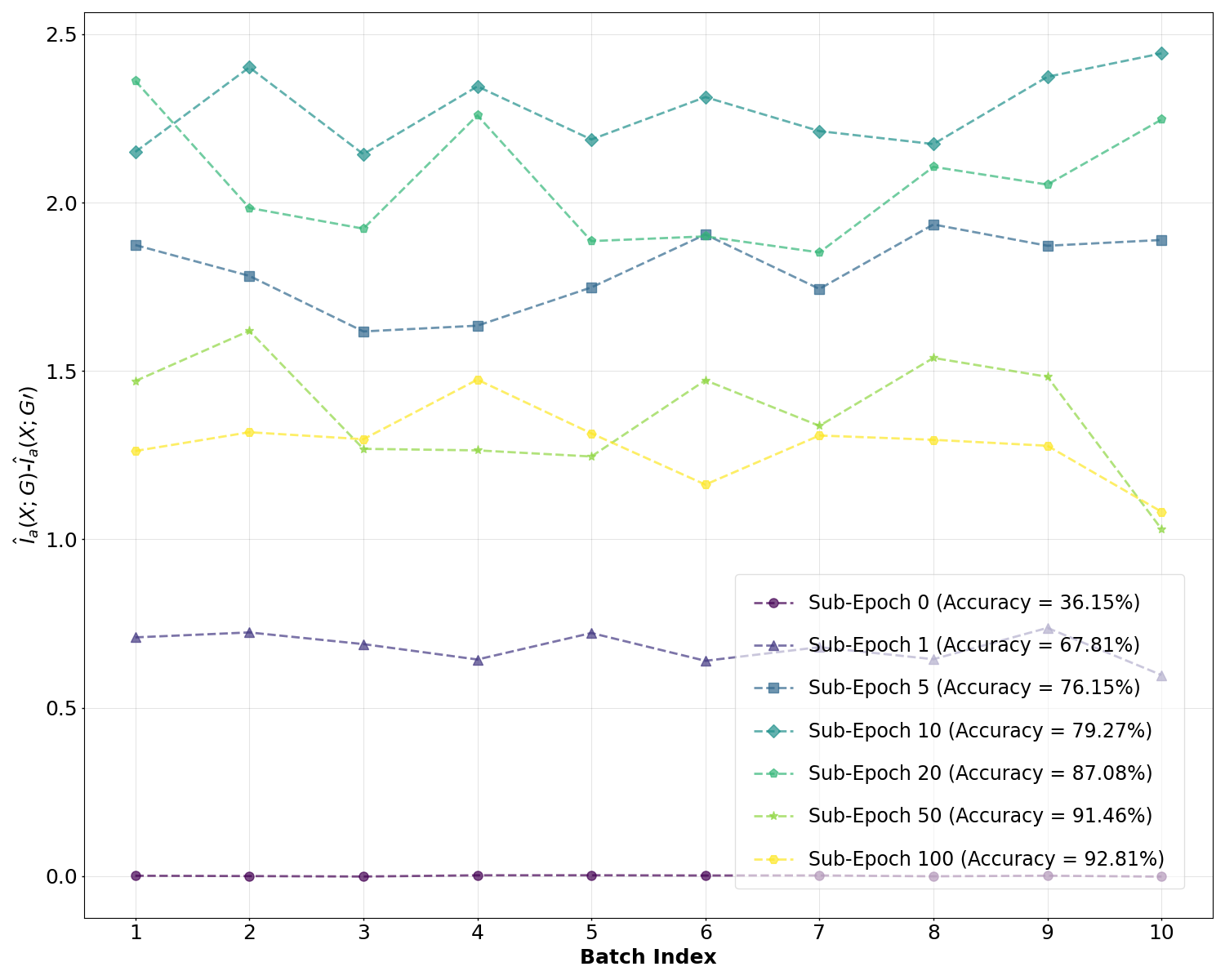}
		\label{diff-mlp-a}
    }
    \subfigure[Transformer]{
		\includegraphics[width=0.42\textwidth]{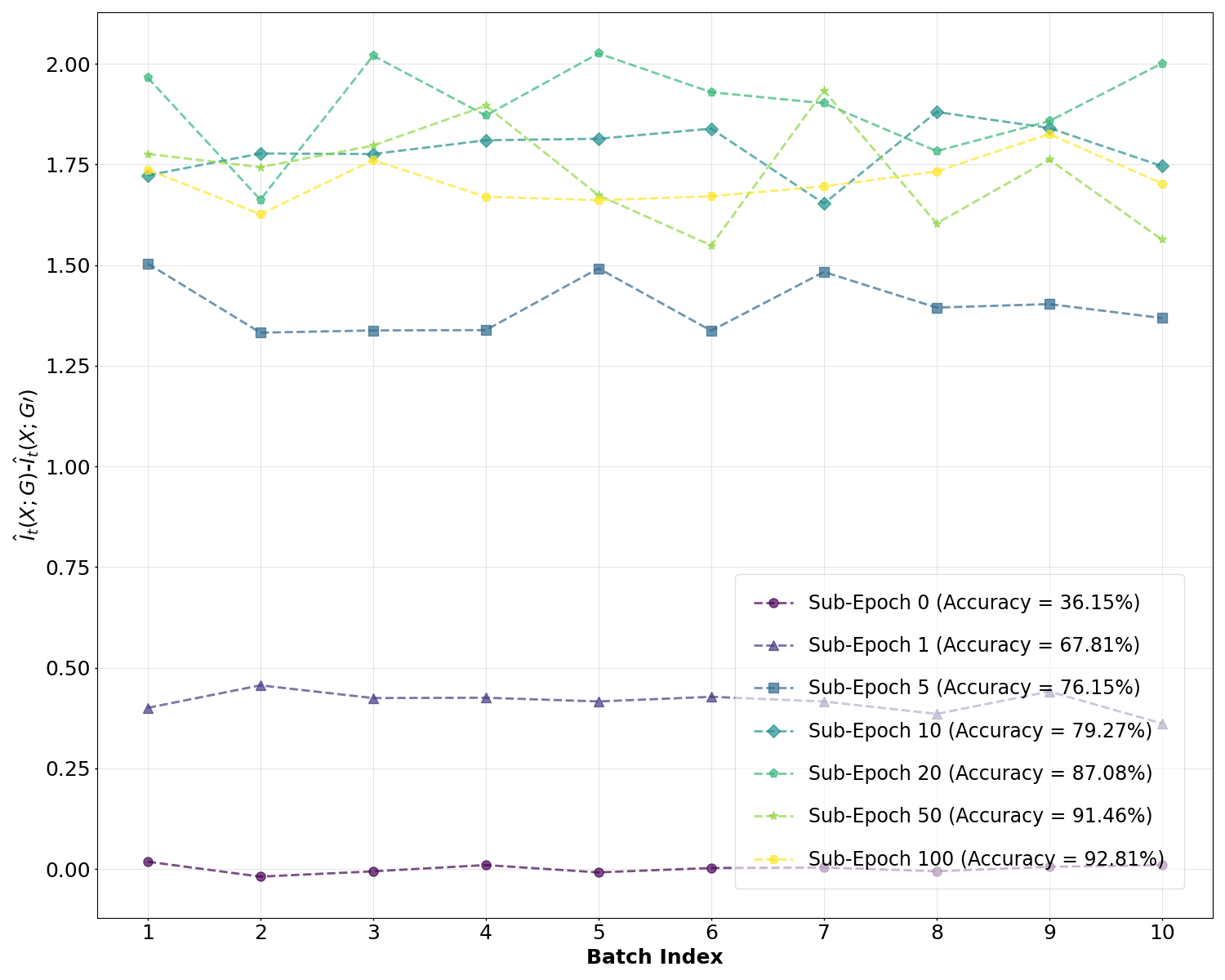}
		\label{diff-mlp-t}
    }
    \caption{Mutual Information Differences Estimation on MLP (CelebA-HQ).}
    \label{diff-MI-mlp}
\end{figure}

\begin{figure}[htp]
    \centering
    \vspace{-2mm}
    \subfigure[Autoencoder]{
		\includegraphics[width=0.42\textwidth]{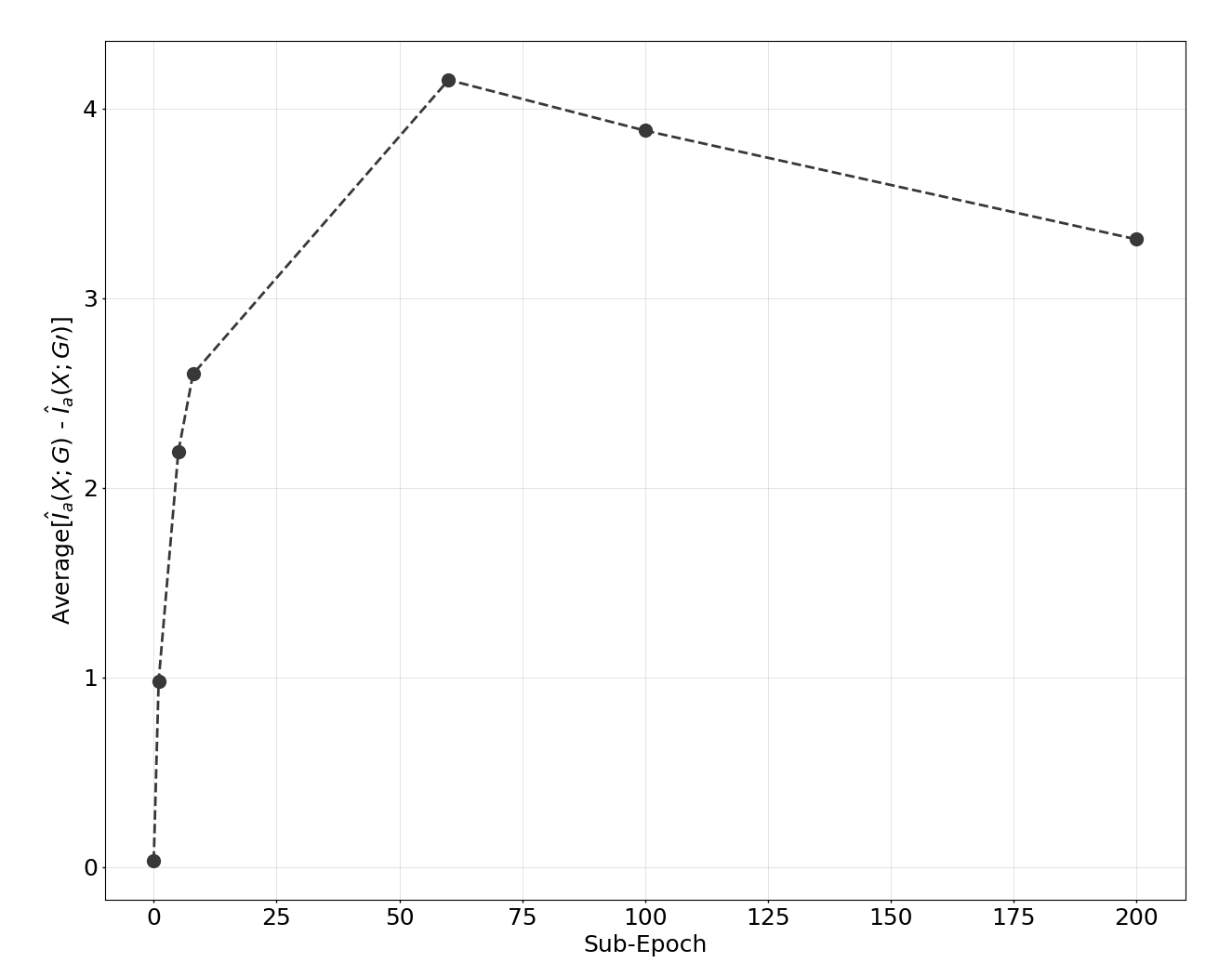}
		\label{diff-ave-let-a}
    }
    \subfigure[Transformer]{
		\includegraphics[width=0.42\textwidth]{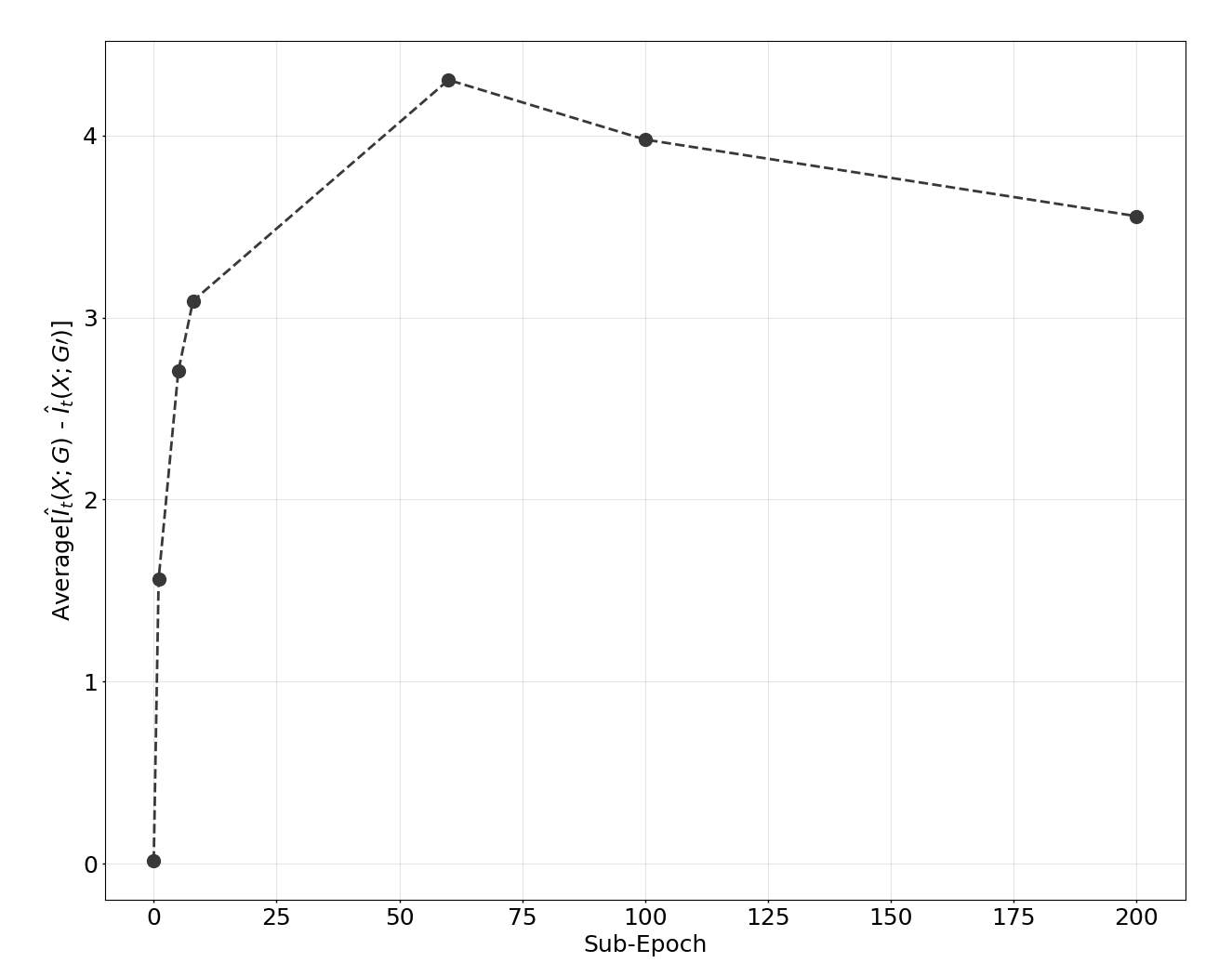}
		\label{diff-ave-let-t}
    }
    \caption{Average Mutual Information Differences Trends on LeNet (CIFAR-10).}
    \label{diff-ave-MI-lenet}
\end{figure}

\begin{figure}[htp]
    \centering
    \vspace{-2mm}
    \subfigure[Autoencoder]{
		\includegraphics[width=0.42\textwidth]{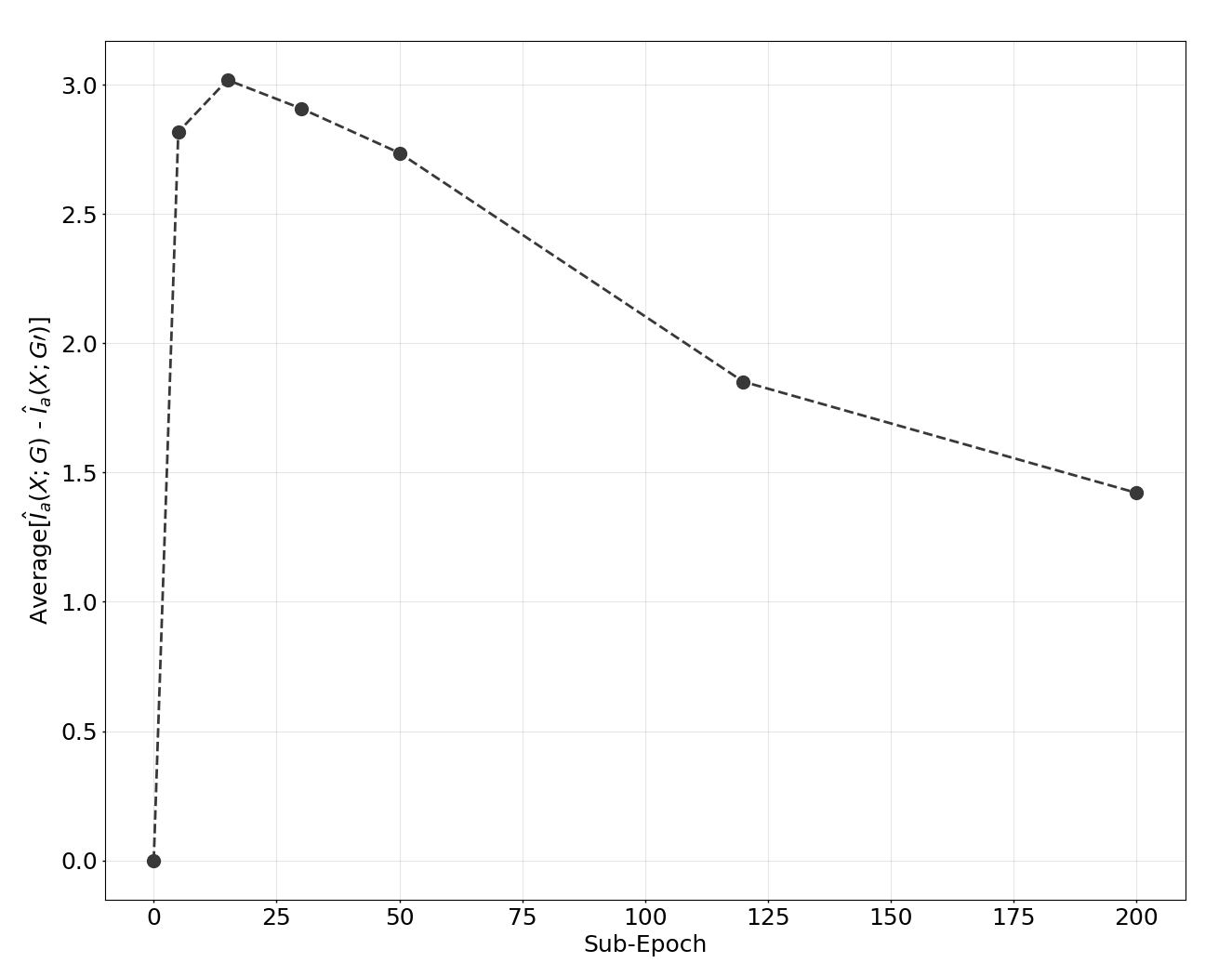}
		\label{diff-ave-alex-a}
    }
    \subfigure[Transformer]{
		\includegraphics[width=0.42\textwidth]{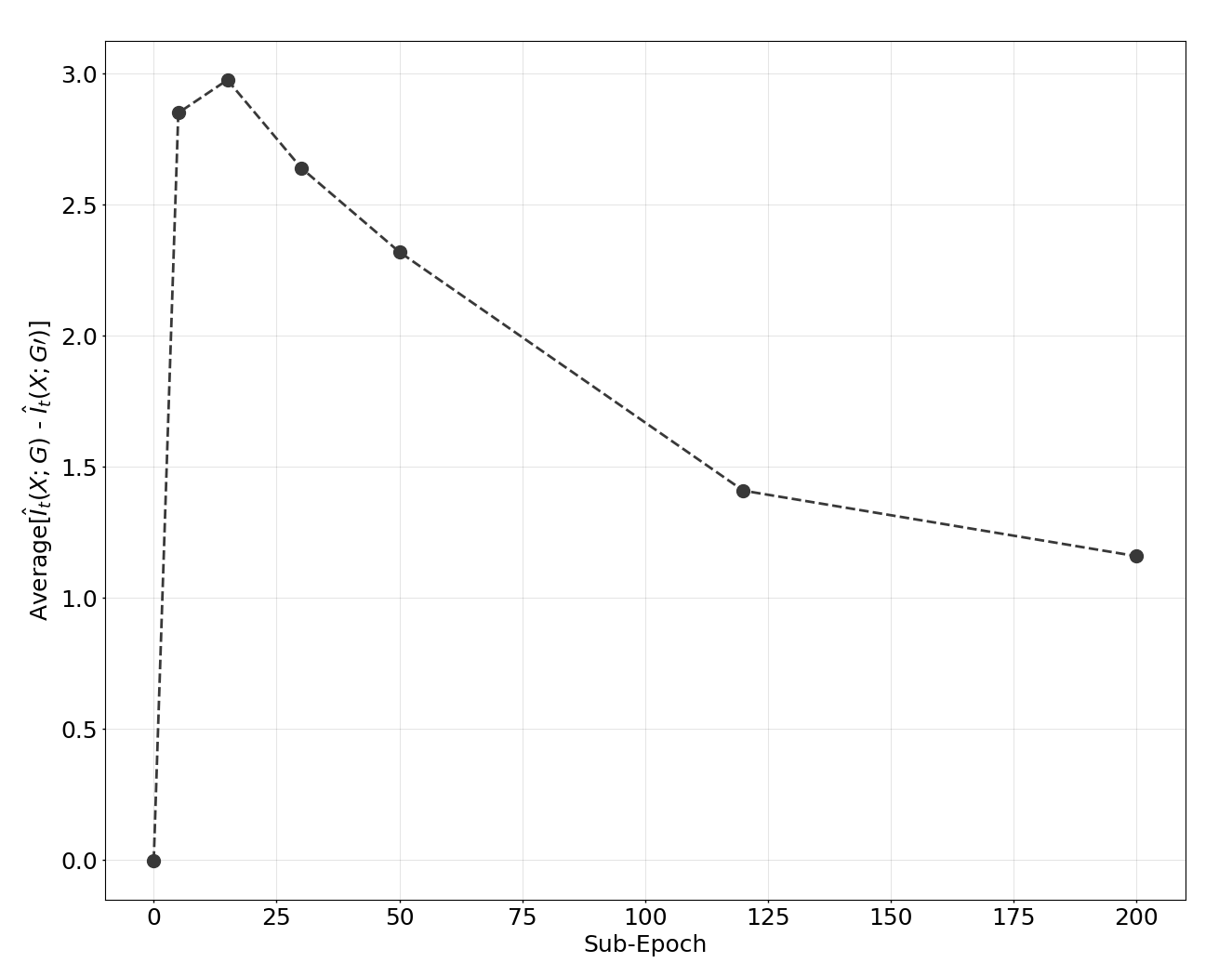}
		\label{diff-ave-alex-t}
    }
    \caption{Average Mutual Information Differences Trends on AlexNet (CIFAR-10).}
    \label{diff-ave-MI-alexnet}
\end{figure}

\begin{figure}[htp]
    \centering
    \vspace{-2mm}
    \subfigure[Autoencoder]{
		\includegraphics[width=0.42\textwidth]{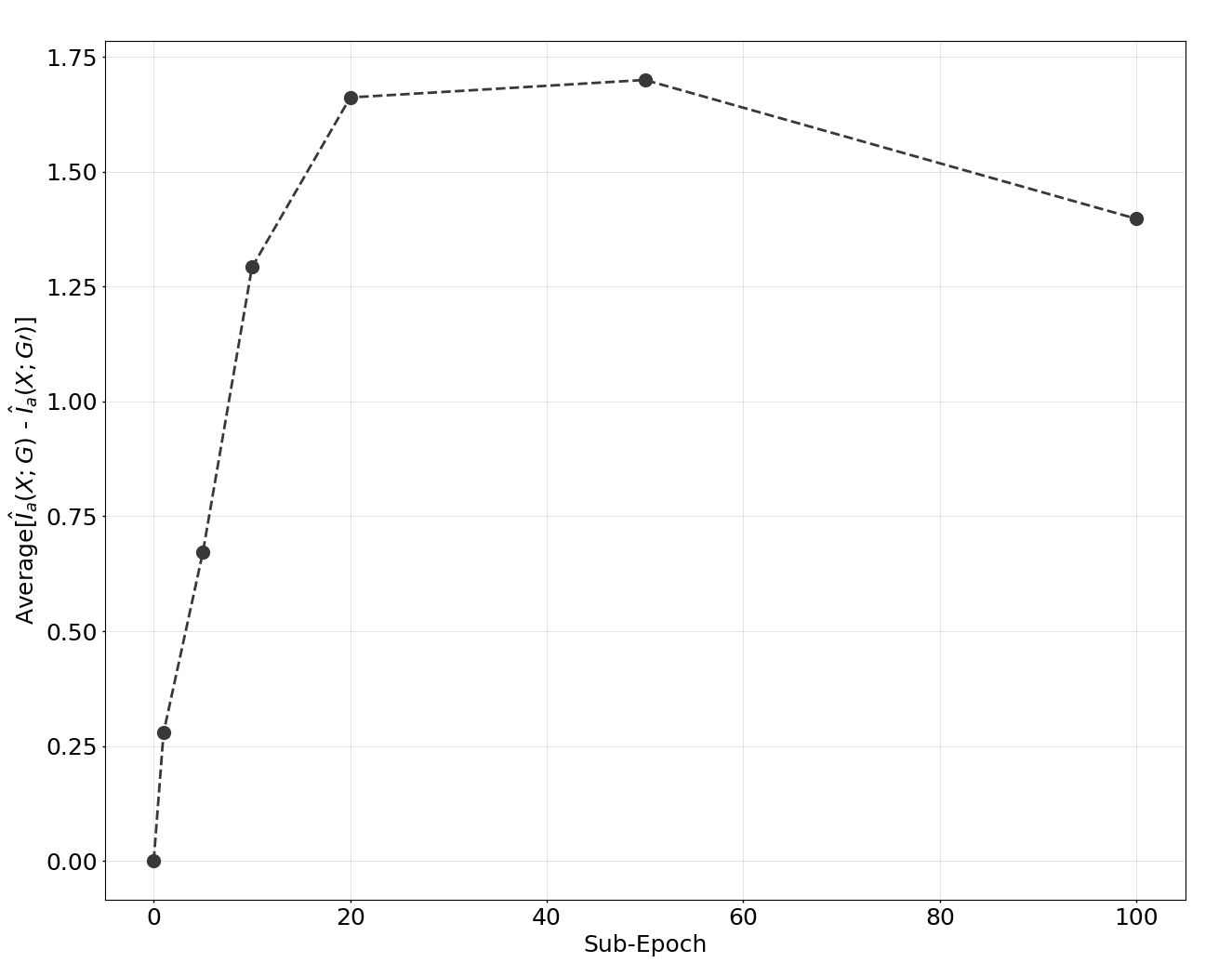}
		\label{diff-ave-cnn-a}
    }
    \subfigure[Transformer]{
		\includegraphics[width=0.42\textwidth]{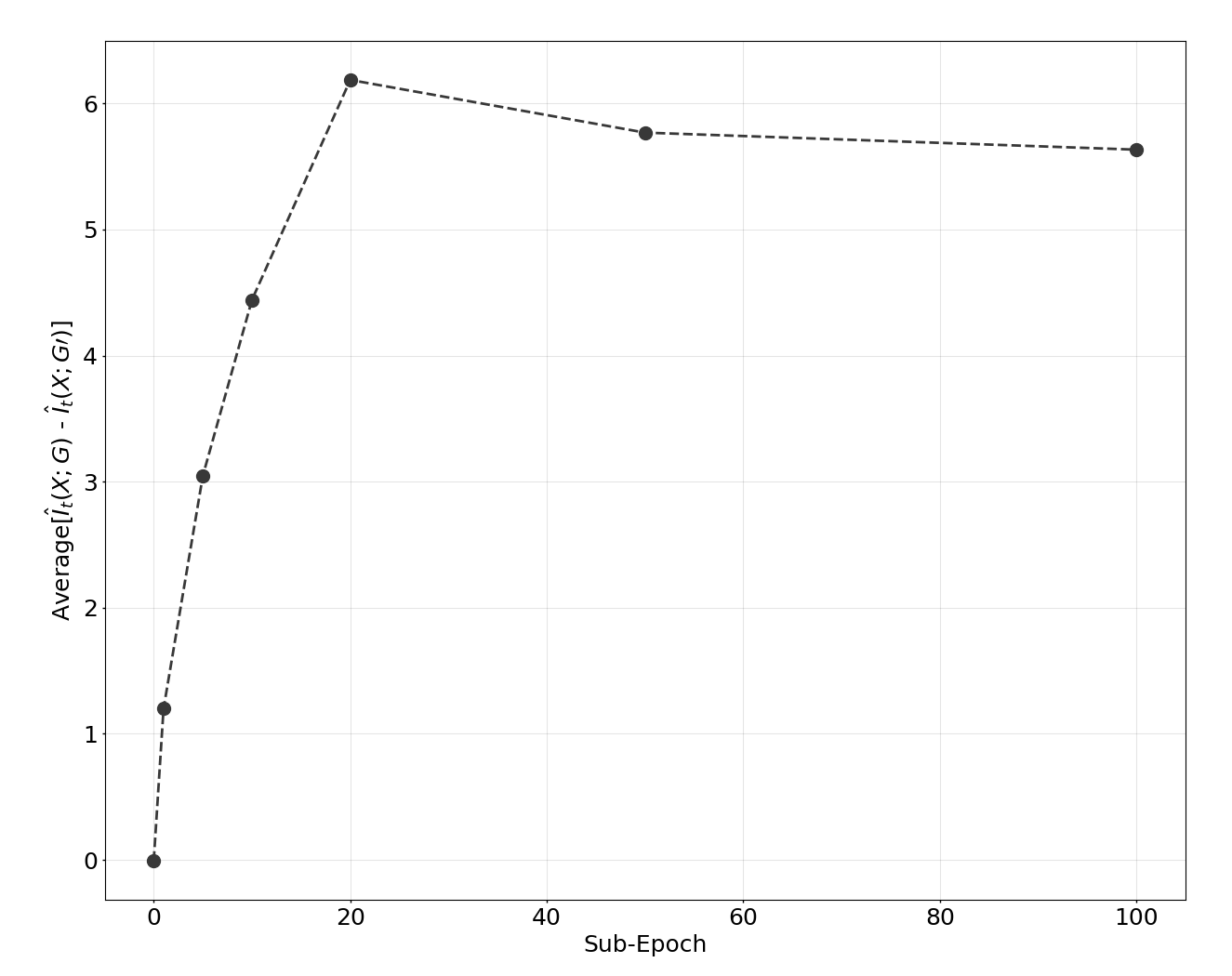}
		\label{diff-ave-cnn-t}
    }
    \caption{Average Mutual Information Differences Trends on CNN (CelebA-HQ).}
    \label{diff-ave-MI-cnn}
\end{figure}

\begin{figure}[htp]
    \centering
    \vspace{-2mm}
    \subfigure[Autoencoder]{
		\includegraphics[width=0.42\textwidth]{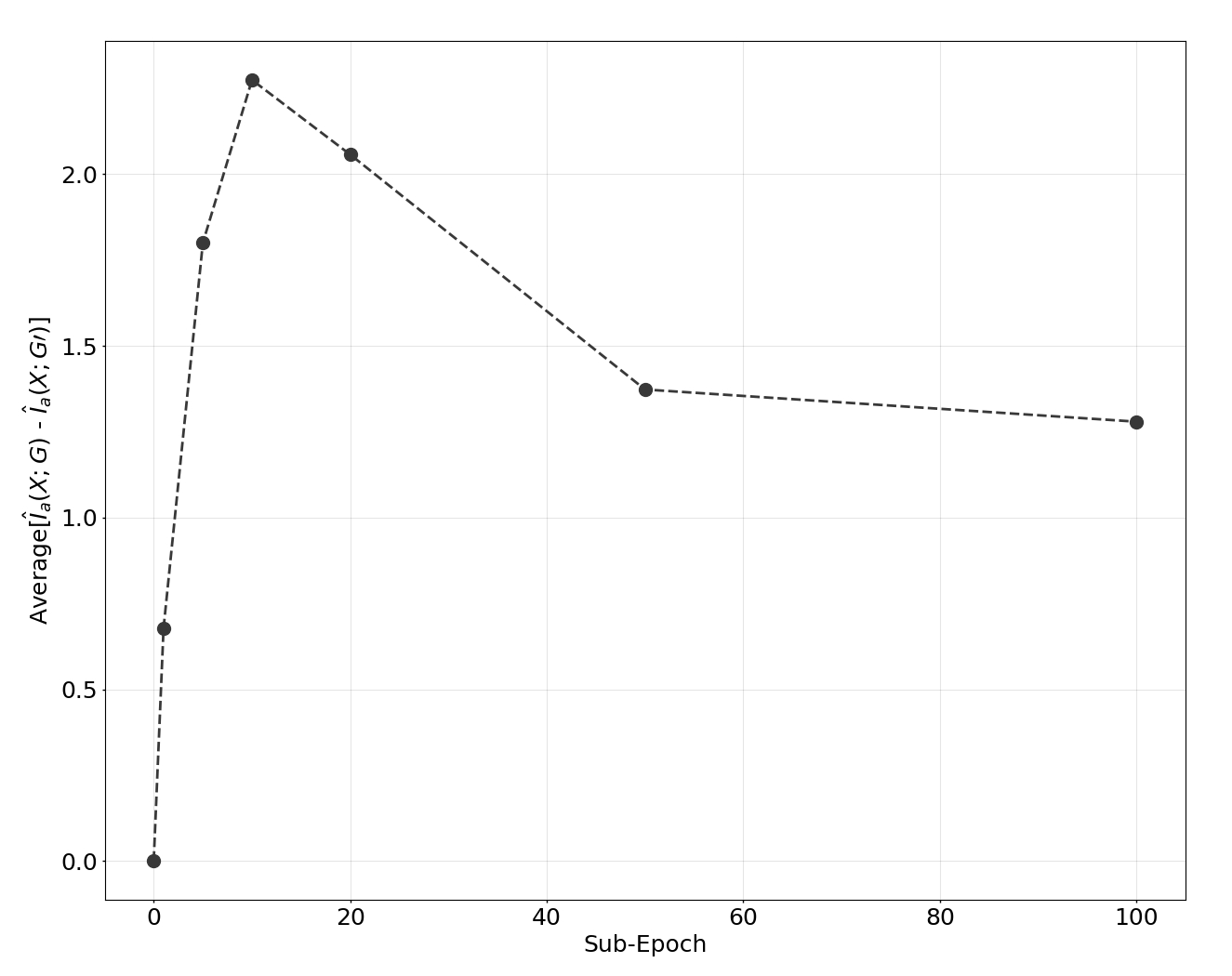}
		\label{diff-ave-mlp-a}
    }
    \subfigure[Transformer]{
		\includegraphics[width=0.42\textwidth]{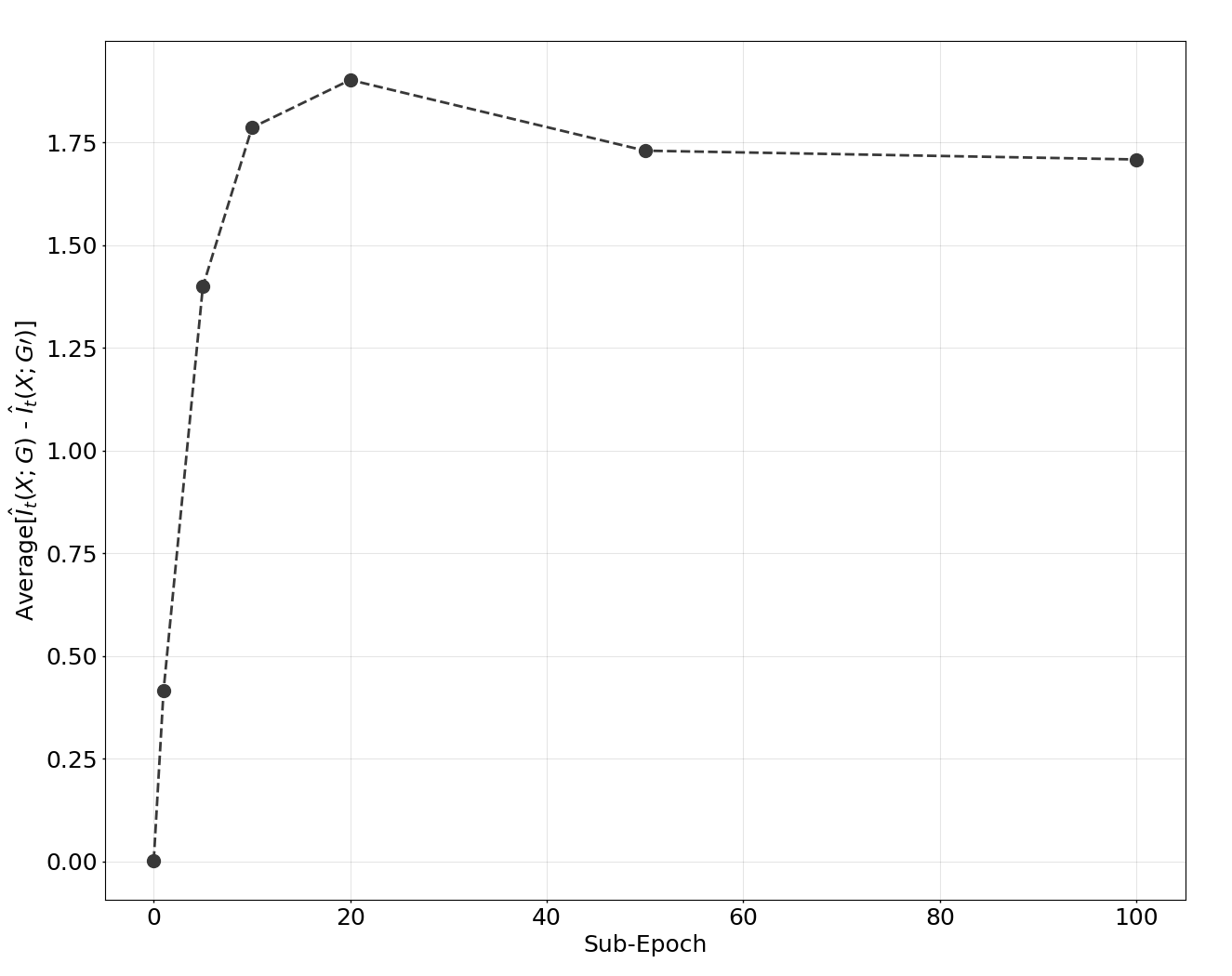}
		\label{diff-ave-mlp-t}
    }
    \caption{Average Mutual Information Differences Trends on MLP (CelebA-HQ).}
    \label{diff-ave-MI-mlp}
\end{figure}

Additionally, we examine the difference in mutual information across all batches at various sub-epochs and record the corresponding test accuracy. The results are shown in Figures \ref{diff-MI-lenet}-\ref{diff-MI-mlp} and summarized in Table \ref{attack}. Figures \ref{diff-ave-MI-lenet}-\ref{diff-ave-MI-mlp} illustrate the evolution of the average mutual information difference over sub-epochs. It is evident that the mutual information difference initially increases and then decreases. A more detailed analysis of these trends is provided in Section \ref{sec:exp}.

The labels "Autoencoder" or "Transformer" in the figures refer to the privacy tokens used for mutual information estimation, which are derived from either the Autoencoder or Transformer.

\newpage

\section{Ablation Study}\label{ablation_study}

\begin{figure}[htp]
    \centering
    \vspace{-2mm}
    \subfigure[Autoencoder]{
		\includegraphics[width=0.48\textwidth]{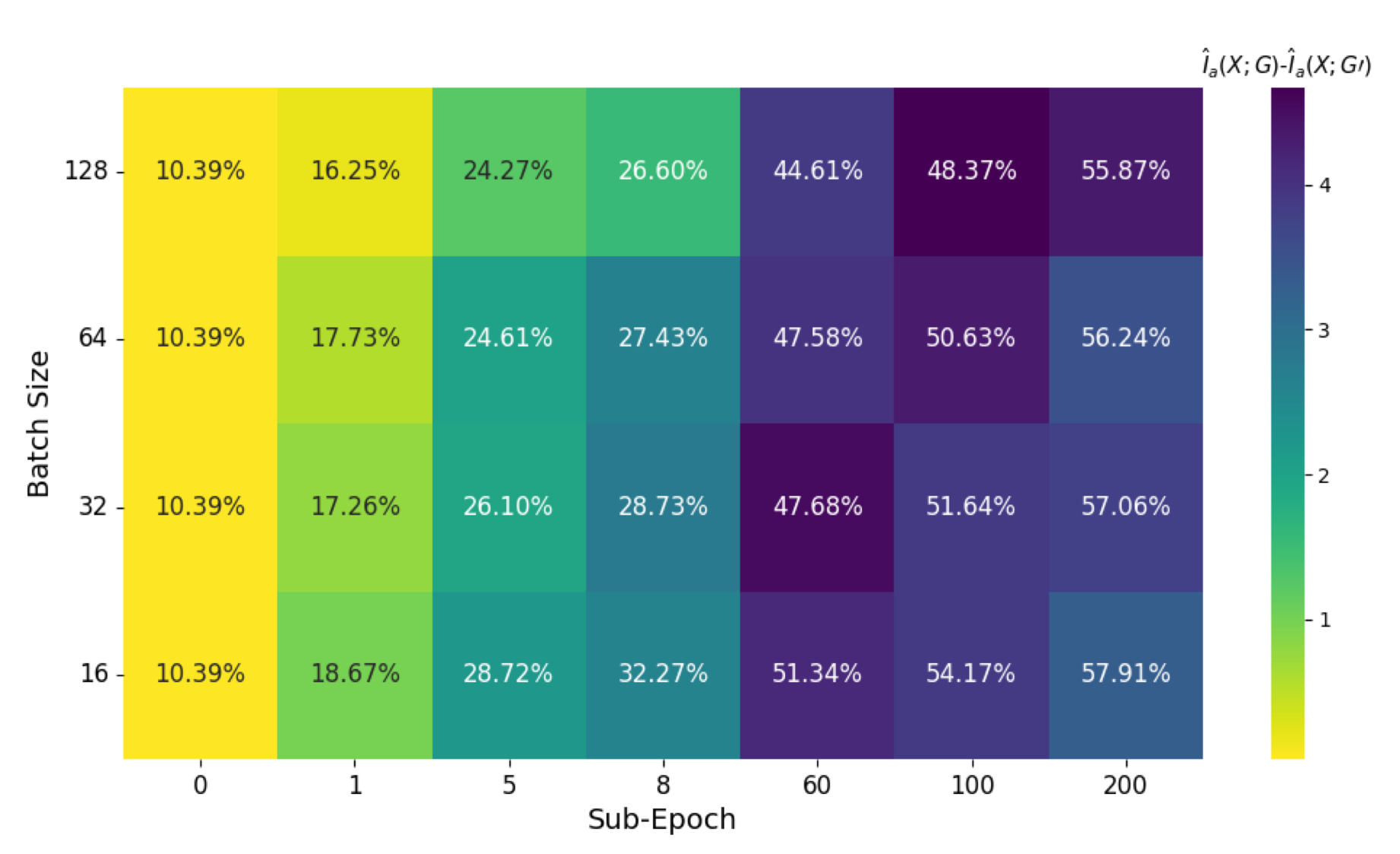}
		\label{heatmap_lenet_auto}
    }
    \subfigure[Transformer]{
		\includegraphics[width=0.48\textwidth]{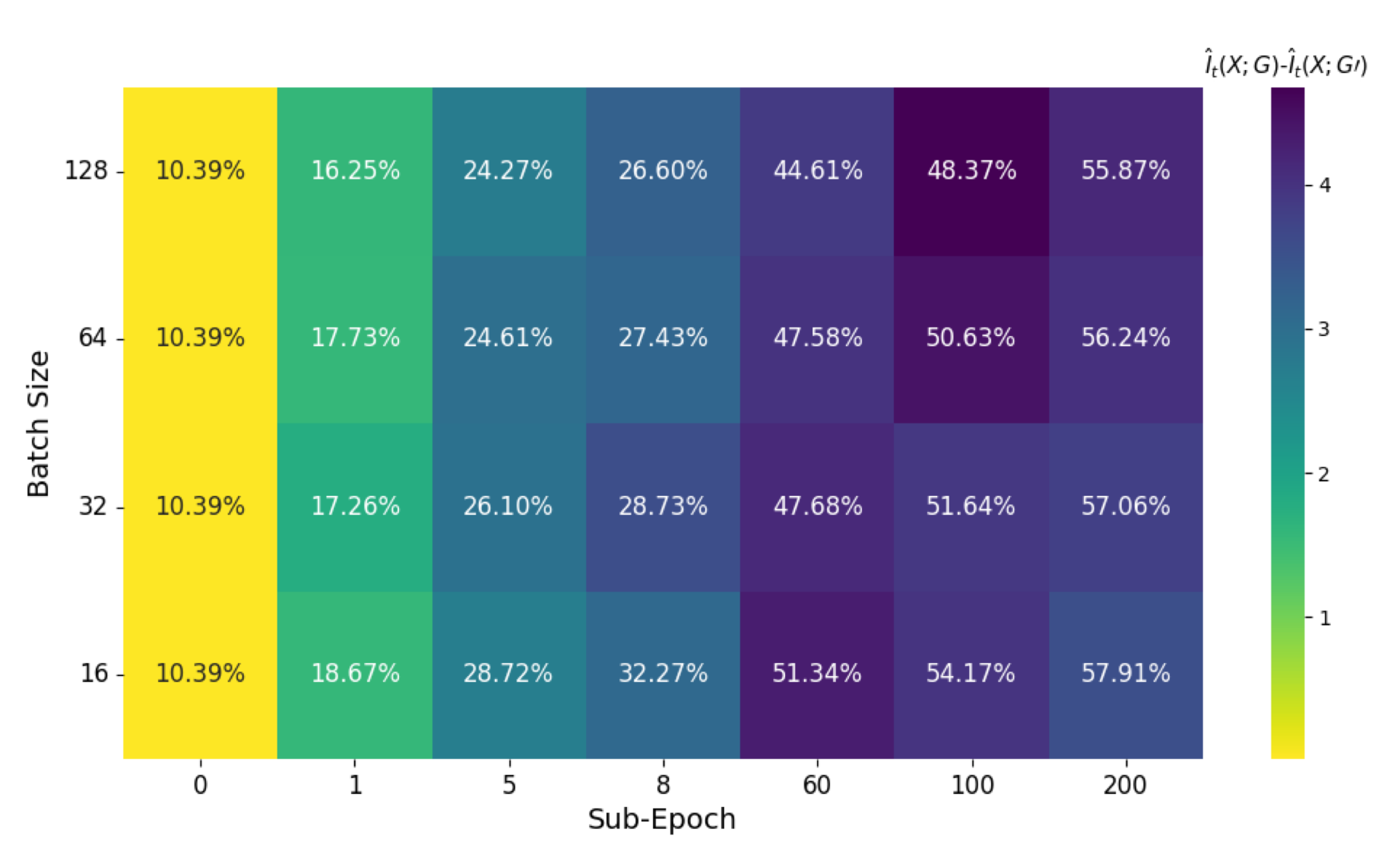}
		\label{heatmap_lenet_trans}
    }
    \caption{Relationship between Mutual Information Difference and Batch Size on LeNet (CIFAR-10).}
    \label{heatmap_lenet}
\end{figure}

\begin{figure}[htp]
    \centering
    \vspace{-2mm}
    \subfigure[Autoencoder]{
		\includegraphics[width=0.48\textwidth]{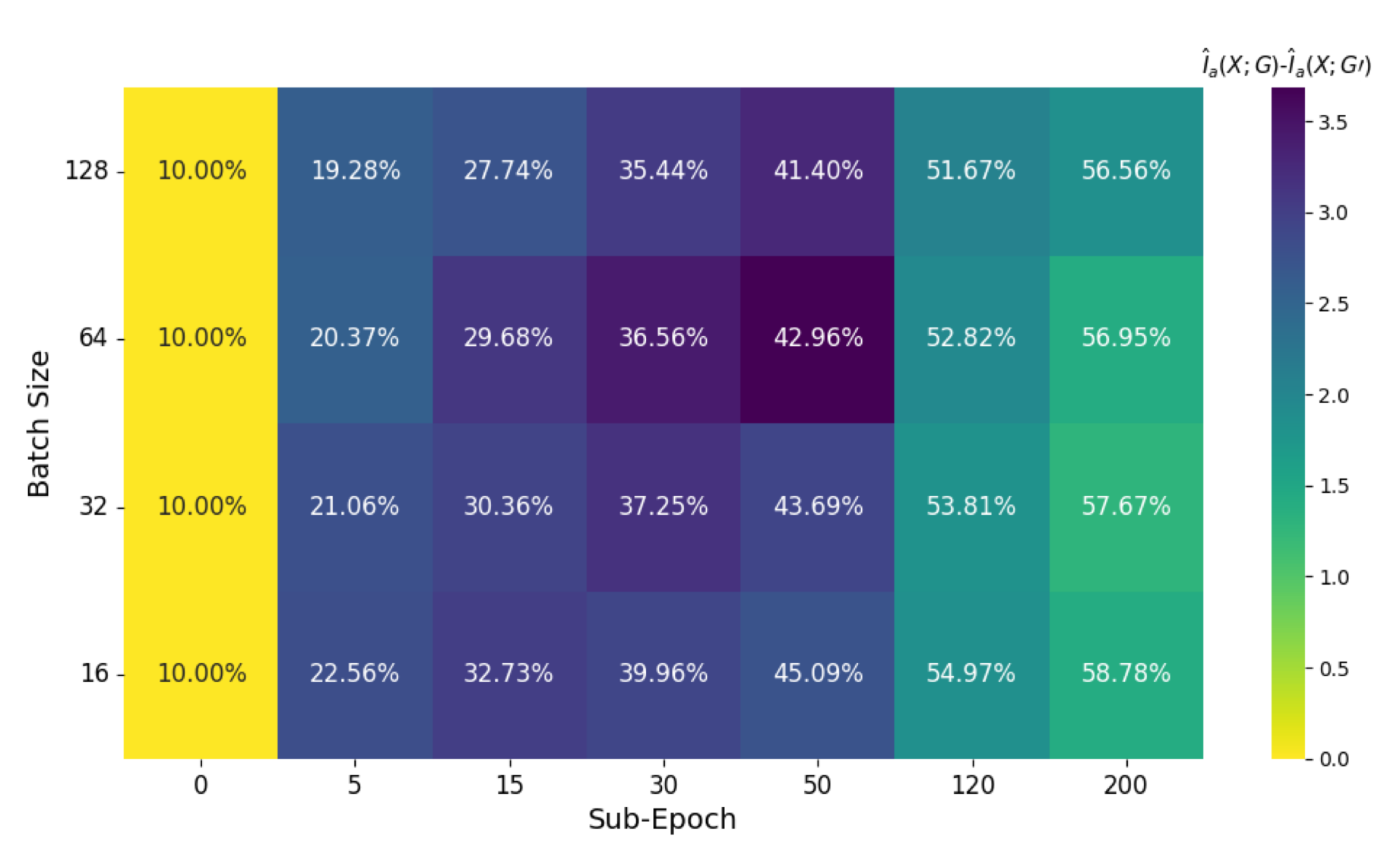}
		\label{heatmap_alex_auto}
    }
    \subfigure[Transformer]{
		\includegraphics[width=0.48\textwidth]{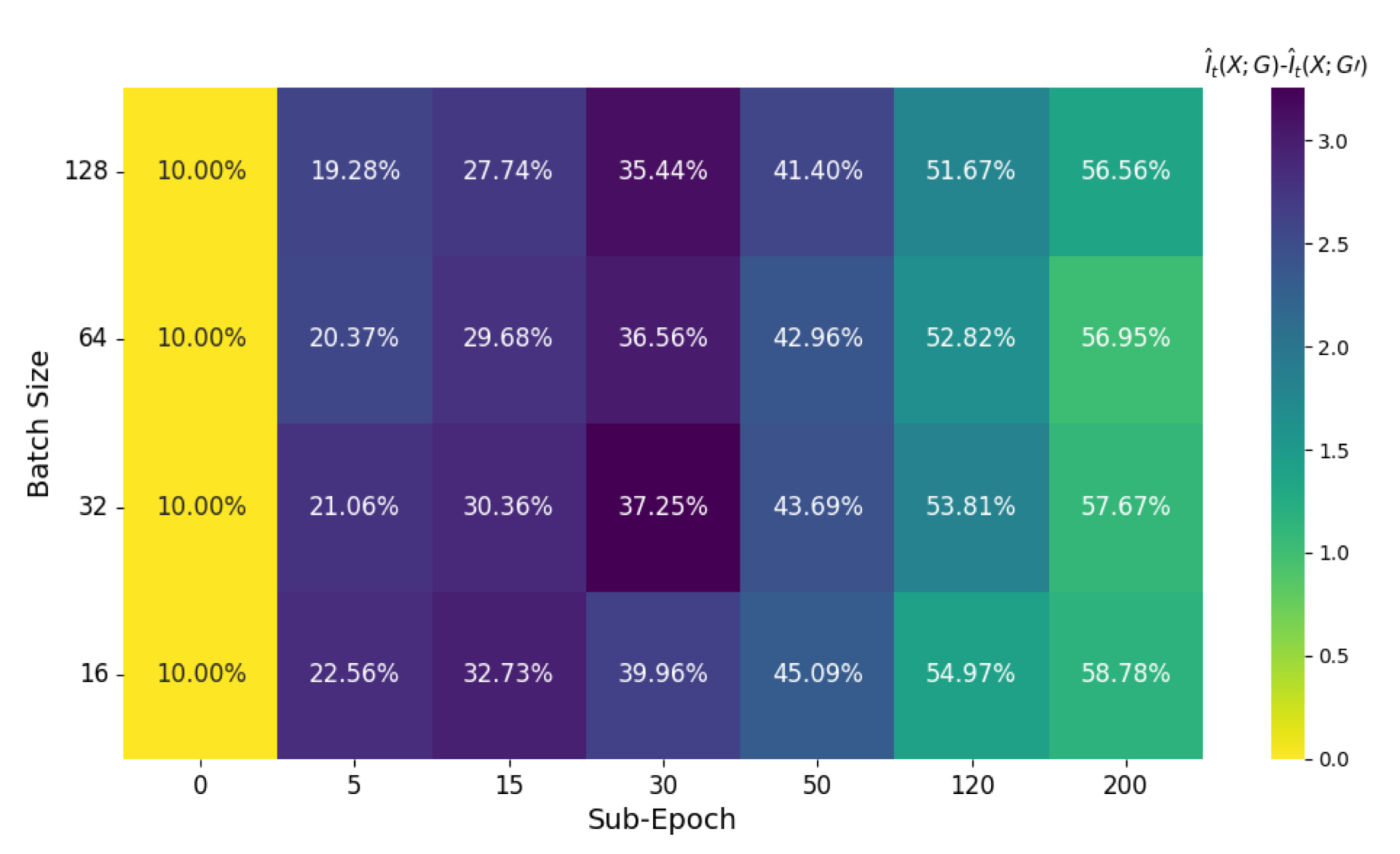}
		\label{heatmap_alex_trans}
    }
    \caption{Relationship between Mutual Information Difference and Batch Size on AlexNet (CIFAR-10).}
    \label{heatmap_alexnet}
\end{figure}

\begin{figure}[htp]
    \centering
    \vspace{-2mm}
    \subfigure[Autoencoder]{
		\includegraphics[width=0.48\textwidth]{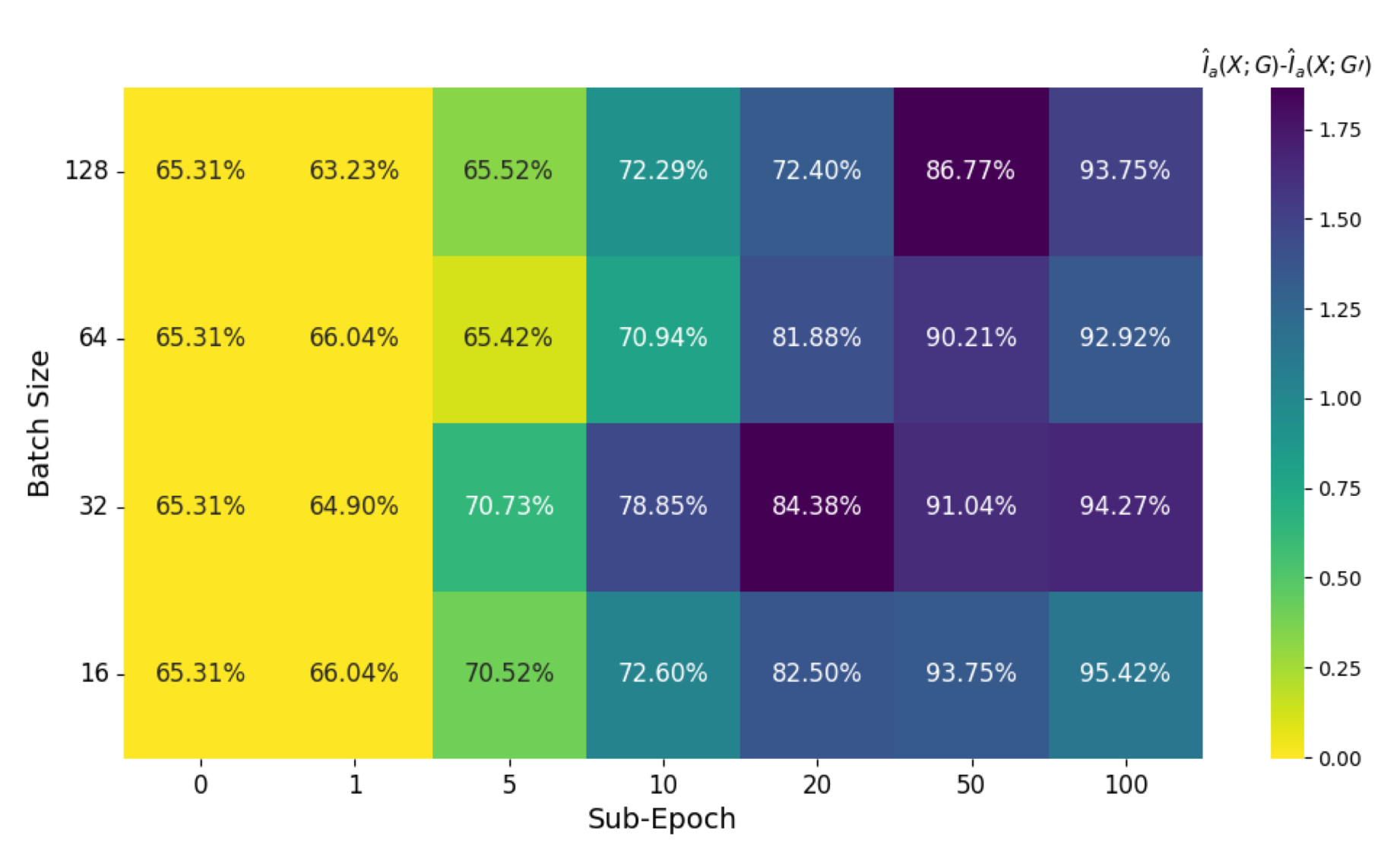}
		\label{heatmap_cnn_auto}
    }
    \subfigure[Transformer]{
		\includegraphics[width=0.48\textwidth]{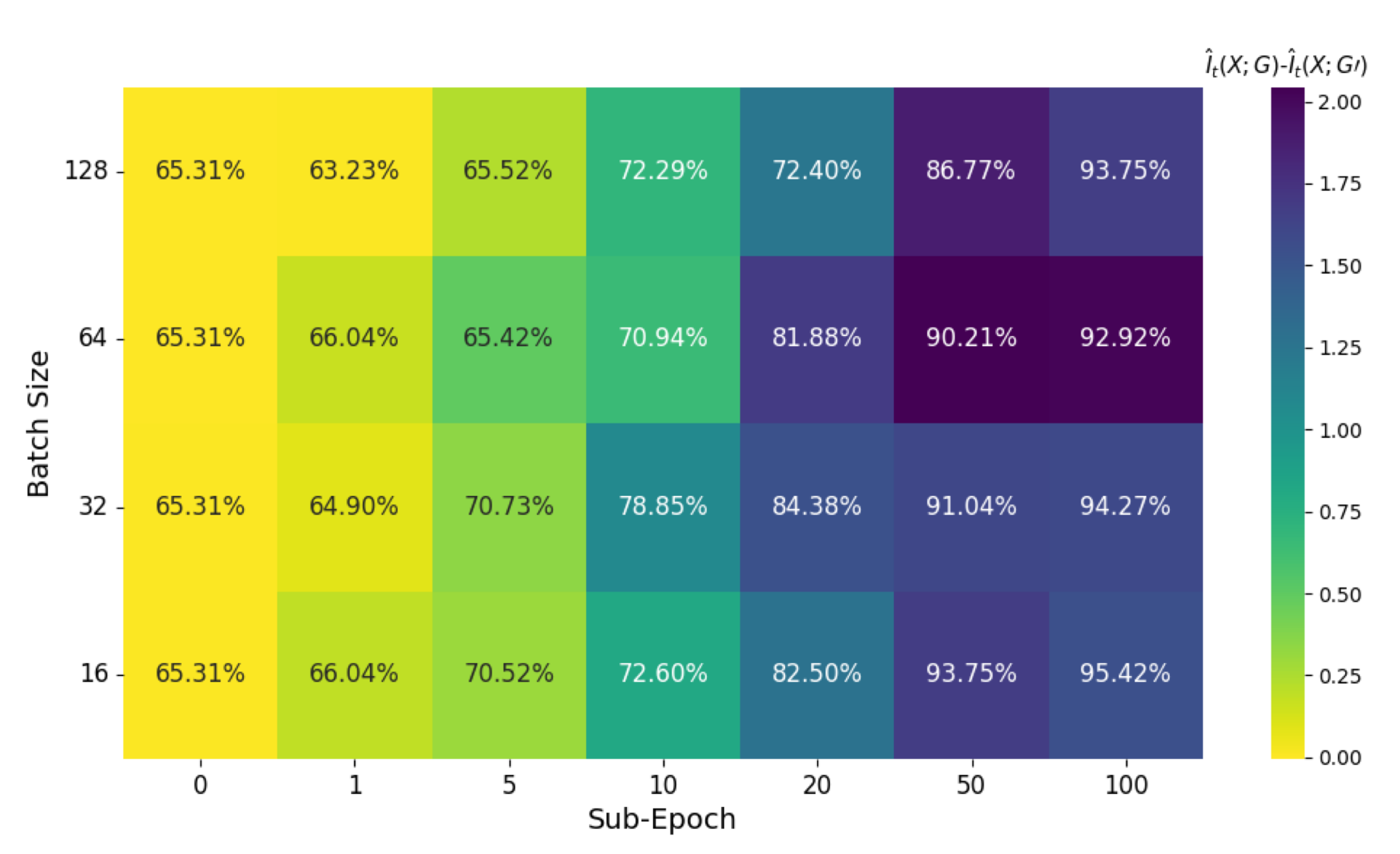}
		\label{heatmap_cnn_trans}
    }
    \caption{Relationship between Mutual Information Difference and Batch Size on CNN (CelebA-HQ).}
    \label{heatmap_cnn}
\end{figure}

\begin{figure}[htp]
    \centering
    \vspace{-2mm}
    \subfigure[Autoencoder]{
		\includegraphics[width=0.48\textwidth]{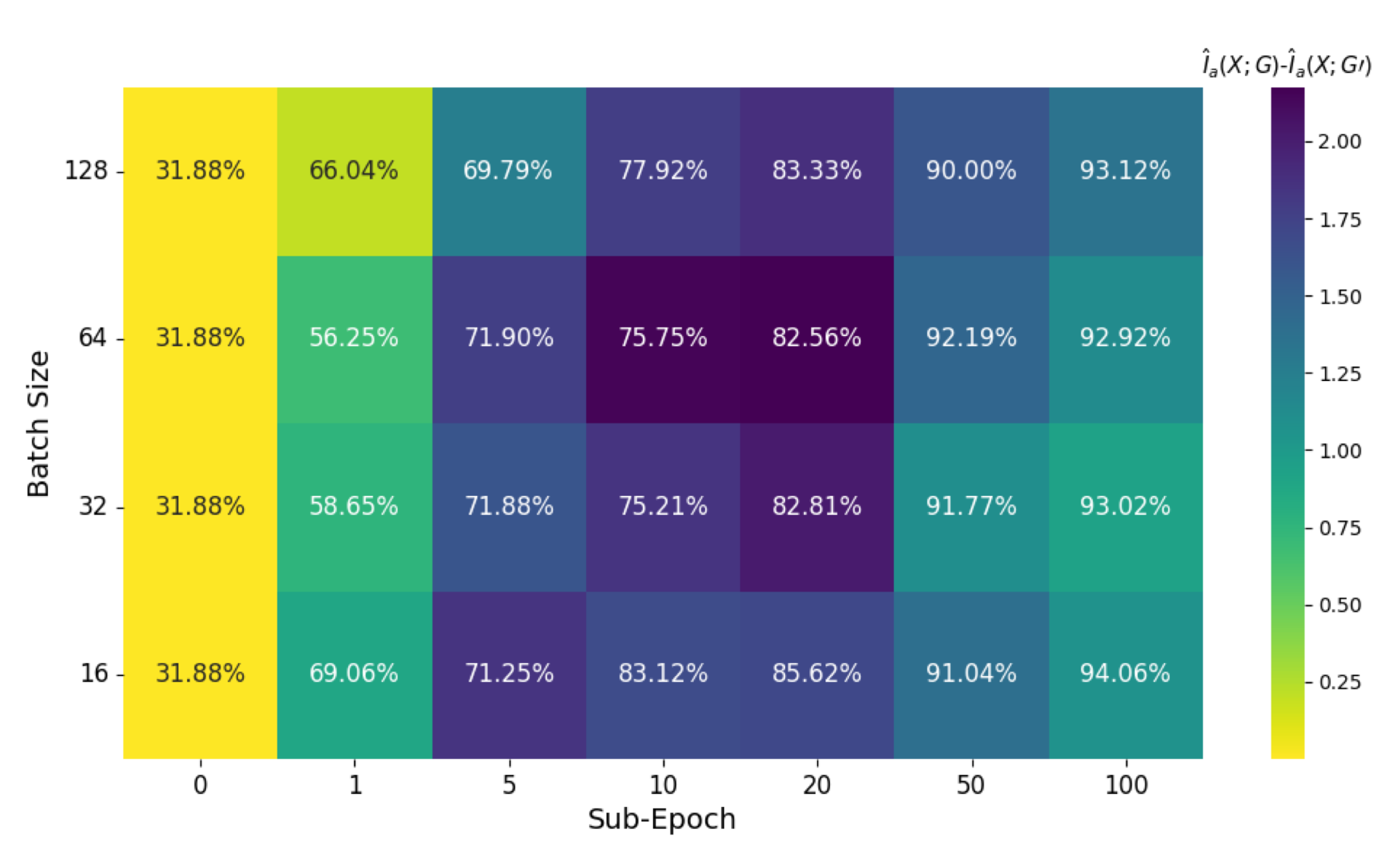}
		\label{heatmap_mlp_auto}
    }
    \subfigure[Transformer]{
		\includegraphics[width=0.48\textwidth]{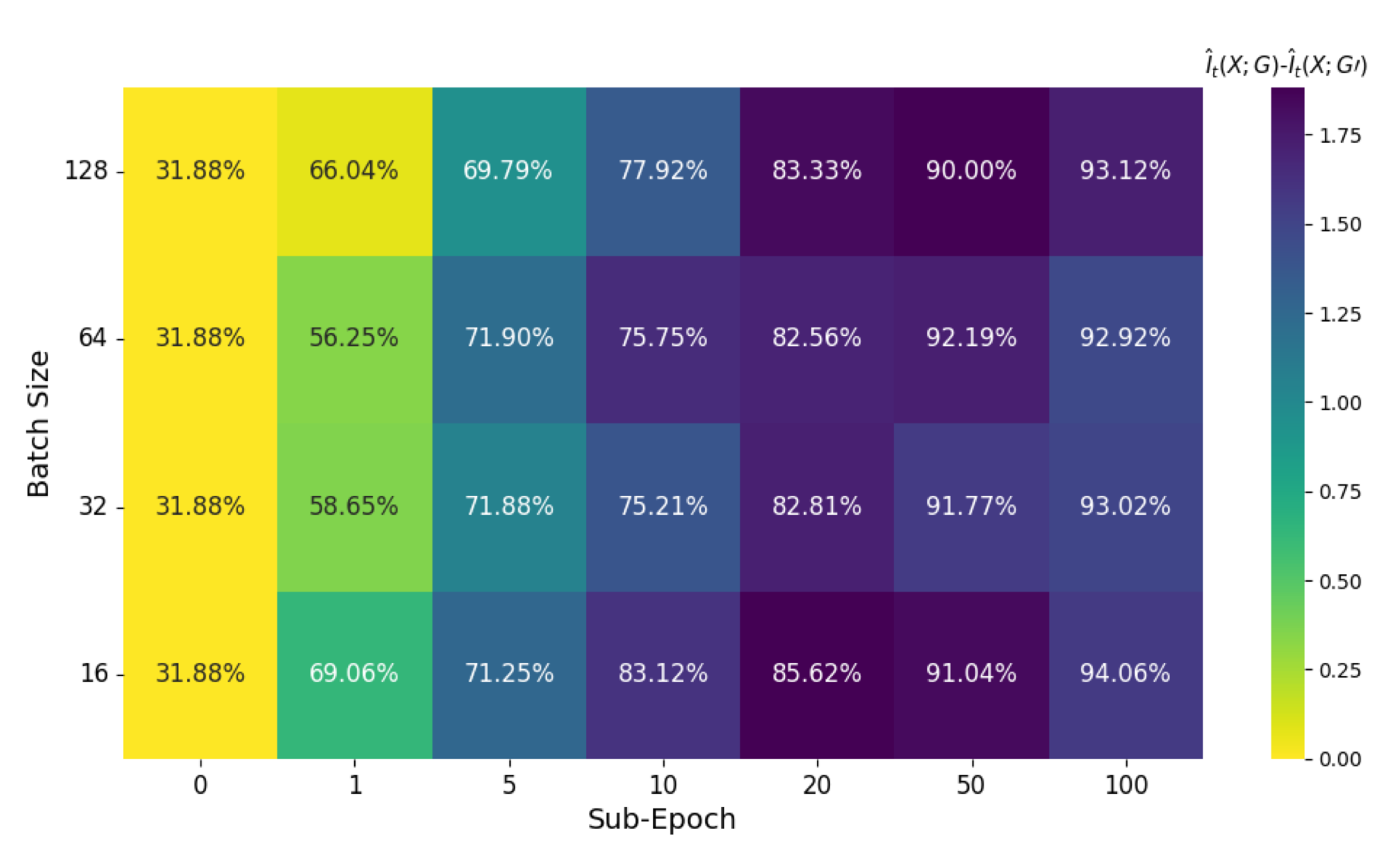}
		\label{heatmap_mlp_trans}
    }
    \caption{Relationship between Mutual Information Difference and Batch Size on MLP (CelebA-HQ).}
    \label{heatmap_mlp}
\end{figure}

The purpose of the ablation study is grounded in the hypothesis that varying batch sizes may influence model performance. The batch size during training significantly affects the model's training process, influencing factors such as test accuracy, convergence speed, and so on. Generally, models that achieve higher test accuracy are considered to have stronger feature extraction capabilities, which contributes to better performance. In Section \ref{experiment_section} of the paper, we observed that the degree of private information leaked by the model varies at different stages of training. A high-performing model, which excels at extracting features for a specific task but has not yet converged, may inadvertently memorize more private information as a result of its improved generalization ability and the broader range of feature extraction. To explore this, we conduct the ablation experiments to record the variation in averaged mutual information differences for different batch sizes across all models and various sub-epochs, as shown in the Figures \ref{heatmap_lenet}-\ref{heatmap_mlp}.

The x-axis represents the configured checkpoint sub-epochs during the training phase, while the y-axis denotes the batch size when training deep learning models, which refers to the number of training samples processed by the model in one sub-epoch before updating its parameters. Each cell combines two key pieces of information: the color indicates the mutual information difference value with the scale provided on the color bar to the right. Darker colors represent larger mutual information difference values, while lighter colors indicate smaller values. Additionally, the text within each cell represents the training accuracy achieved for the corresponding batch size and sub-epoch combination. This visualization highlights the interplay between these factors and their impact on mutual information difference.

From the heatmaps, we can draw the following conclusions:

$\bullet$ For different batch sizes, we observe that smaller batch sizes result in faster model convergence. This can be attributed to the fact that smaller batch sizes allow the model to perform more frequent parameter updates within each sub-epoch, enabling it to adjust weights more frequently and respond more quickly to changes in the loss function. This higher update frequency accelerates the convergence process, especially during the early stages of training. Additionally, smaller batch sizes introduce greater gradient noise, which acts as an "exploration" mechanism, helping the model escape local minima or saddle points and optimize the loss function more effectively.

$\bullet$ For any batch size, when the batch size is fixed, the mutual information difference generally follows a "rise-then-fall" pattern as the number of sub-epochs increases. This observation aligns with the analysis presented in the paper. It indicates that while the size of the batch may affect training speed and convergence behavior, it does not alter the overall trend of mutual information difference during training. This trend reflects the dynamic process of the model transitioning from feature extraction to information compression during training: in the early stages, the model gradually learns and extracts significant features from the data, leading to an increase in the mutual information difference. As training progresses, the model compresses irrelevant information and stabilizes its representations, resulting in a decrease in the mutual information difference. Therefore, regardless of the batch size, the trend in mutual information difference is primarily driven by the optimization objectives and the learning mechanisms of the model, rather than the batch size.

$\bullet$ As the batch size decreases, the sub-epoch corresponding to the maximum mutual information difference (the darkest cell) also decreases. This is because smaller batch sizes lead to faster training convergence, allowing the model to reach a near-converged state at earlier sub-epochs, which subsequently triggers the decline in mutual information difference. This phenomenon indicates that the size of the batch not only affects the speed of convergence but also determines the timing of key processes during training, such as feature extraction and information compression. Consequently, the batch size regulates the timing of the peak in mutual information difference during the training process.

\textbf{Summary:} The model's adversarial robustness is closely tied to its state and training dynamics, with batch size playing a crucial role in the training process. Varying batch sizes lead to differences in the frequency of parameter updates and the level of gradient noise, which in turn affect the model’s training trajectory and convergence speed. As a result, models trained with different batch sizes exhibit varying levels of test accuracy, overall performance, and adversarial robustness. By influencing the dynamic changes in the model’s state, batch size may also affect patterns of information extraction and compression, potentially impacting the risk of information leakage from the model.

\end{document}